\documentclass[twoside,11pt]{article}

\usepackage{jmlr2e}[preprint]

\usepackage[english]{babel}
\usepackage[utf8]{inputenc}
\usepackage{algorithm}
\usepackage{algpseudocode}
\usepackage{pifont}
\usepackage{comment}
\usepackage{amsfonts}
\usepackage{graphicx}
\usepackage{hyperref}
\usepackage{float}
\usepackage{color}
\usepackage{xcolor}
\usepackage{subcaption}
\usepackage{graphicx}
\usepackage{caption}
\usepackage{booktabs}

\usepackage{mathtools}
\mathtoolsset{showonlyrefs}

\ShortHeadings{Forecasting Future and Past Trends in Time Series}{Coelho, Costa and Ferrás}
\firstpageno{1}

\begin{document}


\title{Neural Chronos ODE: Unveiling Temporal Patterns and Forecasting Future and Past Trends in Time Series Data}

\author{\name Cecília Coelho \email cmartins@cmat.uminho.pt\\
\name M. Fernanda P. Costa \email mfc@math.uminho.pt\\
\addr Centre of Mathematics, University of Minho\\
Braga, 4710-057, Portugal
\AND
\name Luís L. Ferrás \email lferras@fe.up.pt\\
       \addr Centre of Mathematics, University of Minho\\
       Braga, 4710-057, Portugal \\
       \addr Department of Mechanical Engineering (Section of Mathematics) - FEUP \\ University of Porto \\ Porto, 4200-465, Portugal
}

\editor{My editor}

\maketitle

\begin{abstract}
This work introduces Neural Chronos Ordinary Differential Equations (Neural CODE), a deep neural network architecture that fits a continuous-time ODE dynamics for predicting the chronology of a system both forward and backward in time. To train the model, we solve the ODE as an initial value problem and a final value problem, similar to Neural ODEs. We also explore two approaches to combining Neural CODE with Recurrent Neural Networks by replacing Neural ODE with Neural CODE (CODE-RNN), and incorporating a bidirectional RNN for full information flow in both time directions (CODE-BiRNN), and variants with other update cells namely GRU and LSTM: CODE-GRU, CODE-BiGRU, CODE-LSTM, CODE-BiLSTM.

Experimental results demonstrate that Neural CODE outperforms Neural ODE in learning the dynamics of a spiral forward and backward in time, even with sparser data. We also compare the performance of CODE-RNN/-GRU/-LSTM and CODE-BiRNN/-BiGRU/-BiLSTM against ODE-RNN/-GRU/-LSTM on three real-life time series data tasks: imputation of missing data for lower and higher dimensional data, and forward and backward extrapolation with shorter and longer time horizons. Our findings show that the proposed architectures converge faster, with CODE-BiRNN/-BiGRU/-BiLSTM consistently outperforming the other architectures on all tasks. 

\end{abstract}


\section{Introduction}




Neural Ordinary Differential Equations (Neural ODEs) are a type of Neural Networks (NNs) that fit an ODE to model the continuous dynamics of time-series data. Unlike traditional NNs, ODEs are continuous-time functions, allowing predictions to be made at any point in time and handling data taken at arbitrary times \citep{chenNeuralOrdinaryDifferential2019, rubanovaLatentOrdinaryDifferential2019}.

Time-series data is commonly used to describe phenomena that change over time and have a fixed order, with consecutive values conveying dependence. However, traditional NNs treat each input as independent, making it difficult to perceive the time relation between consecutive inputs \citep{hewamalageRecurrentNeuralNetworks2021}. Recurrent Neural Networks (RNNs) were introduced to handle this issue by adding a recurrence mechanism, allowing for the learning of dependencies between current and previous values and handling arbitrary length input and output sequences \citep{elmanFindingStructureTime1990}.

However, in certain tasks such as natural language processing (NLP) and speech recognition, the current value depends on both previous and future values too, making it important to use future information when making predictions \citep{schusterBidirectionalRecurrentNeural1997}.




Bidirectional Recurrent Neural Networks (BiRNNs) were introduced as an extension to traditional RNNs, allowing for processing of data sequences in both forward and backward directions. This enables combining information from past and future values to make predictions \citep{schusterBidirectionalRecurrentNeural1997}.

Real-world time-series data may have missing values or be sampled irregularly, creating challenges for RNNs that are designed to recognise order but not time intervals. Preprocessing is often required to convert the data into regularly sampled intervals, which can lead to errors and loss of information due to the frequency of sampling \citep{chenNeuralOrdinaryDifferential2019,rubanovaLatentOrdinaryDifferential2019}.

To address these challenges, ODE-RNNs were proposed. These are RNNs where state transitions are governed by a continuous-time model adjusted by a Neural ODE, eliminating the need for preprocessing by capturing the time dependence of sequential data \citep{rubanovaLatentOrdinaryDifferential2019}.

Neural ODEs solve Initial Value Problems (IVPs) by computing the solution curve forward in time from $t_0$ to $t_f$. However, in practice, predicting a system's behaviour requires leveraging data both forwards and backwards in time. This motivates the development of Neural Chronos\footnote{Greek word for time.} ODE (Neural CODE), an NN architecture that adjusts an ODE using information from both initial and final value problems. Neural CODE can predict the parameters of a previous condition of a system, unlike traditional Neural ODEs.

We compare Neural CODE to Neural ODE in predicting spiral dynamics in both time directions and with varying amounts of training data. Our results show that Neural CODE achieves better generalisation and robustness when predicting forward in time, thanks to the higher amount of information retrieved from the data. Neural CODE also improves backward predictive performance, as expected.

To leverage the advantages of Neural CODE, we refined the ODE-RNN model of \cite{rubanovaLatentOrdinaryDifferential2019} by replacing the Neural ODE by a Neural CODE into ODE-RNN (CODE-RNN). Two variants of the architecture were created by changing the update cell type to GRU and LSTM denoted by CODE-GRU and CODE-LSTM, respectively. However, Neural CODE requires both an initial and final condition to solve forward and backward, respectively. This is problematic for the CODE-RNN architecture, which has a single RNN cell that receives an intermediate hidden state $\boldsymbol{h}_i'$ from a Neural ODE and outputs a single hidden state $\boldsymbol{h}_i$. We address this issue by adding a second RNN cell to the CODE-RNN architecture, given rise to a new architecture CODE-BiRNN. This architecture uses two separate RNN cells to process data forwards and backwards in time, allowing for the incorporation of both initial and final conditions. CODE-BiRNN is further refined by creating two variants, CODE-BiGRU and CODE-BiLSTM, which use GRU and LSTM update cells, respectively. Table \ref{tab:archs} shows the state transitions and the recurrence mechanism  used in each of the continuous-times architectures studied and proposed in this work.

\begin{table}[h]
\centering
\begin{tabular}{lll}
\hline
Continuous-time architectures         & State transitions & Recurrence mechanism \\ \hline
Neural ODE \citep{chenNeuralOrdinaryDifferential2019}             & ODE               & ---                  \\
\textbf{Neural CODE (ours)} & \textbf{CODE}  & ---        \\
ODE-RNN \citep{rubanovaLatentOrdinaryDifferential2019}                & ODE               & RNN                  \\
ODE-GRU \citep{rubanovaLatentOrdinaryDifferential2019}                & ODE               & GRU                  \\
ODE-LSTM \citep{lechnerLearningLongTermDependencies2020}               & ODE               & LSTM                 \\
\textbf{CODE-RNN (ours)}    & \textbf{CODE}  & \textbf{RNN}         \\
\textbf{CODE-GRU (ours)}    & \textbf{CODE}  & \textbf{GRU}         \\
\textbf{CODE-LSTM (ours)}   & \textbf{CODE}  & \textbf{LSTM}        \\
\textbf{CODE-BiRNN (ours)}    & \textbf{CODE}  & \textbf{BiRNN}       \\
\textbf{CODE-BiGRU (ours)}    & \textbf{CODE}  & \textbf{BiGRU}       \\
\textbf{CODE-BiLSTM (ours)}   & \textbf{CODE}  & \textbf{BiLSTM}      \\ \hline
\end{tabular}%
\caption{Evolution of the continuous-time architectures proposed. Neural CODE is an evolution of Neural ODE and is also integrated into the recurrent model ODE-RNN, including into its variants (ODE-GRU/-LSTM) CODE-RNN/-GRU/-LSTM. To incorporate RNN updates into the recurrent state during both forward and backward passes of CODE-RNN, an architecture with BiRNN was proposed including its variants, CODE-BiRNN/-BiGRU/BiLSTM.}
\label{tab:archs}
\end{table}

This paper is organised as follows. Section \ref{sec:background} presents a brief review of essential concepts such as Neural ODE, RNN, ODE-RNN and BiRNN.

Section \ref{sec:TBANeuralODE} is devoted to the three novel deep neural network architectures. First, the Neural CODE architecture is presented along with the mathematical formulation of the initial and final value problems, followed by the proof of existence and uniqueness of solution.
Then, the two recurrent architectures CODE-RNN and CODE-BiRNN, are explained in detail. Followed by the presentation of their respective variants, CODE-GRU, CODE-LSTM, CODE-BiGRU, CODE-BiLSTM.


In Section \ref{sec:experiments} we evaluate and compare the performance of the newly proposed NNs namely, Neural CODE, CODE-RNN, CODE-GRU, CODE-LSTM, CODE-BiRNN, CODE-BiGRU and CODE-BiLSTM with the baseline NNs Neural ODE, ODE-RNN, ODE-GRU and ODE-LSTM.

The paper ends with the conclusions and future work in Section \ref{sec:conclusion}.

\section{Background} \label{sec:background}

This section aims to provide some background information on the developments discussed in Section 3. We will introduce a brief description of Neural ODE, RNN, ODE-RNN, and BiRNN.

\medskip

Let $\boldsymbol{X}=(\boldsymbol{x}_1, \boldsymbol{x}_2, \dots, \boldsymbol{x}_N)$ be a time-series of length $N$, with $\boldsymbol{x}_i \in  \mathbb{R}^d$ the data vector at time step $i$ ($i=1,\dots,N$). Let $\boldsymbol{Y}=(\boldsymbol{y}_1, \boldsymbol{y}_2, \dots, \boldsymbol{y}_N)$ be the ground-truth output time-series, with $\boldsymbol{y}_i \in  \mathbb{R}^{d^*}$ the response vector at time step $i$, and, let  $\boldsymbol{\hat{Y}}=(\boldsymbol{\hat{y}}_1, \boldsymbol{\hat{y}}_2, \dots, \boldsymbol{\hat{y}}_N)$ be the output response sequential data produced by a NN, with $\boldsymbol{\hat{y}}_i \in  \mathbb{R}^{d*}$ the output response vector at time step $i$.

\subsection{Neural ODE}

In traditional NNs, a sequential arrangement of hidden layers $\boldsymbol{h}_i$ is employed to transform an input vector $\boldsymbol{x}_i \in \mathbb{R}^d$ into the desired predicted output $\boldsymbol{\hat{y}}_i \in \mathbb{R}^{d^*}$ \citep{chenNeuralOrdinaryDifferential2019}. This transformation is achieved by propagating the output of each layer to the next one, following equation \eqref{eq:bck_NODE1} until reaching the output layer and obtaining the desired predicted output as indicated in equation \eqref{eq:bck_NODE2},

\begin{equation}
\label{eq:bck_NODE1}
    \boldsymbol{h}_i = \sigma(\boldsymbol{W}^{[i]} \boldsymbol{h}_{i-1}+ \boldsymbol{b}^{[i]}),
\end{equation}

\begin{equation}
\label{eq:bck_NODE2}
    \boldsymbol{\hat{y}}_i = \sigma(\boldsymbol{W}^{[out]} \boldsymbol{h}_{i}+ \boldsymbol{b}^{[out]}).
\end{equation}

Here, $\boldsymbol{h}_0$ is the input layer, $\sigma$ an activation function, $\boldsymbol{W}^{[i]} \in \mathbb{R}^{n \times d}$ the weights matrix of layer $i$ with $n$ neurons, $\boldsymbol{b}^{[i]} \in \mathbb{R}^n$ the biases of layer $i$, $\boldsymbol{W}^{[out]} \in \mathbb{R}^{d^* \times n}$ the weights matrix of the output layer and $\boldsymbol{b}^{[out]} \in \mathbb{R}^{d^*}$ the biases of the output layer.


In \cite{chenNeuralOrdinaryDifferential2019}, the authors introduced Neural ODE, a NN architecture that models hidden states using a continuous-time function. Unlike traditional NNs that rely on discrete layers, Neural ODE employs an infinite number of hidden layers, approaching a continuum. The output of these layers is defined as the solution of an IVP,

\begin{equation}
    \dfrac{d \boldsymbol{h}(t)}{dt}=\boldsymbol{f_\theta}(\boldsymbol{h}(t),t) \,\,\,  \text{with} \,\,\, \boldsymbol{h}(t_0)=\boldsymbol{h}_0
\label{eq:IVP}
\end{equation}

\noindent where $\boldsymbol{f_\theta}$ is given by a NN, parameterised by a set of weights and biases, $\boldsymbol{\theta}$, $t$ is the time step of the solution and $(\boldsymbol{h}_0, t_0)$ is the initial condition of the IVP.

One significant advantage of training a Neural ODE is that it produces a continuous-time function. Consequently, predictions can be made at any desired time point by discretizing the function using a numerical ODE solver:

\begin{equation}
\label{eq:bck_NODE3}
    \boldsymbol{h}(t) = ODESolve(\boldsymbol{f_\theta}, \boldsymbol{h}_0, (t_0,t_f)), 
\end{equation}
with $\boldsymbol{h}(t)$ the solutions computed by solving the IVP \eqref{eq:IVP} in the time interval $(t_0, t_f)$.

Taking $\boldsymbol{f_\theta}(\boldsymbol{h}(t), t)$ to be a NN, which builds an ODE dynamics with learnable parameters $\boldsymbol{\theta}$ and the input vector $(\boldsymbol{x},t)\in \mathbb{R}^{d+1}$ that corresponds to the first time step $t_0$ of the time-series, the input layer $\boldsymbol{h}_0$ is given by \eqref{eq:bck_ODE5}:

\begin{equation}
\label{eq:bck_ODE5}
    \boldsymbol{h}(t_0) = \boldsymbol{h}_0 = \sigma \left( \boldsymbol{W}^{[in]} (\boldsymbol{x},t) + \boldsymbol{b}^{[in]} \right),
\end{equation}
where $\boldsymbol{W}^{[in]} \in \mathbb{R}^{n \times d+1}$ and $\boldsymbol{b}^{[in]} \in \mathbb{R}^n$ are the weight matrix and the bias vector of the input layer, respectively.

\begin{figure}[h]
    \centering
    \includegraphics[scale=0.4]{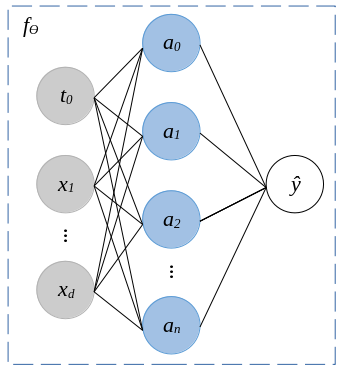}
    \caption{The adjusted ODE dynamics, $\boldsymbol{f_\theta}$, is given by a NN with an input layer with the dimension $d+1$ to accommodate the input vector, $\boldsymbol{x} \in \mathbb{R}^d$ with $d$ features, and its time step, $t \in \mathbb{R}$.}
    \label{fig:nnODEDynamics}
\end{figure}

Consider a single hidden layer, $\boldsymbol{h} \in \mathbb{R}^n$, with $n$ neurons, $a_n$, a batch size of $1$ given by the Neural ODE, $\boldsymbol{f_\theta}$ (see Figure \ref{fig:nnODEDynamics}).

The dynamics built by such a NN has the form:

\begin{equation}
\label{eq:bck_ODE6}
    \boldsymbol{f_\theta}(\boldsymbol{h}(t),t) = \sigma \left( \boldsymbol{W}^{[out]} (\sigma(  \boldsymbol{W}^{[1]} \boldsymbol{h}(t_0)+ \boldsymbol{b}^{[1]} )) + \boldsymbol{b}^{[out]} \right)
\end{equation}
where $\sigma$ represents an activation function, $\boldsymbol{h}(t_0) \in \mathbb{R}^{n}$ is the output of the input layer, \eqref{eq:bck_ODE5}, $\boldsymbol{W}^{[1]} \in \mathbb{R}^{n \times n}, \boldsymbol{W}^{[out]} \in \mathbb{R}^{d^* \times n}$ are the matrix of the weights of the hidden and output layers and $\boldsymbol{b}^{[1]} \in \mathbb{R}^n, \boldsymbol{b}^{[out]} \in \mathbb{R}^{d^*} $ are the bias of the hidden and output layers, respectively.

Upon examining \eqref{eq:bck_ODE5} and \eqref{eq:bck_ODE6}, it becomes apparent that a NN is a composition of linear transformations with the application of various activation functions $\sigma$. These activation functions can either be linear (such as identity or binary step functions) or non-linear (including sigmoid, hyperbolic tangent (tanh), rectified linear unit (ReLU), softmax, etc.).

When all activation functions in the NN that constructs $\boldsymbol{f_\theta}$ are linear, it results in a linear ODE. However, linear activation functions are not commonly used as they cannot effectively capture complex relationships within the data. Consequently, they limit the representational power of the NN and hinder its performance in the given task \citep{Goodfellow-et-al-2016}. On the other hand, by incorporating one or more non-linear activation functions in the dynamics builder NN, it becomes a non-linear ODE.

In addition to the NN that models the function dynamics $\boldsymbol{f_\theta}$ and by optimising its parameters $\boldsymbol{\theta}$, a Neural ODE also incorporates an ODE solver. The ultimate outcome of training a Neural ODE is the learned function $\boldsymbol{f_\theta}$. To make predictions $\boldsymbol{\hat{y}}_i,\,\, i \in 1,\dots,N$ at specific time points $t_i$ within the prediction time interval $(t_1, t_N)$, the IVP is solved. The solutions obtained from solving the IVP, $\boldsymbol{h}(t_i) \equiv \boldsymbol{h}_i$, represent the predictions at each respective time step $t_i, \,\, i \in 1,\dots,N$.

\textbf{Remark:} In some cases the solutions $\boldsymbol{h}(t_i)$ obtained from the Neural ODE do not directly correspond to the desired predictions $\boldsymbol{\hat{y}}_i$. In such cases, an additional NN can be employed to process the intermediate solutions $\boldsymbol{h}(t_i)$ and produce the final predictions $\boldsymbol{\hat{y}}_i$.

\subsection{RNNs}

RNNs \citep{elmanFindingStructureTime1990} are a type of artificial NN that are specifically designed to process sequential data. Unlike feed-forward NNs, which process inputs in a single pass and do not have memory of previous inputs, RNNs have a form of memory that allows them to retain and utilise information from previous steps or time points in the sequence.

The key characteristic of an RNN is its recurrent connection, which forms a loop that allows information to be passed from one step to the next (from one hidden state at time step $t_{i-1}$, $\boldsymbol{h}_{i-1}$, into the next one, $\boldsymbol{h}_i$). This loop enables the network to maintain an internal state or hidden state that captures the context and dependencies of the previous steps. The hidden state serves as a memory that encodes information about the sequence up to the current step, allowing the network to make decisions or predictions based on the entire previous observations rather than just the current input. Mathematically, a RNN cell which forms the layers of an RNN, can be defined for $i=1,\dots,N$:

\begin{equation}
\label{eq:bck_RNN1}
\boldsymbol{h}_i = \sigma(\boldsymbol{W}^{\text{[feedback]}} \boldsymbol{h}_{i-1} + \boldsymbol{W}^{\text{[in]}} \boldsymbol{x}_i + \boldsymbol{b}^{[i]})    
\end{equation}

\noindent where $\boldsymbol{h}_i \in \mathbb{R}^n$ is the initial hidden state vector $\boldsymbol{h}_0$ is initialised at the start of the recurrent computation, $\boldsymbol{W}^{\text{[feedback]}} \in \mathbb{R}^{n \times n}$ is the recurrent matrix that contains the weights that link the two hidden state vectors at time step $t_{i-1}$ and $t_i$, $\boldsymbol{W}^{\text{[in]}} \in \mathbb{R}^{n \times d}$ is the matrix of the weights between the input and all hidden state vectors, $\boldsymbol{x}_i \in \mathbb{R}^d$, $\boldsymbol{b}^{[i]} \in \mathbb{R}^n$ is the bias vector of the current hidden state and $\sigma$ is an arbitrary activation function.

At each step $t_i$ in the sequence, an RNN takes an input, processes it along with the current hidden state, and produces an output and an updated hidden state. The input can be any form of sequential data, such as a time-series, a sentence, or a sequence of images. The output can be a prediction, a classification, or another sequence of values.


The update of an RNN cell at a time step $t_i$ is usually represented concisely as:

\begin{equation}
\label{eq:bck_RNN2}
\boldsymbol{h}_i = RNNCell(\boldsymbol{h}_{i-1}, \boldsymbol{x}_i).
\end{equation}

The predicted output at time step $t_i$, $\boldsymbol{\hat{y}}_i$, is computed by applying an output layer to the hidden state $\boldsymbol{h}_i$ \citep{elmanFindingStructureTime1990}:

\begin{equation}
\label{eq:bck_RNN3}
\boldsymbol{\hat{y}}_i = \sigma(\boldsymbol{W}^{\text{[out]}} \boldsymbol{h}_i + \boldsymbol{b}^{\text{[out]}})
\end{equation}

 \noindent where $\boldsymbol{W}^{\text{[out]}} \in \mathbb{R}^{d^* \times n}$ is the matrix of the weights between the hidden state and the output vectors,  $\boldsymbol{b}^{\text{[out]}} \in \mathbb{R}^{d^*}$ is the bias vector of the output layer and $\boldsymbol{\hat{y}}_i \in \mathbb{R}^{d^*}$ is the prediction at time step $i$.

\subsection{ODE-RNN}



RNNs adjust a discrete-time dynamics with states defined at each time step, denoted as $t_i$. However, the states between data observations, occurring in the interval $[t_{i}, t_{i+1}]$, remain undefined. Moreover, RNNs face challenges when handling irregularly sampled data due to the inability to perceive the sampling intervals between observations. Consequently, preprocessing of such data becomes essential to achieve regular sampling intervals, although this process introduces potential errors \citep{rubanovaLatentOrdinaryDifferential2019}.

To address these limitations, \cite{rubanovaLatentOrdinaryDifferential2019} introduced the ODE-RNN architecture, which defines the states between observations using an ODE.

Instead of using the previous hidden state $\boldsymbol{h}_{i-1}$, as in \eqref{eq:bck_RNN2}, for the RNN update, it employs an intermediate hidden state $\boldsymbol{h}'_i$ instead,

\begin{equation}
\label{eq:bck_ODERNN1}
\boldsymbol{h}_i = RNNCell(\boldsymbol{h}'_{i}, \boldsymbol{x}_i).
\end{equation}
The intermediate hidden states are obtained as the solution of an IVP:

\begin{equation}
\label{eq:IVPODE-RNN}
    \dfrac{d \boldsymbol{h'}(t)}{dt} = \boldsymbol{f_\theta} (\boldsymbol{h}(t),t) \,\,\, \text{with} \,\,\, \boldsymbol{h}(t_{i-1}) = \boldsymbol{h}_{i-1}
\end{equation}

Thus, the solution is given by:

\begin{equation}
\label{eq:bck_ODERNN2}
\boldsymbol{h}'_i = ODESolve(\boldsymbol{f_\theta}, \boldsymbol{h}_{i-1},(t_{i-1}, t_i))
\end{equation}

\noindent where $\boldsymbol{h}'_i \in \mathbb{R}^n$, $\boldsymbol{f_\theta}$ is the ODE dynamics given by the NN with parameters $\boldsymbol{\theta}$, $\boldsymbol{h}_{i-1}$ is the previous hidden state, and $t_{i-1}$ and $t_i$ are the time steps of the previous and current hidden state, respectively.


ODE-RNNs are well-suited for handling irregularly sampled data due to their ability to leverage the sampling intervals and adapt to time-dependent dynamics. This advantage eliminates the necessity for preprocessing steps, which are typically required when using traditional RNNs.

\subsection{BiRNNs}




RNNs have been widely used, within encoder-decoder architectures, in various applications such as machine translation \citep{choLearningPhraseRepresentations2014}, NLP \citep{sutskeverSequenceSequenceLearning2014}, and time-series prediction \citep{hewamalageRecurrentNeuralNetworks2021}.

As previously mentioned, RNNs enable the prediction of the output at a specific time step $t_i$, $\boldsymbol{\hat{y}}_i$, by using past information stored in the hidden state from the previous time step, $\boldsymbol{h}_{i-1}$. This captures the dependency between the current value and the preceding one \citep{elmanFindingStructureTime1990}. However, certain tasks, such as NLP and machine translation, require predictions that depend not only on the past but also on the future. Therefore, incorporating future information can lead to improved predictive performance.

To address this, \cite{schusterBidirectionalRecurrentNeural1997} introduced BiRNNs, an architecture composed of two stacked RNNs. The first RNN receives the input sequence in its original order, from $\boldsymbol{x}_1$ to $\boldsymbol{x}_N$, while the second RNN receives the same sequence but in reverse order, from $\boldsymbol{x}_N$ to $\boldsymbol{x}_1$. Consequently, one row of the unfolded RNNs captures the dependence between the previous value and the current one through forward hidden states, denoted as $\boldsymbol{\overrightarrow{h}}_i$. Simultaneously, the other row captures the dependence between the next value and the current one through backward hidden states, denoted as $\boldsymbol{\overleftarrow{h}}_i$. This is represented in the two following equations for $i=1,\dots,N$:

\begin{equation}
\label{eq:bck_BiRNN1}
\boldsymbol{\overrightarrow{h}}_i = \sigma (\boldsymbol{\overrightarrow{W}}^{\text{[feedback]}} \boldsymbol{\overrightarrow{h}}_{i-1} + \boldsymbol{W}^{[f]} \boldsymbol{x}_{i}  + \boldsymbol{b}^{[f]}),
\end{equation}

\begin{equation}
\label{eq:bck_BiRNN2}
\boldsymbol{\overleftarrow{h}}_i = \sigma (\boldsymbol{\overleftarrow{W}}^{\text{[feedback]}} \boldsymbol{\overleftarrow{h}}_{i-1} + \boldsymbol{W}^{[b]} \boldsymbol{x}_{i} + \boldsymbol{b}^{[b]}),
\end{equation}

\noindent where $\boldsymbol{W}^{[f]} \in \mathbb{R}^{n \times d}$ is the weight matrix between the input and forward hidden state vectors, $\boldsymbol{b}^{[f]} \in \mathbb{R}^n$ is the bias vector of the forward hidden states, $\boldsymbol{W}^{[b]} \in \mathbb{R}^{n \times d}$ is the weight matrix between the input and backward hidden states and $\boldsymbol{b}^{[b]} \in \mathbb{R}^n$ the bias vector of the backward hidden states.

\begin{figure}[h]
    \centering
    \includegraphics[]{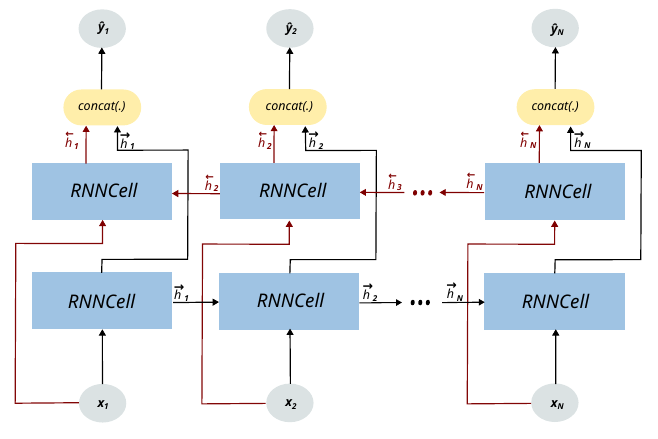}
    \caption{Scheme of a BiRNN unfolded through time. }
    \label{fig:BiRNN}
\end{figure}

To get the predicted output at a given time step, $\boldsymbol{\hat{y}}_i$, the full hidden state, $\boldsymbol{h}_i$, is computed by performing an aggregation/merging operation (concatenate, sum, mean, etc.) of the forward, $\boldsymbol{{\overrightarrow{h}}}_i$ , and backward, $\boldsymbol{{\overleftarrow{h}}}_i$,  hidden states of the same time step $i$ \citep{schusterBidirectionalRecurrentNeural1997} as follows, for $i=1,\dots,N$,

\begin{equation}
\label{eq:bck_BiRNN3}
\boldsymbol{\hat{y}}_i = \sigma(\boldsymbol{W}^{\text{[out]}} \boldsymbol{h}_i + \boldsymbol{b}^{\text{[out]}}) \,\,\, \text{with} \,\,\, \boldsymbol{h}_i = \boldsymbol{\overrightarrow{h}}_i \oplus \boldsymbol{\overleftarrow{h}}_i,
\end{equation}

\noindent where $\oplus$ denotes the aggregation/merging operation used. In our study and experiments, we chose concatenation. Figure \ref{fig:BiRNN} depicts the architecture of a BiRNN.

If multiple layers of BiRNNs are used, the complete hidden state, $\boldsymbol{h}_i$, is the input for the subsequent BiRNN \citep{schusterBidirectionalRecurrentNeural1997}.

\section{Method} \label{sec:TBANeuralODE}


In this section, we introduce the innovative architecture called Neural CODE, providing the mathematical details that underlie its workings. Additionally, we demonstrate the versatility of Neural CODE by applying it to enhance the ODE-RNN family through the introduction of two new distinct architectures: CODE-RNN and CODE-BiRNN.

\subsection{Neural CODE}




In this work, we propose a novel continuous-time model called Neural CODE. It builds upon the concepts of Neural ODEs but introduces a new strategy to adjust the ODE dynamics by leveraging possible relationships between past and current as well as future and current values. 
This idea was directly inspired by BiRNNs and, by allowing both past and future information to influence the current prediction, Neural CODE aims to enhance the model's ability to handle long-range dependencies and capture subtle patterns in the data.

Thus, the ODE dynamics $\boldsymbol{f_\theta}$ is adjusted by optimising the parameters $\boldsymbol{\theta}$ using the information given by capturing the dependence between the previous and the current values through forward hidden states $\boldsymbol{\overrightarrow{h}}_i$, as well as the dependence between the next and the current values through backward hidden states $\boldsymbol{\overleftarrow{h}}_i$.

In this context, the IVP \eqref{eq:IVP} is solved in a forward way within the time interval $(t_0, t_f)$, using a numerical ODE solver. 

\begin{equation}
    \overrightarrow{\boldsymbol{h}}_i = ODESolve(\boldsymbol{f_\theta}, \boldsymbol{h}_f, (t_0, t_f)).
\end{equation}

For considering possible relationships between next and current values, it is defined a Final Value Problem (FVP) as follows: 

\begin{equation}
    \dfrac{d \boldsymbol{h}(t)}{dt}=\boldsymbol{f_\theta}(\boldsymbol{h}(t),t) \,\,\,  \text{with} \,\,\, \boldsymbol{h}(t_f)=\boldsymbol{h}_f,
\label{eq:FVP}
\end{equation}

\noindent where the final condition is given by the value at the last time step, $\boldsymbol{h}_f$. Furthermore, the FVP is solved in a backward way within the reverse time interval $(t_f, t_0)$:

\begin{equation}
    \overleftarrow{\boldsymbol{h}}_i = ODESolve(\boldsymbol{f_\theta}, \boldsymbol{h}_f, (t_f, t_0)).
\end{equation}

\begin{figure}[h]
    \centering
    \includegraphics[]{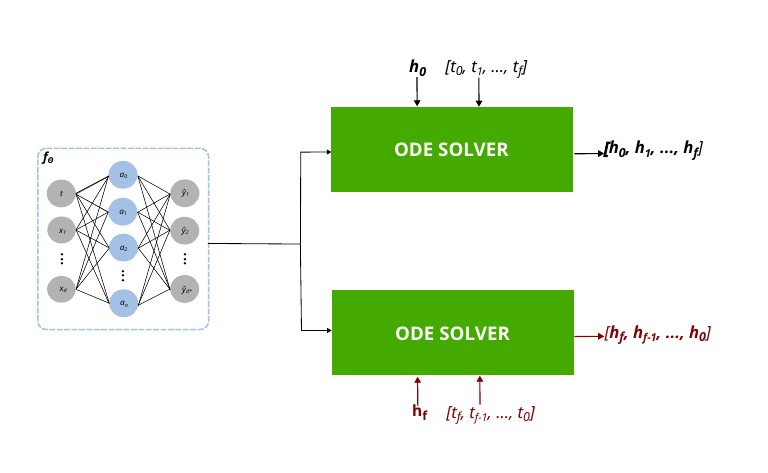}
    \caption{Scheme of a Neural CODE able to make predictions forward, by solving an IVP, and backward, by solving an FVP, in time.}
    \label{fig:TBANeuralODE}
\end{figure}

Note that, by solving the IVP \eqref{eq:IVP} and FVP \eqref{eq:FVP} two sets of solutions are obtained. The first set, denoted as $\boldsymbol{\overrightarrow{h}}_i$, represents the solutions computed in the positive direction of time. The second set, $\boldsymbol{\overleftarrow{h}}_i$, corresponds to the solutions computed in the negative direction of time. For a visual representation, see Figure \ref{fig:TBANeuralODE}.

\begin{algorithm}[h]
\caption{Neural CODE training process.}
\label{alg:TBA}
\begin{algorithmic}
\State \textbf{Input:} start time $t_0$, end time $t_f$, initial condition $(\boldsymbol{h}_0, t_0)$, final condition $(\boldsymbol{h}_f, t_f)$, maximum number of iterations $MAXITER$; 
\State Choose $Optimiser$;
\State $\boldsymbol{f_\theta} = DynamicsNN()$;
\State Initialise $\boldsymbol{\theta}$;
\For{$k=1:MAXITER$}
    \State $\boldsymbol{\overrightarrow{h}}_{i} \leftarrow ODESolve(\boldsymbol{f_\theta}, \boldsymbol{h}_{0}, (t_{0},t_f))$;
    \State $\boldsymbol{\overleftarrow{h}}_{i} \leftarrow ODESolve(\boldsymbol{f_\theta}, \textbf{h}_{f}, (t_{f},t_{0}))$;
    \State Evaluate loss $\mathcal{L}_{\text{total}}$ using \eqref{eq:L_total};
    \State $\nabla \mathcal{L}_{\text{total}} \leftarrow \text{Compute gradients of } \mathcal{L}_{\text{total}}$;
    \State $\boldsymbol{\theta} \leftarrow Optimiser.Step(\nabla \mathcal{L}_{\text{total}})$;
\EndFor
\State \textbf{Return:} $\boldsymbol{\theta}$;
\end{algorithmic}
\end{algorithm}

To optimise the parameters $\boldsymbol{\theta}$ of the NN we designed a loss function given by the sum of two terms, namely the mean squared error (MSE) of the forward and backward predictions, $\mathcal{L}(\boldsymbol{\overrightarrow{h}}_{i})$ and $\mathcal{L}(\boldsymbol{{\overleftarrow{h}}}_{i})$ respectively:

\begin{align}
 \mathcal{L}_{\text{total}}(\boldsymbol{\theta}) =&\begin{multlined}
    \mathcal{L} \left( \boldsymbol{h}(t_0) + \int_{t_0}^{t_f} \boldsymbol{f_\theta}(\boldsymbol{h}(t), t) dt \right) + \mathcal{L} \left( \boldsymbol{h}(t_{f}) + \int_{t_f}^{t_0} \boldsymbol{f_\theta}(\boldsymbol{h}(t), t) dt \right) \end{multlined}\\ 
    =& \begin{multlined}
    \mathcal{L} \left( ODESolve(\boldsymbol{f_\theta}, \boldsymbol{h}_0,  (t_0, t_f)) \right) +  \mathcal{L} \left( ODESolve(\boldsymbol{f_\theta}, \boldsymbol{h}_f, (t_f, t_0)) \right) \end{multlined} \\
    =& \begin{multlined}
    \mathcal{L}(\boldsymbol{\overrightarrow{h}}_i) + \mathcal{L}(\boldsymbol{\overleftarrow{h}}_i)  \end{multlined}
\label{eq:L_total}
\end{align}

\noindent where $\mathcal{L}(\boldsymbol{\overrightarrow{h}}_t) = MSE \left( \overrightarrow{\boldsymbol{\hat{y}}}, \overrightarrow{\boldsymbol{y}} \right)$ and $\mathcal{L}(\boldsymbol{\overleftarrow{h}}_t) = MSE \left( \overleftarrow{\boldsymbol{\hat{y}}}, \overleftarrow{\boldsymbol{y}} \right)$, with $\overrightarrow{\boldsymbol{y}}$ and $\overleftarrow{\boldsymbol{y}}$ the ground-truth in the forward and backward directions.

By minimising the loss function $\mathcal{L}_{\text{total}}$, the optimisation process in Neural CODE aims to find the parameters $\boldsymbol{\theta}$ that can produce solutions to the IVP that closely match the true values and, simultaneously, solutions to the FVP that also closely match the true values.

Similarly to Neural ODEs, the adjoint sensitivity method is used to compute the gradients of the loss function \citep{chenNeuralOrdinaryDifferential2019, pontryaginMathematicalTheoryOptimal2018}.

After the optimisation process has been completed in training a Neural CODE, the ultimate outcome is the ODE dynamics $\boldsymbol{f_\theta}$, adjusted by a NN. To make predictions, a numerical ODE solver is used, which can solve an IVP by starting at the initial condition and solving forward in time (Algorithm \ref{alg:TBA_test}), or a FVP by starting at the final condition and solving backward in time (Algorithm \ref{alg:TBA_test2}).  

One advantage of Neural CODE is the ability to model the ODE dynamics to effectively capture the data dynamics when solved in both forward (IVP \eqref{eq:IVP}) and backward (FVP \eqref{eq:FVP}) time directions. Therefore, the proposed approach allows for the computation of valid predictions within or outside the training time interval. Within the training time interval, $t \in (t_0, t_f)$, the method is capable of performing data completion tasks, where missing data points are estimated. Moreover, the approach also facilitates prediction of past data, $t \in (t_m, t_f)$, where $t_m$ represents a time preceding time $t_0$, and prediction of future data, denoted as $t \in (t_0, t_M)$, where $t_M$ represents a time beyond time $t_f$.

\begin{algorithm}[h]
\caption{Neural CODE for making future predictions, by solving IVP \eqref{eq:IVP}.}
\label{alg:TBA_test}
\begin{algorithmic}
\State \textbf{Input:} NN $\boldsymbol{f_\theta}$, start time $t_0$, end time $t_M$,  initial condition $(\boldsymbol{h}_0, t_0)$; 
\State  $\boldsymbol{\overrightarrow{h}}_{i} \leftarrow ODESolve(\boldsymbol{f_\theta}, \boldsymbol{h}_{0}, (t_{0},t_M))$;
\State  $\overrightarrow{\boldsymbol{\hat{Y}}} \leftarrow \boldsymbol{\overrightarrow{h}}_{i}$;
\State \textbf{Return:} $\overrightarrow{\boldsymbol{\hat{Y}}}$;
\end{algorithmic}
\end{algorithm}

\begin{algorithm}[h]
\caption{Neural CODE for making past predictions, by solving FVP \eqref{eq:FVP}.}
\label{alg:TBA_test2}
\begin{algorithmic}
\State \textbf{Input:} NN $\boldsymbol{f_\theta}$, start time $t_f$, end time $t_m$,  final condition $(\boldsymbol{h}_f, t_f)$; 
\State  $\boldsymbol{\overleftarrow{h}}_{i} \leftarrow ODESolve(\boldsymbol{f_\theta}, \textbf{h}_{f}, (t_{f},t_{m}))$;
\State  $\overleftarrow{\boldsymbol{\hat{Y}}} \leftarrow \boldsymbol{\overleftarrow{h}}_{i}$;
\State \textbf{Return:} $\overleftarrow{\boldsymbol{\hat{Y}}}$;
\end{algorithmic}
\end{algorithm}

\textbf{Remark:} Note that, we consider that the solution $\boldsymbol{h}_i$ outputted by the ODE solver is the prediction $\boldsymbol{\hat{y}}_i$. However, in some cases the solutions $\boldsymbol{h}(t_i)$ obtained from solving the IVP or FVP may not directly correspond to the desired predictions $\boldsymbol{\hat{y}}_i$. In such cases, an additional NN can be employed to process the intermediate solutions $\boldsymbol{h}(t_i)$ to produce the final predictions $\boldsymbol{\hat{y}}_i$.

\subsubsection{Existence and uniqueness of the solution}

Neural CODE aims to adjust the dynamics of an ODE that makes valid predictions when solved both IVP \eqref{eq:IVP} and FVP \eqref{eq:FVP}, originating an Endpoint Value Problem (EVP) \citep{wangDefinitionNumericalMethod2018}.

To prove the existence and uniqueness of a solution to the EVP we use the Picard-Lindelöf theorem for both the initial and final value problems, \eqref{eq:IVP} and \eqref{eq:FVP} respectively. \\


\noindent
 {\bf Theorem 1} {\it Let $\boldsymbol{f_\theta}(\boldsymbol{h}, t) :  \mathbb{R}^d \times [t_0,t_f]  \to \mathbb{R}^{d}$, be a continuous vector-valued function that is Lipschitz continuous, \textit{i.e.}, there exists a constant $K$ such that for all $\boldsymbol{\boldsymbol{h}}_1, \boldsymbol{\boldsymbol{h}}_2 \in  \mathbb{R}^{d}$ and $ t \in [t_0,t_f]$,

$$||\boldsymbol{\boldsymbol{f_\theta}}(\boldsymbol{\boldsymbol{h}}_1, t) - \boldsymbol{\boldsymbol{f_\theta}}(\boldsymbol{\boldsymbol{h}}_2, t)|| \le K \lVert \boldsymbol{\boldsymbol{h}}_1 - \boldsymbol{\boldsymbol{h}}_2 \rVert.$$

\noindent Then, for any initial value $\boldsymbol{h}_0 \in \mathbb{R}^d$, the IVP \eqref{eq:IVP} has a unique solution on the interval $[t_0,t_f]$.
} 
\\

{\bf Proof}. The theorem can be proved by Picard's Existence and Uniqueness Theorem.

\textbf{Remark} It should be noted that:

\begin{enumerate}
    \item  
    

        \noindent
        By constructing $\boldsymbol{f_\theta}$, given by a NN, we ensure the continuity of the function in both variables since it is a composition of continuous functions. For any pair $({\boldsymbol{h}_i}, t_i) \in \mathbb{R}^{d} \times  \in [t_0, t_f]$, there is a neighbourhood such that $\boldsymbol{f_\theta}(\boldsymbol{h},t)$ is continuous within it.

    \item 

        \noindent
         Since the function $\boldsymbol{f_\theta}(\boldsymbol{h}, t)$, given by a NN, is a composition of Lipschitz functions of all layers, each with an associated Lipschitz constant, and the weights are finite, the function is Lipschitz continuous in $\boldsymbol{h}$ with $K$ the product of the Lipschitz constants of its layers, i.e.,
        $$||\boldsymbol{f_\theta}(\boldsymbol{h}_1,t) - \boldsymbol{f_\theta}(\boldsymbol{h}_2,t)|| \le K||\boldsymbol{h}_1 - \boldsymbol{h}_2||$$ holds for all $(\boldsymbol{h}_1, t), (\boldsymbol{h}_2, t) \in \mathbb{R}^{d} \times  \in [t_0, t_f]$
        \citep{goukRegularisationNeuralNetworks2020}. 

\end{enumerate}


\noindent
{\bf Theorem 2} {\it Let $\boldsymbol{f_\theta}(\boldsymbol{h}, t) :  \mathbb{R}^d \times [t_0,t_f]  \to \mathbb{R}^{d}$, be a continuous vector-valued function that is Lipschitz continuous, \textit{i.e.}, there exists a constant $K$ such that for all $\boldsymbol{\boldsymbol{h}}_1, \boldsymbol{\boldsymbol{h}}_2 \in  \mathbb{R}^{d}$ and $ t \in [t_0,t_f]$,

$$||\boldsymbol{\boldsymbol{f_\theta}}(\boldsymbol{\boldsymbol{h}}_1,t) - \boldsymbol{\boldsymbol{f_\theta}}(\boldsymbol{\boldsymbol{h}}_2,t)|| \le K \lVert \boldsymbol{\boldsymbol{h}}_1 - \boldsymbol{\boldsymbol{h}}_2 \rVert.$$

\noindent Then, for any final value $\boldsymbol{h}_f \in \mathbb{R}^d$, the FVP \eqref{eq:FVP} has a unique solution on the interval $[t_0,t_f]$.} \\


\noindent
{\bf Proof}.
As the form of an IVP \eqref{eq:IVP} and a FVP \eqref{eq:FVP} are similar \citep{wangDefinitionNumericalMethod2018} the proof of Theorem 2 
can also be proved using the Picard's Existence and Uniqueness Theorem
. {\color{black} We use time reversibility, and make the change of variable $\boldsymbol{H}(s)=\boldsymbol{h}(t_f-s)$ which results in the IVP, $\frac{d\boldsymbol{H}}{ds} = -\boldsymbol{f_\theta}(\boldsymbol{H},t_f-s)$, $\boldsymbol{H}(0) = \boldsymbol{h}(t_f) = \boldsymbol{\boldsymbol{h}}_f$.} \hfill\BlackBox


Since all the conditions needed for the Picard-Lindelöf theorem are satisfied by both IVP \eqref{eq:IVP} and FVP \eqref{eq:FVP}, then it is guaranteed the existence and uniqueness of the solutions.


\subsection{Recurrent architectures based on Neural CODE}

We now demonstrate the versatility of Neural CODE by applying it to enhance the ODE-RNN family through the introduction of two new distinct recurrent architectures: CODE-RNN and CODE-BiRNN. The reason to introduce recurrent architectures based-on Neural CODE is that this architecture adjusts an ODE dynamics which is determined by the initial (IVP \eqref{eq:IVP}) and final condition (FVP \eqref{eq:FVP}). By employing a recurrence mechanism we aim to update the ODE dynamics at each observation, having as many initial and final conditions as time steps, similar to ODE-RNN.

\subsubsection{CODE-RNN} 




As mentioned, with the aim of further improving the performance of predictions in both forward and backward directions of Neural CODE, we propose the recurrent recurrent architecture CODE-RNN. This new architecture is built by refining and redesigning the ODE-RNN architecture of \cite{rubanovaLatentOrdinaryDifferential2019} in which we replaced the Neural ODE by a Neural CODE. 

In CODE-RNN, the intermediate hidden state $\boldsymbol{h'}_i$ in the RNN cell update \eqref{eq:bck_ODERNN1}, is obtained by combining the forward intermediate hidden state, $\boldsymbol{\overrightarrow{h'}}_i$, and the backward intermediate hidden state, $\boldsymbol{\overleftarrow{h'}}_i$.

The forward intermediate hidden state, $\boldsymbol{\overrightarrow{h'_i}}$, is obtained as the solution of an IVP within the time interval between the previous and current time steps, $(t_{i-1}, t_i)$:

\begin{equation}
\label{eq:IVPCODE-RNN}
    \dfrac{d \boldsymbol{\overrightarrow{h'}}(t)}{dt} = \boldsymbol{f_\theta} (\boldsymbol{\overrightarrow{h'}}(t),t) \,\,\, \text{with} \,\,\, \boldsymbol{\overrightarrow{h'}}(t_{i-1}) = \boldsymbol{\overrightarrow{h'}}_{i-1}.
\end{equation}

On the other hand, the backward intermediate hidden state, $\boldsymbol{\overleftarrow{h'_i}}$, is obtained as the solution of a FVP within the time interval between the current and previous time steps, $(t_i, t_{i-1})$: 

\begin{equation}
\label{eq:FVPCODE-RNN}
    \dfrac{d \boldsymbol{\overleftarrow{h'}}(t)}{dt} = \boldsymbol{f_\theta} (\boldsymbol{\overleftarrow{h'}}(t),t) \,\,\, \text{with} \,\,\, \boldsymbol{\overleftarrow{h'}}(t_{i}) = \boldsymbol{\overleftarrow{h'}}_{i}.
\end{equation}

Thus, the solutions are given by:

\begin{equation}
\label{eq:TBAODERNN1}    
\boldsymbol{\overrightarrow{h'}}_i= ODESolve(\boldsymbol{f_\theta}, \boldsymbol{\overrightarrow{h'}}_{i-1}, (t_{i-1},t_i)),
\end{equation}

\begin{equation}
\label{eq:TBAODERNN2}    
\boldsymbol{\overleftarrow{h'}}_{i} = ODESolve(\boldsymbol{f_\theta}, \boldsymbol{\overleftarrow{h'}}_{i-1}, (t_{i},t_{i-1}).
\end{equation}

Then, the two intermediate hidden states $\boldsymbol{\overrightarrow{h'}}_i$ and $\boldsymbol{\overleftarrow{h'}}_i$ are merged to form the full intermediate state, $\boldsymbol{h'_i}=\boldsymbol{\overrightarrow{h'}}_i \oplus \boldsymbol{\overleftarrow{h'}}_i$. Hence, the RNN cell update is: 

\begin{equation}
\label{eq:CODERNNupdate}
\boldsymbol{h}_i = RNNCell(\boldsymbol{h}'_{i}, \boldsymbol{x}_i) \,\,\, \text{with} \,\,\, \boldsymbol{h'}_i = \boldsymbol{\overrightarrow{h}}_i \oplus \boldsymbol{\overleftarrow{h}}_i.
\end{equation}

\begin{figure}[h]
    \centering
    \includegraphics[width=\linewidth]{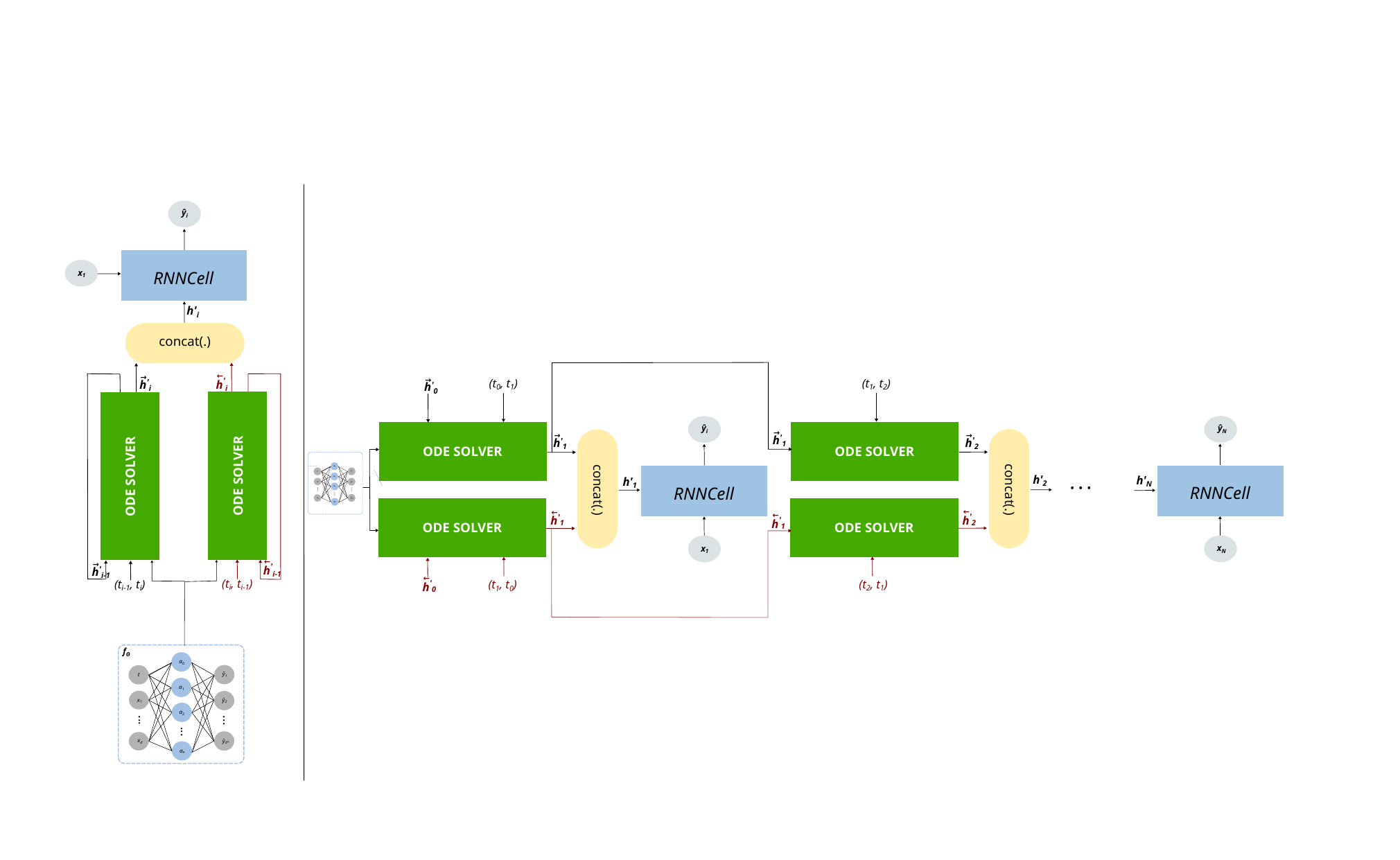}
    \caption{Training scheme of CODE-RNN and CODE-RNN unfolded through time (on the right). It uses a Neural CODE to adjust an ODE dynamics $\boldsymbol{f_\theta}$.}
\label{fig:TBANODERNN}
\end{figure}

The training process of CODE-RNN is described in Algorithm \ref{alg:TBAODERNN}. 
It is important to note that in CODE-RNN, the initial and final conditions of both the IVP and FVP are determined by the forward $\boldsymbol{\overrightarrow{h'}}_i$ and backward $\boldsymbol{\overleftarrow{h'}}_i$ intermediate hidden states. This is in contrast to BiRNNs, where the hidden states that propagate from one time step to another also incorporate the input values $\boldsymbol{x}_i$. Figure \ref{fig:TBANODERNN} depicts the architecture of CODE-RNN.

To optimise the parameters $\boldsymbol{\theta}$ of CODE-RNN, the loss function is defined by MSE:

\begin{equation}
\label{eq:L_CODERNN}
    \mathcal{L}(\boldsymbol{\theta}) = MSE(\boldsymbol{\hat{Y}}, \boldsymbol{Y}).
\end{equation}

This is a standard loss function commonly used in neural networks, as opposed to the loss function \eqref{eq:L_total} specifically designed for Neural CODE.

\begin{algorithm}[h]
\caption{CODE-RNN training process.}
\label{alg:TBAODERNN}
\begin{algorithmic}
\State \textbf{Input:} Data points and their timestamps $\{(\boldsymbol{x}_i,t_i)\}_{i=1, \dots, N}$, maximum number of iterations $MAXITER$;
\State $\boldsymbol{\overrightarrow{h}}_0\leftarrow\textbf{0}$, $\boldsymbol{\overleftarrow{h}}_0\leftarrow\textbf{0}$;
\State Choose $Optimiser$;
\State Initialise $\boldsymbol{\theta}$;
\For{$k=1:MAXITER$}
    \For{$i = 1:N$ }
        \State $\boldsymbol{\overrightarrow{h'}}_i\leftarrow ODESolve(\boldsymbol{f_\theta}, \boldsymbol{\overrightarrow{h'}}_{i-1}, (t_{i-1},t_i))$;
        \State $\boldsymbol{\overleftarrow{h'}}_{i} \leftarrow ODESolve(\boldsymbol{f_\theta}, \boldsymbol{\overleftarrow{h'}}_{i-1}, (t_{i},t_{i-1}))$;
        \State $\boldsymbol{h'}_i \leftarrow \boldsymbol{\overrightarrow{h'}}_i \oplus \boldsymbol{\overleftarrow{h'}}_{i}$;
        \State $\boldsymbol{h}_i \leftarrow RNNCell(\boldsymbol{h'}_i,\boldsymbol{x}_i)$;
    \EndFor
    \State $\boldsymbol{\hat{Y}} \leftarrow \boldsymbol{h}_i$ for all $i=1, \dots, N$;
    \State Evaluate loss $\mathcal{L}$ using \eqref{eq:L_CODERNN};
    \State $\nabla \mathcal{L} \leftarrow \text{Compute gradients of }\mathcal{L}$;
    \State $\boldsymbol{\theta} \leftarrow Optimiser.Step(\nabla \mathcal{L})$;
\EndFor
\State \textbf{Return:} $\boldsymbol{\theta}$;
\end{algorithmic}
\end{algorithm}



To make predictions in both forward and backward directions using CODE-RNN, the IVP and FVP are solved to construct the intermediate hidden state $\boldsymbol{h'}_i$, which is used to compute the output of the RNN update cell. The process of making predictions forward and backward in time follows a similar procedure, with the only difference being the order of the time steps $t_i$.
When making predictions in the forward direction, the time steps are given starting at the pairs $(t_{i-1}, t_i)$ for the IVP \eqref{eq:IVPCODE-RNN} and $(t_i, t_{i-1})$ for the FVP \eqref{eq:FVPCODE-RNN}, for $i=1,\dots,M$. Algorithm \ref{alg:TBAODERNN_test} describes the process of making future predictions with CODE-RNN. However, when making predictions in the backward direction, the time steps are given starting at the pairs $(t_{i-1}, t_i)$ for the IVP \eqref{eq:IVPCODE-RNN}, and $(t_i, t_{i-1})$ for the FVP \eqref{eq:FVPCODE-RNN}, for $i=N,\dots,m$. Algorithm \ref{alg:TBAODERNN_test2} describes the process of making past predictions with CODE-RNN. Similar to Neural CODE, predictions can be computed within ($t \in (t_0, t_f)$) or outside the training time interval for the past ($t \in (t_m, t_f))$ or the future ($t \in (t_0, t_M)$).

Note that, once more we consider that the hidden state $\boldsymbol{h}_i$ outputted by the RNN cell is the prediction $\boldsymbol{\hat{y}}_i$. 

\begin{algorithm}[h]
\caption{CODE-RNN for making future predictions.}
\label{alg:TBAODERNN_test}
\begin{algorithmic}
\State \textbf{Input:} start time $t_0$, end time $t_M$; 
\State $\boldsymbol{\overrightarrow{h}}_0\leftarrow\textbf{0}$, $\boldsymbol{\overleftarrow{h}}_0\leftarrow\textbf{0}$;
\For{$i=1:M$}
    \State $\boldsymbol{\overrightarrow{h'}}_i\leftarrow ODESolve(\boldsymbol{f_\theta}, \boldsymbol{\overrightarrow{h'}}_{i-1}, (t_{i-1},t_i))$;
    \State $\boldsymbol{\overleftarrow{h'}}_{i} \leftarrow ODESolve(\boldsymbol{f_\theta}, \boldsymbol{\overleftarrow{h'}}_{i-1}, (t_{i},t_{i-1}))$;
    \State $\boldsymbol{h'}_i \leftarrow \boldsymbol{\overrightarrow{h'}}_i \oplus \boldsymbol{\overleftarrow{h'}}_{i}$;
    \State $\boldsymbol{h}_i \leftarrow RNNCell(\boldsymbol{h'}_i,\boldsymbol{x}_i)$;
\EndFor
\State  $\overrightarrow{\boldsymbol{\hat{Y}}} \leftarrow \boldsymbol{h}_i$ for all $i=1, \dots, M$;
\State \textbf{Return:} $\,\, \overrightarrow{\boldsymbol{\hat{Y}}}$;
\end{algorithmic}
\end{algorithm}

\begin{algorithm}[h]
\caption{CODE-RNN for making past predictions.}
\label{alg:TBAODERNN_test2}
\begin{algorithmic}
\State \textbf{Input:} start time $t_N$, end time $t_m$;  
\State $\boldsymbol{\overrightarrow{h}}_{N-1}\leftarrow\textbf{0}$, $\boldsymbol{\overleftarrow{h}}_{N-1}\leftarrow\textbf{0}$;
\For{$i=N:m$}
    \State $\boldsymbol{\overrightarrow{h'}}_i\leftarrow ODESolve(\boldsymbol{f_\theta}, \boldsymbol{\overrightarrow{h'}}_{i-1}, (t_{i-1},t_i))$;
    \State $\boldsymbol{\overleftarrow{h'}}_{i} \leftarrow ODESolve(\boldsymbol{f_\theta}, \boldsymbol{\overleftarrow{h'}}_{i-1}, (t_{i},t_{i-1}))$;
    \State $\boldsymbol{h'}_i \leftarrow \boldsymbol{\overrightarrow{h'}}_i \oplus \boldsymbol{\overleftarrow{h'}}_{i}$;
    \State $\boldsymbol{h}_i \leftarrow RNNCell(\boldsymbol{h'}_i,\boldsymbol{x}_i)$;
\EndFor
\State  $\overleftarrow{\boldsymbol{\hat{Y}}} \leftarrow \boldsymbol{h}_i$ for all $i=N, \dots, m$;
\State \textbf{Return:} $\overleftarrow{\boldsymbol{\hat{Y}}}$;
\end{algorithmic}
\end{algorithm}

In the literature, several variants of update cells have emerged to enhance the performance of the RNN cell. The recurrent architectures based on Neural CODE can incorporate any of these variants. In this study, we conducted experiments with two additional update cells: GRU (CODE-GRU) and LSTM (CODE-LSTM).

In contrast to ODE-RNNs, the hidden state $\boldsymbol{h}_i$ resulting from the RNN cell update is not transmitted to the next cell through the feedback loop. Our approach uses intermediate hidden states in the forward ($\boldsymbol{\overrightarrow{h'}}_i$) and backward ($\boldsymbol{\overleftarrow{h'}}_{i}$) directions. This is necessary because the initial and final conditions for solving the ODE cannot be the same, which would occur if we used the hidden state $\boldsymbol{h}_i$ in the feedback loop. This limitation reveals a significant drawback of CODE-RNN/-GRU/-LSTM, as the input information $\boldsymbol{x}_i$ is not considered or passed to the next time step iteration; it is only used to compute the output prediction. To address this issue, we propose an upgraded architecture called CODE-BiRNN.

\subsection{CODE-BiRNN}


We introduce CODE-BiRNN, an enhanced bidirectional recurrent architecture based on Neural CODE, which is an improvement over CODE-RNN. The aim of this enhancement is to incorporate the input information $\boldsymbol{x}_i$ into the hidden state of the feedback loop, thus updating the ODE dynamics at each observation.

To accomplish this, we made modifications to the architecture. Instead of having a single cell update that generates a single hidden state, $\boldsymbol{h}_i$, we redesigned the structure to include two independent cell updates. The first update computes the forward hidden states, $\boldsymbol{\overrightarrow{h}}_i$, while the second update computes the backward hidden states, $\boldsymbol{\overleftarrow{h}}_{i}$. 

The forward intermediate hidden state $\boldsymbol{\overrightarrow{h'}}_i$ is given by solving the IVP \eqref{eq:IVPCODE-BiRNN} within the time interval between the previous and current time steps, $(t_{i-1}, t_i)$:

\begin{equation}
\label{eq:IVPCODE-BiRNN}
    \dfrac{d \boldsymbol{\overrightarrow{h'}}(t)}{dt} = \boldsymbol{f_\theta} (\boldsymbol{\overrightarrow{h}}(t),t) \,\,\, \text{with} \,\,\, \boldsymbol{\overrightarrow{h}}(t_{i-1}) = \boldsymbol{\overrightarrow{h}}_{i-1}.
\end{equation}

The backward intermediate hidden state, $\boldsymbol{\overleftarrow{h'}}_i$, is obtained by solving the FVP \eqref{eq:FVPCODE-BiRNN} within the time interval between the current and previous time steps, $(t_i, t_{i-1})$:

\begin{equation}
\label{eq:FVPCODE-BiRNN}
    \dfrac{d \boldsymbol{\overleftarrow{h'}}(t)}{dt} = \boldsymbol{f_\theta} (\boldsymbol{\overleftarrow{h}}(t),t) \,\,\, \text{with} \,\,\, \boldsymbol{\overleftarrow{h}}(t_{i}) = \boldsymbol{\overleftarrow{h}}_{i}.
\end{equation}

Thus, unlike CODE-RNN, the forward, $\boldsymbol{\overrightarrow{h'}}_i$, and backward, $\boldsymbol{\overleftarrow{h'}}_i$, intermediate hidden states are computed using the previous forward $\boldsymbol{\overrightarrow{h}}_{i-1}$  \eqref{eq:TBAODEBiRNN1} and backward $\boldsymbol{\overleftarrow{h}}_{i-1}$  \eqref{eq:TBAODEBiRNN2} hidden states as initial and final value, respectively.

\begin{equation}
\label{eq:TBAODEBiRNN1}    
\boldsymbol{\overrightarrow{h'}}_i = ODESolve(\boldsymbol{f_\theta}, \boldsymbol{\overrightarrow{h}}_{i-1}, (t_{i-1},t_i))
\end{equation}

\begin{equation}
\label{eq:TBAODEBiRNN2}    
\boldsymbol{\overleftarrow{h'}}_{i} = ODESolve(\boldsymbol{f_\theta}, \boldsymbol{\overleftarrow{h}}_{i-1}, (t_{i},t_{i-1}))
\end{equation}

The two intermediate hidden states $\boldsymbol{\overrightarrow{h'}}_i$ and $\boldsymbol{\overleftarrow{h'}}_i$, are then passed onto separate RNN cells which output the forward $\boldsymbol{\overrightarrow{h}}_i$ \eqref{eq:CODEBiRNNupdateF} and backward $\boldsymbol{\overleftarrow{h}}_i$ \eqref{eq:CODEBiRNNupdateB} hidden states:

\begin{equation}
\label{eq:CODEBiRNNupdateF}
\boldsymbol{\overrightarrow{h}}_i = RNNCell(\boldsymbol{\overrightarrow{h'}}_{i}, \boldsymbol{x}_i),
\end{equation}

\begin{equation}
\label{eq:CODEBiRNNupdateB}
\boldsymbol{\overleftarrow{h}}_i = RNNCell(\boldsymbol{\overleftarrow{h'}}_{i}, \boldsymbol{x}_i),
\end{equation}

To get the predicted output at a given time step, $\boldsymbol{\hat{y}}_i$, the full hidden state, $\boldsymbol{h}_i$, is computed by performing an aggregation/merging operation of the forward $\boldsymbol{\overrightarrow{h}}_i$ and backward $\boldsymbol{\overleftarrow{h}}_i$ hidden states of the same time step $t_i$:

\begin{equation*}
    \boldsymbol{\hat{y}}_i = \sigma(\boldsymbol{W}^{\text{[out]}} \boldsymbol{h}_i + \boldsymbol{b}^{\text{[out]}}) \,\,\, \text{with} \,\,\, \boldsymbol{h}_i = \boldsymbol{\overrightarrow{h}}_i \oplus \boldsymbol{\overleftarrow{h}}_i.
\end{equation*}

By incorporating two separate hidden layers within the RNN, processing the data in opposite temporal directions, our network inherits the name BiRNN. Algorithm \ref{alg:BiODE-RNN} outlines the implementation details. 

\begin{algorithm}[h]
\caption{CODE-BiRNN training process.}
\label{alg:BiODE-RNN}
\begin{algorithmic}
\State \textbf{Input:} Data points and their timestamps $\{(\boldsymbol{x}_i,t_i)\}_{i=1, \dots, N}$, maximum number of iterations $MAXITER$;
\State $\boldsymbol{\overrightarrow{h}}_0\leftarrow\textbf{0}$, $\boldsymbol{\overleftarrow{h}}_0\leftarrow\textbf{0}$;
\State Choose $Optimiser$;
\State Initialise $\boldsymbol{\theta}$;
\For{$k=1:MAXITER$}
\For{$i=1:N$}
    \State $\boldsymbol{\overrightarrow{h'}}_i \leftarrow ODESolve(\boldsymbol{f_\theta}, \boldsymbol{\overrightarrow{h}}_{i-1}, (t_{i-1},t_i))$ ;
    \State $\boldsymbol{\overrightarrow{h}}_i \leftarrow RNNCell(\boldsymbol{\overrightarrow{h'}}_i,\boldsymbol{x}_i)$;
\EndFor
\For{$i= 1, \dots, N$ }
    \State $\boldsymbol{\overleftarrow{h'}}_{i} \leftarrow ODESolve(\boldsymbol{f_\theta}, \boldsymbol{\overleftarrow{h}}_{i-1}, (t_{i},t_{i-1}))$; 
    \State $\boldsymbol{\overleftarrow{h}}_{i} \leftarrow RNNCell(\boldsymbol{\overleftarrow{h'}}_{i},\boldsymbol{x}_i)$;
\EndFor
\State $\boldsymbol{h}_i \leftarrow \boldsymbol{\overrightarrow{h}}_i \oplus \boldsymbol{\overleftarrow{h}}_{i}$;
\State $\boldsymbol{\hat{Y}} \leftarrow \boldsymbol{h}_i$ for all $i=1, \dots, N$;
\State Evaluate loss $\mathcal{L}$ using \eqref{eq:L_CODERNN};
\State $\nabla \mathcal{L} \leftarrow \text{Compute gradients of }\mathcal{L}$;
\State $\boldsymbol{\theta} \leftarrow Optimiser.Step(\nabla \mathcal{L})$;
\EndFor
\State \textbf{Return:} $\boldsymbol{\theta}$;
\end{algorithmic}
\end{algorithm}

\begin{figure}[h]
    \centering
    \includegraphics[width=\textwidth]{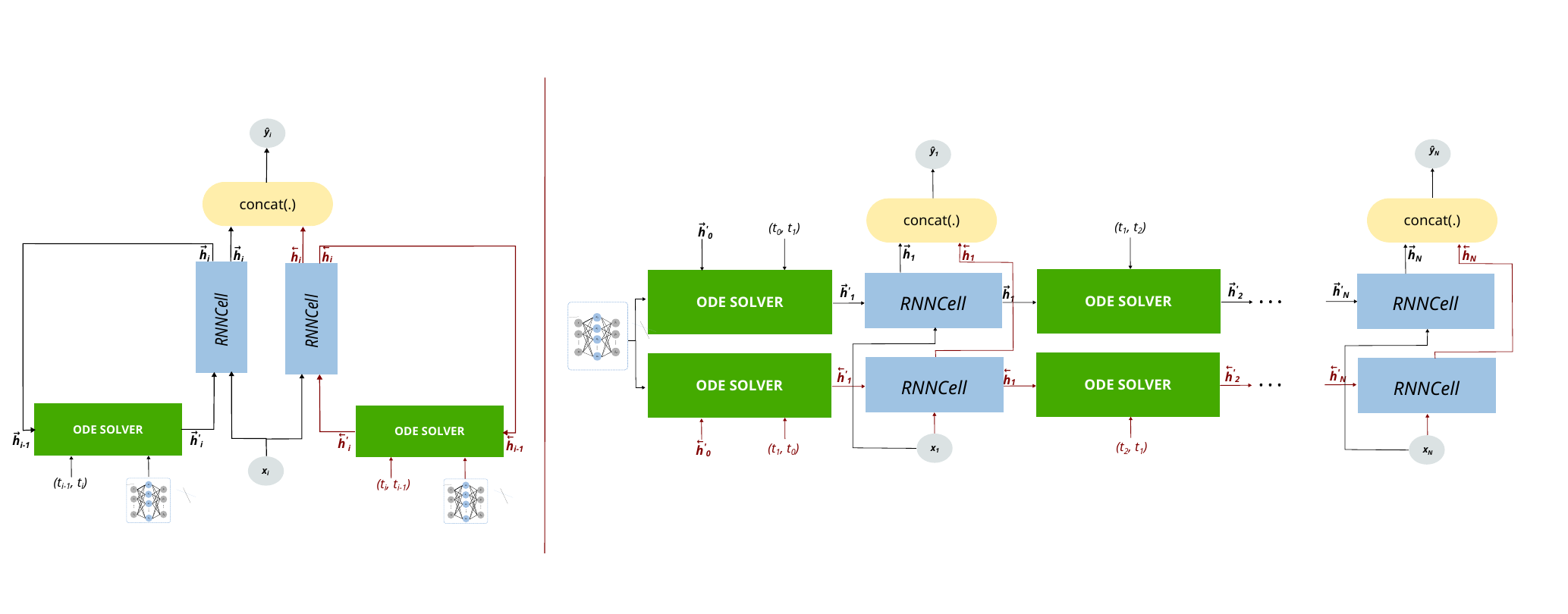}
    \caption{Scheme of CODE-BiRNN and CODE-BiRNN unfolded through time.}
    \label{fig:TBANODEBiRNN}
\end{figure}

Replacement of the RNN by a BiRNN makes it possible to have the full independent recurrent process for each time direction, Figure \ref{fig:TBANODEBiRNN}.

Similar to CODE-RNN, in CODE-BiRNN, the optimisation process aims to find the optimal parameters $\boldsymbol{\theta}$ by minimising the MSE between the predicted values and the true values. 

To make predictions in both forward and backward directions using CODE-BiRNN, a similar approach is followed as in CODE-RNN. The IVP \eqref{eq:IVPCODE-BiRNN} and FVP \eqref{eq:FVPCODE-BiRNN} need to be solved to construct the forward and backward outputs of the RNN cells, respectively, which are then used to generate the final hidden state by combining the contributions from both directions.
The process of making predictions in forward and backward directions with CODE-BiRNN is analogous to that of CODE-RNN, with the directional difference lying in the order of the time steps. When making predictions in the forward direction, the time steps are given in the pairs $(t_{i-1}, t_i)$ for the IVP and $(t_i, t_{i-1})$ for the FVP, for $i = 1,\dots,M$. Algorithm \ref{alg:BiODE-RNN_test} describes the process of making future predictions with CODE-RNN. On the other hand, when making predictions in the backward direction, the time steps are given in the pairs $(t_{i-1}, t_i)$ for the IVP, and $(t_i, t_{i-1})$ for the FVP, , for $i = N,\dots,m$. Algorithm \ref{alg:BiODE-RNN_test2} describes the process of making future predictions with CODE-RNN. Predictions can be computed within ($t \in (t_0, t_f)$) or outside the training time interval for the past ($t \in (t_m, t_f))$ or the future ($t \in (t_0, t_M)$).

Note that, once more we consider that the hidden state $\boldsymbol{h}_i$ that results of aggregating/merging the forward and backward outputs of the RNN cells is the prediction $\boldsymbol{\hat{y}}_i$.

\begin{algorithm}[h]
\caption{CODE-BiRNN for making future predictions.}
\label{alg:BiODE-RNN_test}
\begin{algorithmic}
\State \textbf{Input:} start time $t_0$, end time $t_M$;
\State $\boldsymbol{\overrightarrow{h}}_0\leftarrow\textbf{0}$, $\boldsymbol{\overleftarrow{h}}_0\leftarrow\textbf{0}$;
\For{ $i=1:M$}
    \State $\boldsymbol{\overrightarrow{h'}}_i \leftarrow ODESolve(\boldsymbol{f_\theta}, \boldsymbol{\overrightarrow{h}}_{i-1}, (t_{i-1},t_i))$ ;
    \State $\boldsymbol{\overrightarrow{h}}_i \leftarrow RNNCell(\boldsymbol{\overrightarrow{h'}}_i,\boldsymbol{x}_i)$;
\EndFor
\For{ $i$ in $1:M$}
    \State $\boldsymbol{\overleftarrow{h'}}_{i} \leftarrow ODESolve(\boldsymbol{f_\theta}, \boldsymbol{\overleftarrow{h}}_{i-1}, (t_{i},t_{i-1}))$; 
    \State $\boldsymbol{\overleftarrow{h}}_{i} \leftarrow RNNCell(\boldsymbol{\overleftarrow{h'}}_{i},\boldsymbol{x}_i)$;
\EndFor
\State $\boldsymbol{h}_i \leftarrow \boldsymbol{\overrightarrow{h}}_i \oplus \boldsymbol{\overleftarrow{h}}_{i}$;
\State  $\overrightarrow{\boldsymbol{\hat{Y}}} \leftarrow \boldsymbol{h}_i$ for all $i=1, \dots, M$;
\State \textbf{Return:} $\overrightarrow{\boldsymbol{\hat{Y}}}$;
\end{algorithmic}
\end{algorithm}

\begin{algorithm}[h]
\caption{CODE-BiRNN for making past predictions.}
\label{alg:BiODE-RNN_test2}
\begin{algorithmic}
\State \textbf{Input:} start time $t_N$, end time $t_m$;
\State $\boldsymbol{\overrightarrow{h}}_{N-1}\leftarrow\textbf{0}$, $\boldsymbol{\overleftarrow{h}}_{N-1}\leftarrow\textbf{0}$;
\For{$i=N:m$}
    \State $\boldsymbol{\overrightarrow{h'}}_i \leftarrow ODESolve(\boldsymbol{f_\theta}, \boldsymbol{\overrightarrow{h}}_{i-1}, (t_{i-1},t_i))$ ;
    \State $\boldsymbol{\overrightarrow{h}}_i \leftarrow RNNCell(\boldsymbol{\overrightarrow{h'}}_i,\boldsymbol{x}_i)$;
\EndFor
\For{$i=N:m$}
    \State $\boldsymbol{\overleftarrow{h'}}_{i} \leftarrow ODESolve(\boldsymbol{f_\theta}, \boldsymbol{\overleftarrow{h}}_{i-1}, (t_{i},t_{i-1}))$; 
    \State $\boldsymbol{\overleftarrow{h}}_{i} \leftarrow RNNCell(\boldsymbol{\overleftarrow{h'}}_{i},\boldsymbol{x}_i)$;
\EndFor
\State $\boldsymbol{h}_i \leftarrow \boldsymbol{\overrightarrow{h}}_i \oplus \boldsymbol{\overleftarrow{h}}_{i}$;
\State  $\overleftarrow{\boldsymbol{\hat{Y}}} \leftarrow \boldsymbol{h}_i$ for all $i=N, \dots, m$;
\State \textbf{Return:} $\overleftarrow{\boldsymbol{\hat{Y}}}$;
\end{algorithmic}
\end{algorithm}

The bidirectional recurrent architecture based on Neural CODE can use any update cell. In this work we present results with RNN (CODE-BiRNN), GRU (CODE-BiGRU) and LSTM (CODE-BiLSTM) cells.

\section{Numerical Experiments} \label{sec:experiments}

In order to evaluate the performance of the architectures presented in this work, a set of experiments were conducted using various datasets and tasks. Specifically, we compared Neural CODE to Neural ODE by reconstructing spiral ODE dynamics, both forward and backward in time, using two synthetic datasets that differ in the number of training and testing points. By varying the number of data points, we aimed to investigate how Neural CODE performs in scenarios where data availability is limited, which is often encountered in real-world applications. 

Then, CODE-RNN/-GRU/-LSTM, CODE-BiRNN/-BiGRU/-BiLSTM were tested and ODE-RNN/-GRU/-LSTM were used as baselines. These comparison were carried out using three real-world time-series datasets: regularly-sampled data from the climate domain, sparsely regularly-sampled data from the hydrological domain, and irregularly-sampled data from the stock market domain. The evaluation covered three distinct tasks: missing data imputation, future extrapolation, and backward extrapolation (past discovery). 
The first task, missing data imputation, aimed to assess the models' ability to fill in missing values in the time-series data, being crucial when  dealing with incomplete datasets where the missing information needs to be estimated. The second task, future extrapolation, focused on predicting future values beyond the observed time range, being essential for forecasting and decision-making in many domains. The third task, past discovery, aimed to evaluate the models' capacity to uncover and reconstruct past patterns in the time-series data. This task is particularly relevant for analysing historical trends and understanding the dynamics of the data over time. 
Furthermore, in the context of the missing data imputation task, we conducted testing using two different settings: one involving a single input/output feature, and the other involving higher-dimensional input/output vectors. This allowed us to leverage the tests conducted on Neural CODE and examine its behaviour when learning from higher-dimensional data. By evaluating the model's performance in  capturing the dynamics of the time-series in higher-dimensional settings, we gained insights into its ability to handle and learn from datasets with increased complexity. In the context of future and backward extrapolation, the testing was conducted using sequences of 7 or 15 observations to predict 7 or 15 observations. This allowed us to study the performance of the proposed architectures when predicting for longer time-horizons.

\subsection{Case Study 1: Synthetic Spiral ODE Dynamics}

We compared the performance of Neural CODE and Neural ODE in restoring the dynamics of a spiral ODE both forward and backward in time. By comparing the performance of Neural CODE and Neural ODE on this specific task, we aimed to gain insights into the strengths and limitations of each model in terms of their ability to capture and reproduce complex dynamics. 
To create the training and testing sets, we defined a simple linear ODE \eqref{eq:expSpiral1} and solved it numerically to retrieve data points in the time interval [0, 25]. For each time step, values for the two coordinates of the spiral are available $(x, y)$. 

We adapted the code from the \textit{Python Torchdiffeq} \citep{chenTorchdiffeq2021} for our implementation:

\begin{equation}
\label{eq:expSpiral1}
\begin{cases}
    \dfrac{dx}{dt} = -0.1x^3 + 2.0y^3 \\
    \dfrac{dy}{dt} = -2.0x^3 - 0.1y^3
\end{cases}
\end{equation}

Two datasets were created: one with 2000 training points and 1000 testing points, denoted by $2000/1000$ dataset, and another with 1000 training points and 2000 testing points, denoted by $1000/2000$ dataset. We aimed to study the impact of dataset size on the models' performance.

For training, we used a batch size of 1, a sequence length of 10 time steps, 2000 iterations for MAXITER, and the Adam optimiser with a learning rate of 0.001. For the $ODESolve$, we employed the Runge-Kutta method of order 5 (Dormand-Prince-Shampine) with default configurations. The loss function of Neural CODE is based on the MSE (see \eqref{eq:L_total}).
The NN architecture used in this study consists of three layers. The input layer contains 2 neurons, representing the input features, $(x_i,y_i)$. The hidden layer is composed of 50 neurons, each using the hyperbolic tangent activation function. Finally, the output layer consists of 2 neurons, representing the output or prediction of the model, $(\hat{x}_i,\hat{y}_i)$.

\begin{figure}[h]
\centering
\begin{subfigure}{\textwidth}
  \centering
  \includegraphics[width=1\textwidth]{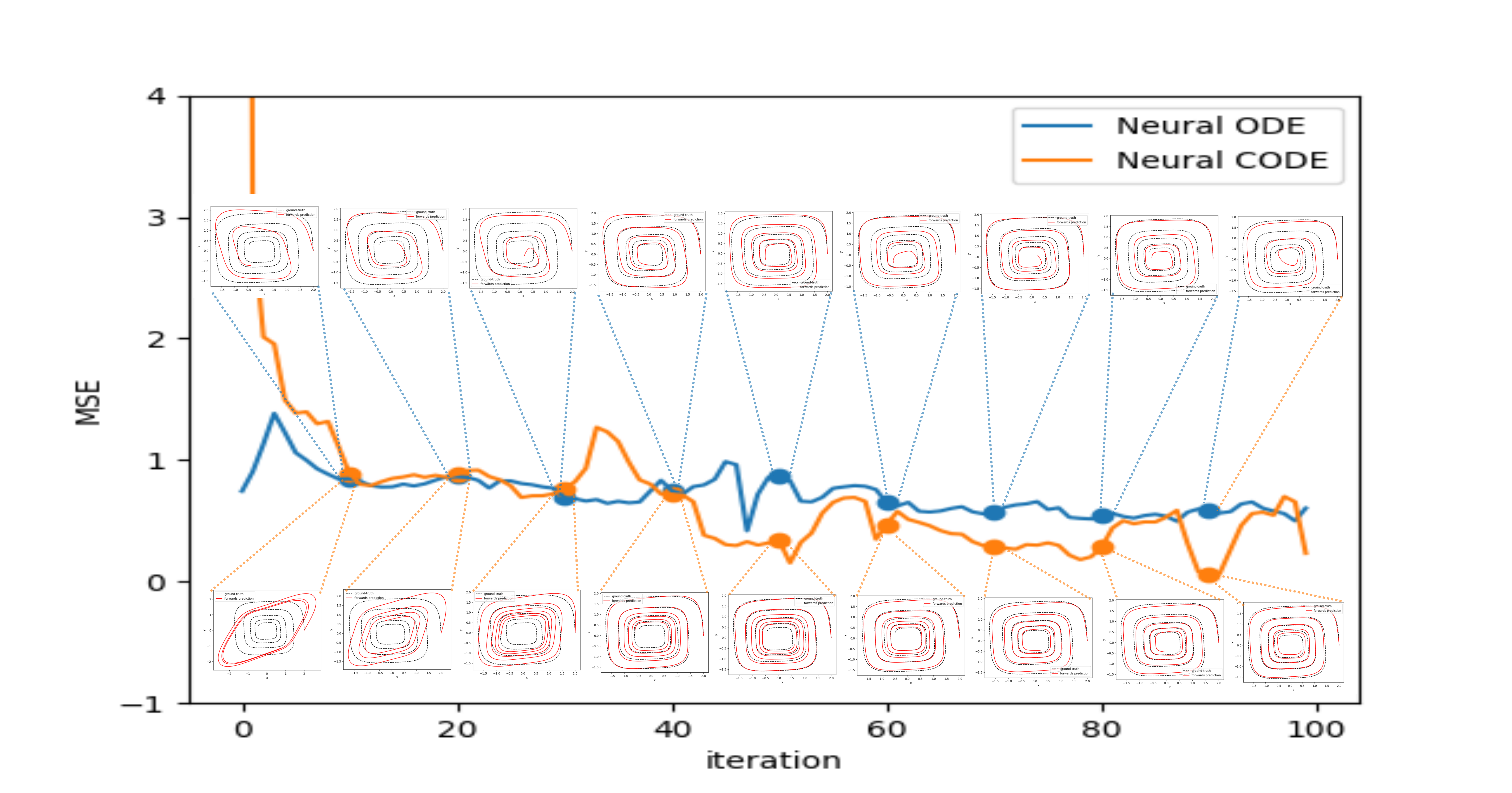}
  \caption{2000/1000 dataset.}
  \label{fig:TBAODE_spiralLoss1sub1}
\end{subfigure} \\
\begin{subfigure}{1\textwidth}
  \centering
  \includegraphics[width=\textwidth]{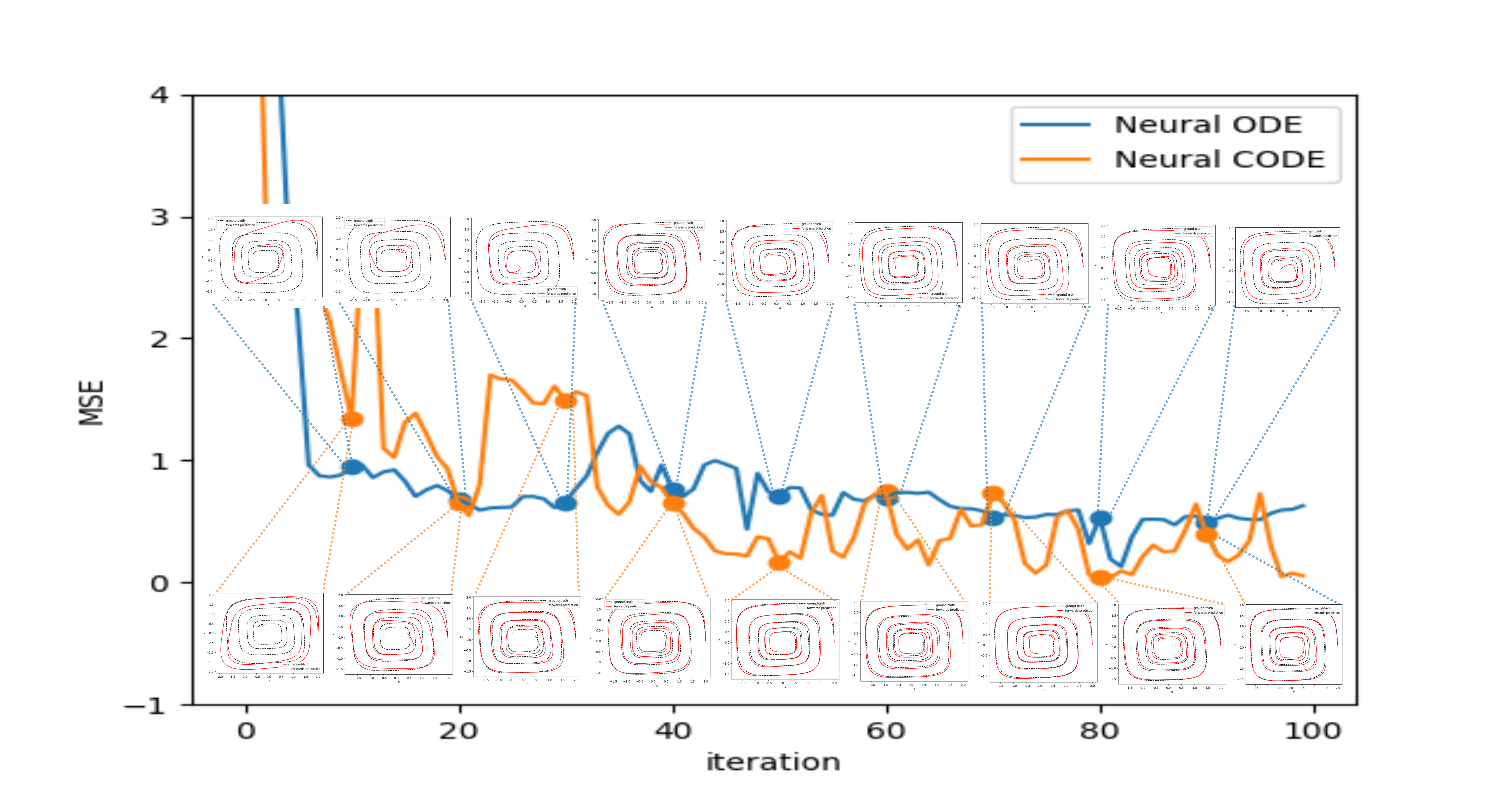}
  \caption{1000/2000 dataset.}
  \label{fig:TBAODE_spiralLoss1sub2}
\end{subfigure}
\caption{Training MSE values for forward predictions, with $20$ iterations frequency, with the respective forward learnt spiral dynamics , (a) $2000/1000$ dataset and (b) $1000/2000$ dataset, for Neural ODE (in blue) and Neural CODE (in orange).}
\label{fig:TBAODE_spiralLoss1}
\end{figure}

\begin{figure}[h]
\centering
\begin{subfigure}{\textwidth}
  \centering
  \includegraphics[width=1\linewidth]{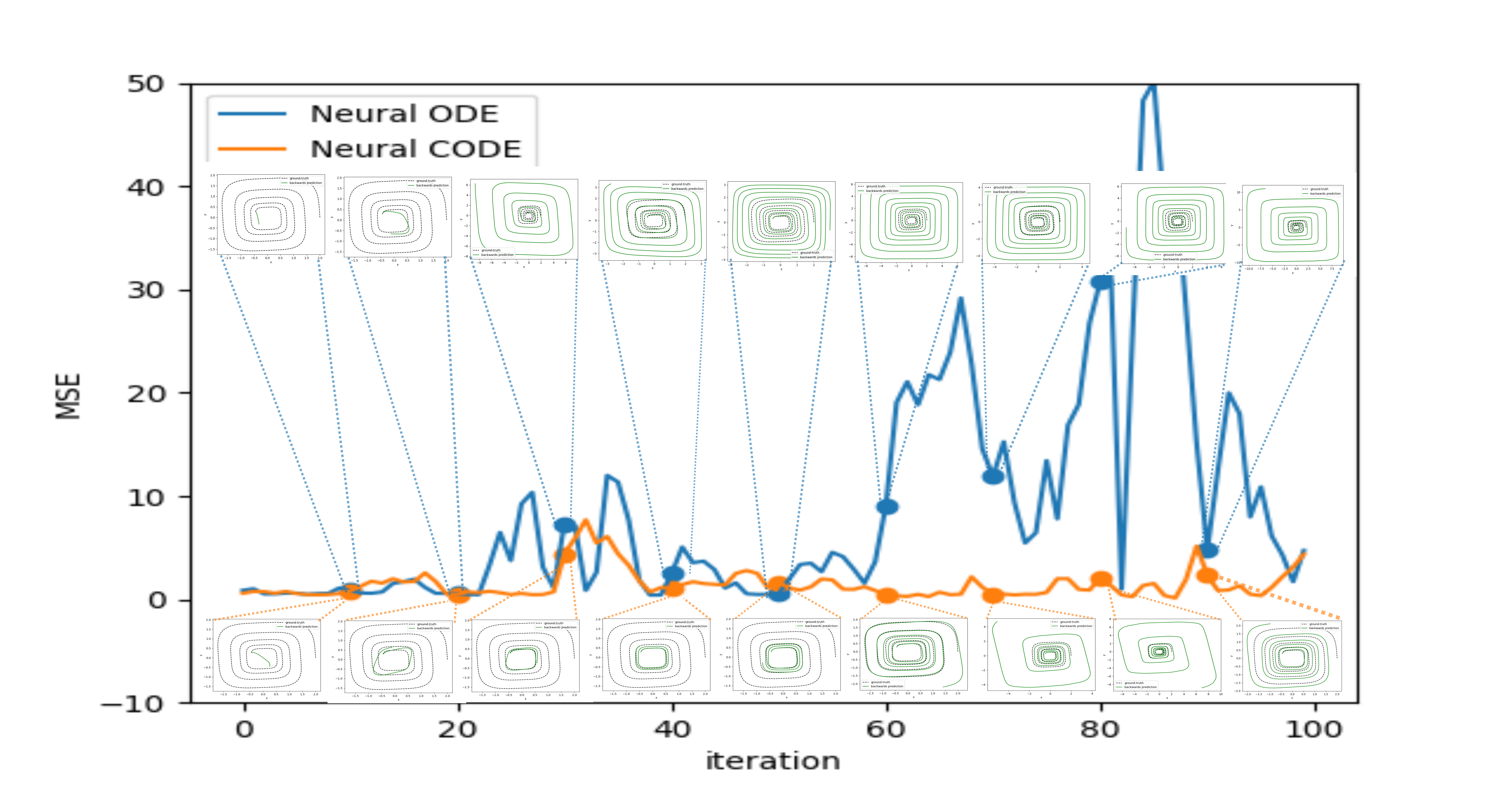}
  \caption{2000/1000 dataset.}
  \label{fig:sub1}
\end{subfigure} \\
\begin{subfigure}{\textwidth}
  \centering
  \includegraphics[width=1\linewidth]{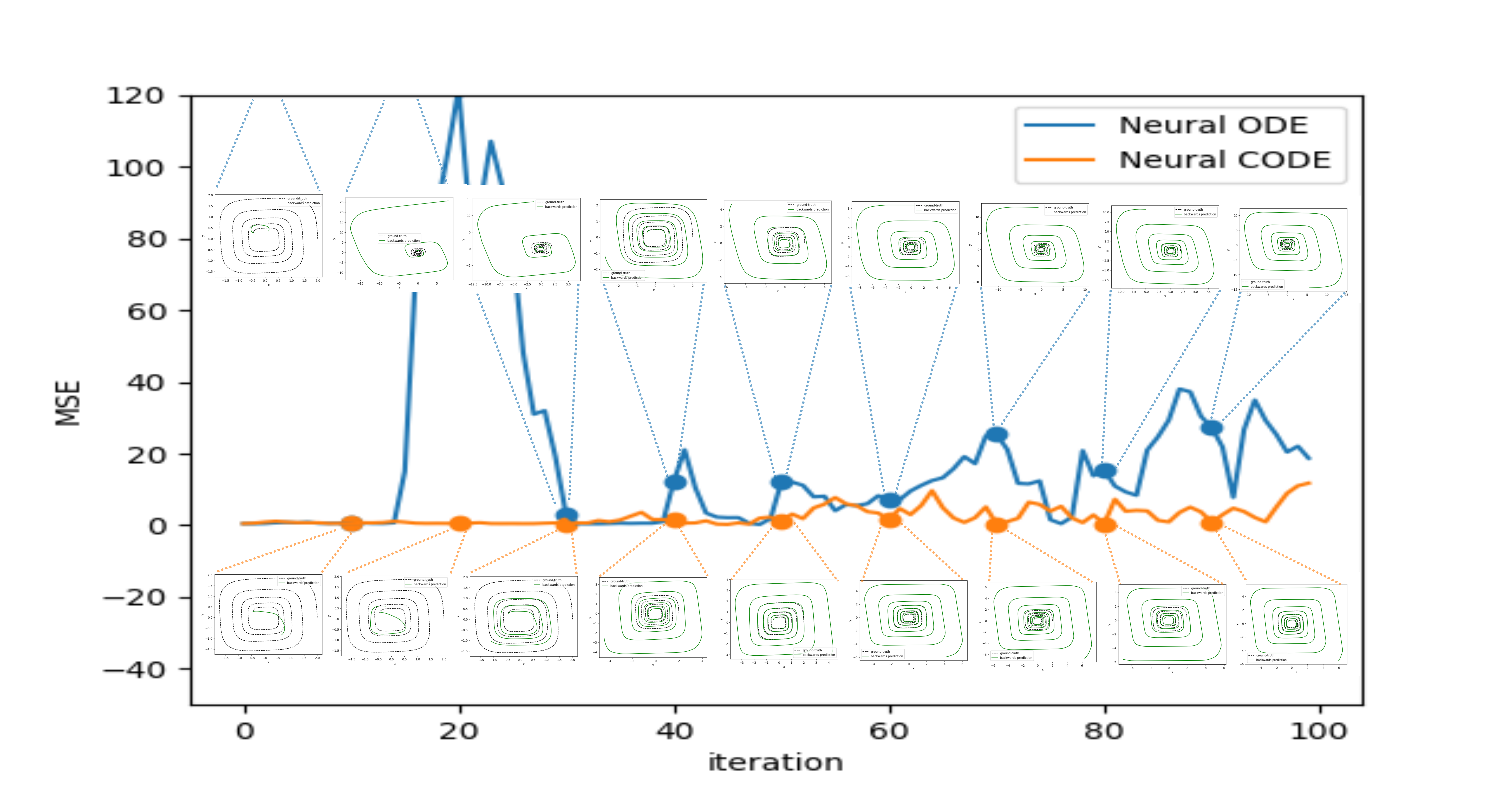}
  \caption{1000/2000 dataset.}
  \label{fig:sub2}
\end{subfigure}
\caption{Training MSE values for backward predictions, with $20$ iterations frequency, with the respective backward learnt spiral dynamics, (a) $2000/1000$ dataset and (b) $1000/2000$ dataset, for Neural ODE (in blue) and Neural CODE (in orange).}
\label{fig:TBAODE_spiralLoss2}
\end{figure}

To account for random weight initialisation, we trained and tested each model three times (R=3). To evaluate the performance of the models, we compute the average of the MSE ($MSE_{avg}$) and standard deviation ($std_{avg}$) values for the test sets, from the three runs. Additionally, to analyse the evolution of the adjusted ODE during training, we plotted the MSE for forward and backward prediction on the training set at every 20 iterations (Figures \ref{fig:TBAODE_spiralLoss1} and \ref{fig:TBAODE_spiralLoss2}, respectively).

\paragraph{Forward reconstruction}


Table \ref{tab:periodicResults1} presents the numerical results of the Neural ODE and Neural CODE for the forward reconstruction of the spiral. The first column displays the number of points in the training and testing sets, the second displays the time interval considered for the reconstruction task. The next two columns display, for each model, the $MSE_{avg}$ and $std_{avg}$ values. 

The results presented in Table \ref{tab:periodicResults1} demonstrate that Neural CODE outperforms Neural ODE in reconstructing the spiral through the solution of an IVP. It is worth noting that as the number of training points decreases, there is an expected increase in the prediction error.

\begin{table}[h]
\centering
\begin{tabular}{@{}cccc@{}}
\toprule
                        &              & Neural ODE               & Neural CODE                                 \\ \midrule
training/testing points & $(t_0, t_M)$ & $MSE_{avg}\pm std_{avg}$ & $MSE_{avg}\pm std_{avg}$                    \\ \midrule
2000/1000               & $(0, 25)$    & 4.75e-1$\pm$1.21e-1        & \textbf{1.56e-1$\pm$8.37e-2} \\
1000/2000               & $(0, 25)$    & 5.89e-1$\pm$4.48e-2        & \textbf{3.57e-1$\pm$2.29e-1} \\ \bottomrule
\end{tabular}
\caption{Numerical results on the test spiral sets for predictions forward in time by Neural ODE and Neural CODE, by solving IVP \eqref{eq:IVP}.}
\label{tab:periodicResults1}
\end{table}



In Figure \ref{fig:TBAODE_spiralLoss1}, the plots illustrating the MSE evolution during training demonstrate that Neural CODE achieved lower MSE values in the $2000$ iterations. Additionally, Neural CODE exhibited faster adaptation to spiral dynamics compared to Neural ODE. By examining the MSE curves for both datasets, depicted in Figures \ref{fig:TBAODE_spiralLoss1sub1} and \ref{fig:TBAODE_spiralLoss1sub2}, we can conclude that using a smaller training dataset leads to more unstable learning.

\paragraph{Backward reconstruction}



Table \ref{tab:periodicResults2} presents the numerical results of the Neural ODE and Neural CODE for the backward reconstruction of the spiral. The first column displays the
number of points in the training and testing sets, the second displays the time interval considered for the reconstruction task. The next two columns display, for each model, the
$MSE_{avg}$ and $std_{avg}$ values.

The results presented in Table \ref{tab:periodicResults2} demonstrate that when using the fitted ODEs to predict backward in time by solving a FVP \eqref{eq:FVP}, Neural CODE demonstrates superior performance compared to Neural ODE for both datasets: $2000/1000$ and $1000/2000$. In contrast to the forward reconstruction results presented in Table \ref{tab:periodicResults1}, Neural CODE performs better or at least similarly when trained on a smaller number of data points ($1000/2000$ dataset).

In Figure \ref{fig:TBAODE_spiralLoss2}, the evolution of MSE during training provides evidence of the increased difficulty faced by Neural ODE when solving the FVP. However, it exhibits a similar behaviour to Neural CODE after completing half of the total iterations when trained with fewer points.

This discrepancy can be attributed to the fact that Neural CODE minimises a function that accounts for backward prediction, while Neural ODE does not. The improved prediction capabilities of Neural CODE with the $1000/2000$ dataset can be attributed to achieving a better generalised representation due to the limited number of training points, reducing the likelihood of overfitting.

\begin{table}[h]
\centering
\begin{tabular}{@{}llll@{}}
\toprule
                        &              & Neural ODE               & Neural CODE                                \\ \midrule
training/testing points & $(t_f, t_m)$ & $MSE_{avg}\pm std_{avg}$ & $MSE_{avg}\pm std_{avg}$                   \\ \midrule
2000/1000               & $(25, 0)$    & 8.95$\pm$6.50          & \textbf{2.31$\pm$7.88e-1} \\
1000/2000               & $(25, 0)$    & 7.87$\pm$6.82          & \textbf{1.37$\pm$7.99e-1} \\ \bottomrule
\end{tabular}
\caption{Numerical results on the test spiral sets for predictions backward in time by Neural ODE and Neural CODE by solving FVP \eqref{eq:FVP}.}
\label{tab:periodicResults2}
\end{table}

\footnotetext{For doing predictions backward in time using a Neural ODE model, the time interval is given in the reverse order to the $ODESolve$.}

\begin{figure}[h]
\centering
\begin{subfigure}{0.3\textwidth}
  \centering
  \includegraphics[width=\linewidth]{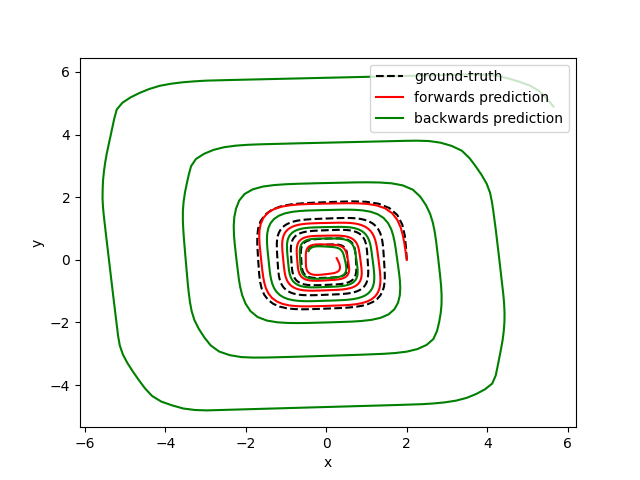}
  \caption{Spiral dynamics.}
  \label{fig:sub1}
\end{subfigure}%
\begin{subfigure}{.3\textwidth}
  \centering
  \includegraphics[width=\linewidth]{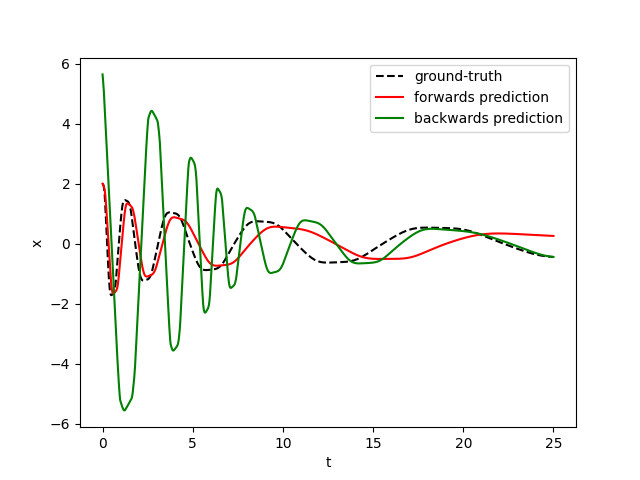}
  \caption{x values prediction.}
  \label{fig:sub2}
\end{subfigure}
\begin{subfigure}{.3\textwidth}
  \centering
  \includegraphics[width=\linewidth]{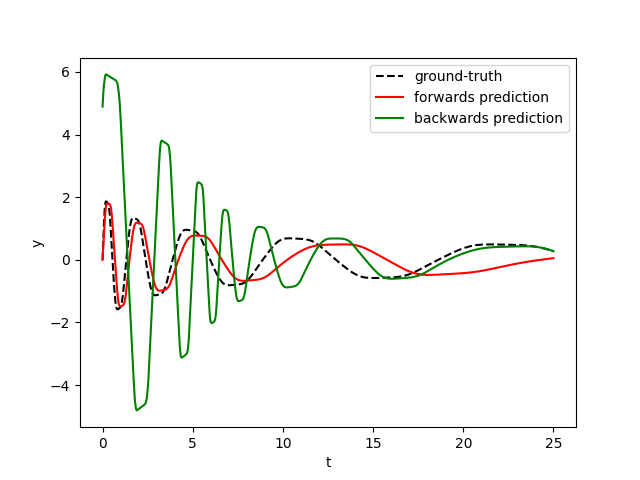}
  \caption{y values prediction.}
  \label{fig:sub2}
\end{subfigure}
\caption{Neural ODE at reconstructing the dynamics of a spiral ODE using the $2000/1000$ dataset. In dashed black the true dynamics, in red the predictions forward and in green the predictions backward in time.}
\label{fig:NODE_spiral}
\end{figure}

\begin{figure}[h]
\centering
\begin{subfigure}{0.3\textwidth}
  \centering
  \includegraphics[width=\linewidth]{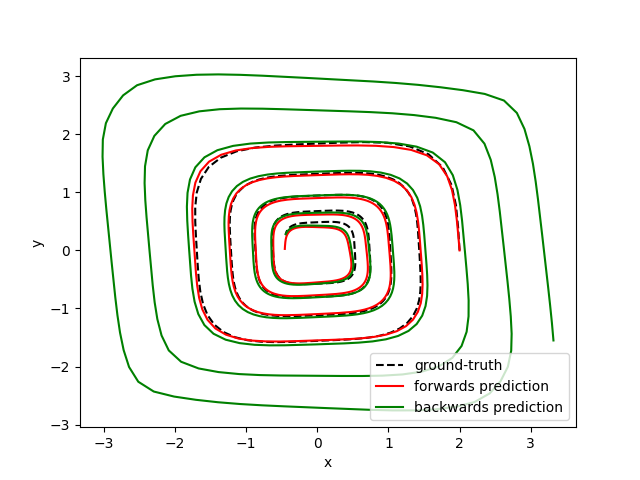}
  \caption{Spiral dynamics.}
  \label{fig:sub1}
\end{subfigure}%
\begin{subfigure}{.3\textwidth}
  \centering
  \includegraphics[width=\linewidth]{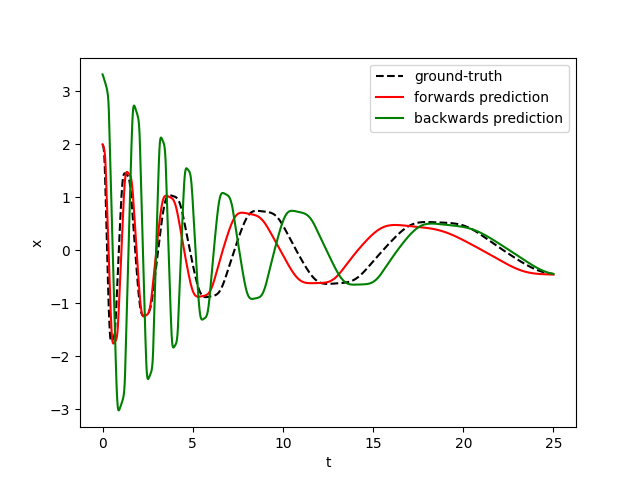}
  \caption{x values prediction.}
  \label{fig:sub2}
\end{subfigure}
\begin{subfigure}{.3\textwidth}
  \centering
  \includegraphics[width=\linewidth]{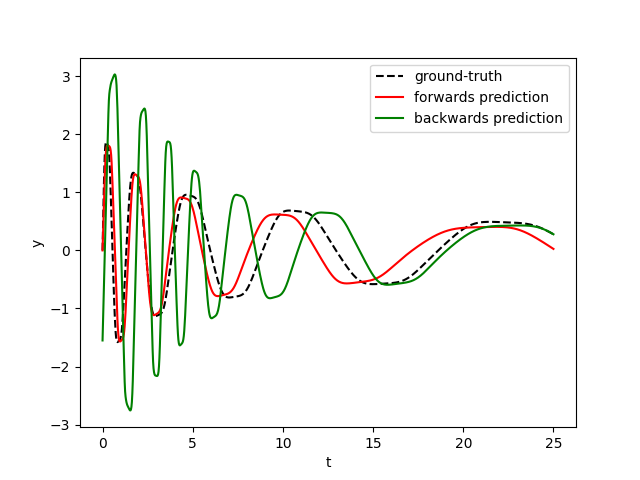}
  \caption{y values prediction.}
  \label{fig:sub2}
\end{subfigure}
\caption{Neural CODE at reconstructing the dynamics of a spiral ODE using the $2000/1000$ dataset. In dashed black the true dynamics, in red the predictions forward and in green the predictions backward in time.}
\label{fig:TBAODE_spiral}
\end{figure}

\begin{figure}[h]
\centering
\begin{subfigure}{0.3\textwidth}
  \centering
  \includegraphics[width=\linewidth]{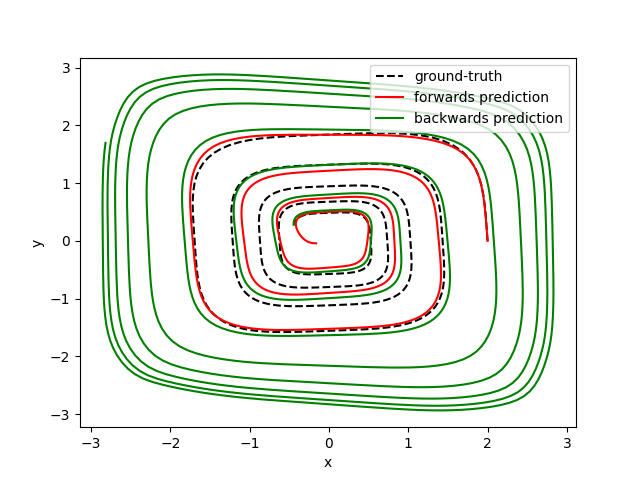}
  \caption{Spiral dynamics.}
  \label{fig:sub1}
\end{subfigure}%
\begin{subfigure}{.3\textwidth}
  \centering
  \includegraphics[width=\linewidth]{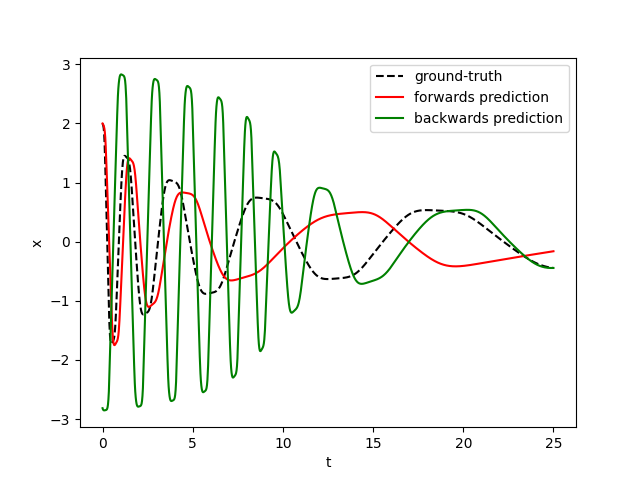}
  \caption{x values prediction.}
  \label{fig:sub2}
\end{subfigure}
\begin{subfigure}{.3\textwidth}
  \centering
  \includegraphics[width=\linewidth]{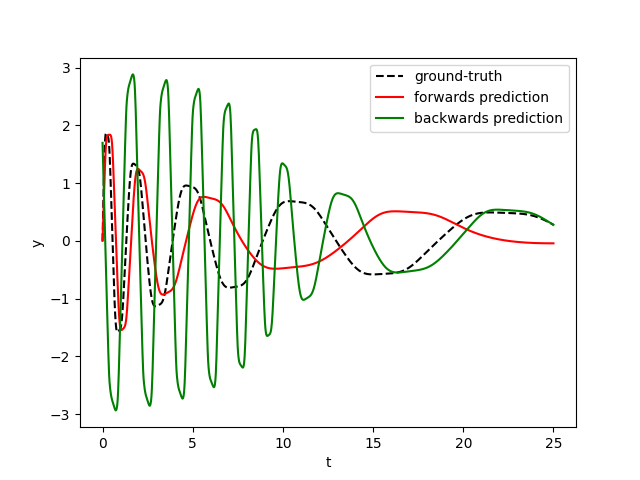}
  \caption{y values prediction.}
  \label{fig:sub2}
\end{subfigure}
\caption{Neural ODE at reconstructing the dynamics of a spiral ODE using the $1000/2000$ dataset. In dashed black the true dynamics, in red the predictions forward and in green the predictions backward in time.}
\label{fig:NODE_spiral2}
\end{figure}

\begin{figure}[h]
\centering
\begin{subfigure}{0.3\textwidth}
  \centering
  \includegraphics[width=\linewidth]{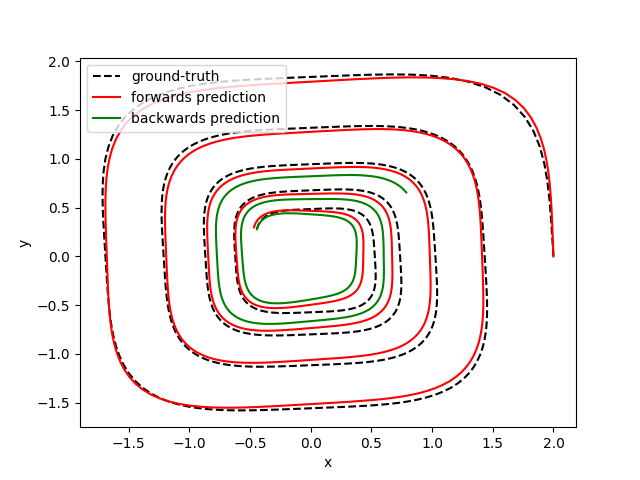}
  \caption{Spiral dynamics.}
  \label{fig:sub1}
\end{subfigure}%
\begin{subfigure}{.3\textwidth}
  \centering
  \includegraphics[width=\linewidth]{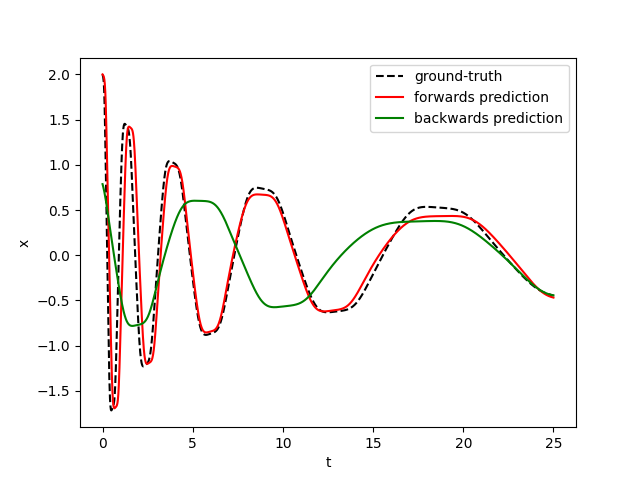}
  \caption{x values prediction.}
  \label{fig:sub2}
\end{subfigure}
\begin{subfigure}{.3\textwidth}
  \centering
  \includegraphics[width=\linewidth]{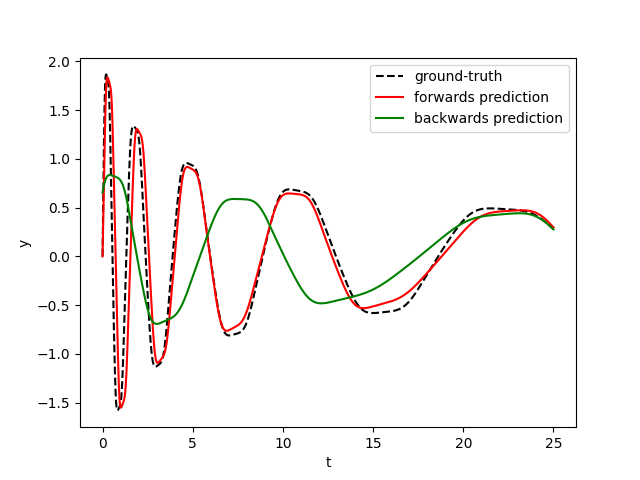}
  \caption{y values prediction.}
  \label{fig:sub2}
\end{subfigure}
\caption{Neural CODE at reconstructing the dynamics of a spiral ODE using the $1000/2000$ dataset. In dashed black the true dynamics, in red the predictions forward and in green the predictions backward in time.}
\label{fig:TBAODE_spiral2}
\end{figure}

In Figures \ref{fig:NODE_spiral} and \ref{fig:TBAODE_spiral}, the visualisation of spiral dynamics and the predicted values $(\hat{x}_i,\hat{y}_i)$ with the testing set, of the $2000/1000$ dataset, clearly demonstrates that Neural CODE outperforms Neural ODE in terms of prediction accuracy in both forward and backward reconstruction tasks.
Furthermore, for the $1000/2000$ dataset, the visual representations, in Figures \ref{fig:NODE_spiral2} and \ref{fig:TBAODE_spiral2}, confirm our earlier conclusions. In forward time predictions, Neural CODE exhibits superior modelling capabilities compared to Neural ODE. Additionally, the backward dynamics of Neural CODE better aligns with the true dynamics, albeit deviating from the periodic function that models the $x$ and $y$ values.
It's important to note that these visualisations were generated using the models at the final training iteration. Upon examining the evolution plots of MSE, it becomes evident that implementing an early stopping criterion would be beneficial.

\paragraph{Case Study 2: Real-world time-series}




We evaluated and compared the recurrent models CODE-RNN/-GRU/-LSTM and CODE-BiRNN/-BiGRU/-BiLSTM, using ODE-RNN/-GRU/-LSTM baselines, when modelling three real-world time-series datasets with different characteristics (available in \emph{Kaggle}):

\begin{itemize}
    \item Daily Climate time-series Data (regularly sampled), denoted by DC,  \citep{vraoDailyClimateTime};
    \item Hydropower modelling with hydrological data (regularly sampled with sparse data), denoted by HM, \citep{defeliceHydropowerModellingHydrological};
    \item DJIA 30 Stock time-series (irregularly sampled), denoted by DJIA,  \citep{szrleeDJIA30Stock}.
\end{itemize}

For each dataset, the performance of the models was analysed at three distinct tasks: 
\begin{enumerate}
    \item[(i)] Missing data imputation: predicting an observation $\boldsymbol{\hat{y}}_{i+1}$ at $t_{i+1}$ between observations at $t_i \text{ and } t_{i+2}$, for $i = 1,3,6\dots$. \\
    \textbf{Remark:} Recurrent architectures based-on Neural ODE (ODE-RNN/-GRU/-LSTM) only receive the observation at $t_i$ and predict the observation at $t_{i+1}$ while architectures based-on Neural CODE (CODE-RNN/-GRU/-LSTM and CODE-BiRNN/-BiGRU/-BiLSTM) receive both observations at $t_i$ (for the IVP) and at $t_{i+2}$ (for the FVP) to predict the observation at $t_{i+1}$, being able to capture more information.

    \item[(ii)] Forward extrapolation: given a sequence of length 7 or 15 predict the observations for the next 7 or 15 time-steps.

    \item[(iii)] Backward extrapolation: given a sequence of length 7 or 15 predict the observations for the past 7 or 15 time-steps. \\
    \textbf{Remark:} For doing predictions backward in time with ODE/-RNN/-GRU/-LSTM, the $ODESolve$ receives the time interval in the reverse order.
\end{enumerate}
        
The same training conditions were applied to all models for all datasets and tasks. The datasets were divided into 75\% training and 25\% testing points. The training was conducted with a batch size of 1, 50 epochs, and the Adam optimiser with a learning rate of 0.0005. For the $ODESolve$ we use the Runge-Kutta method of order 5 (Dormand-Prince-Shampine) with the default configurations. 
The NN architecture that builds the ODE dynamics used in this study consists of three layers. The input layer contains 1 neuron (or 4 neurons when predicting 4 features or 39 neurons when predicting 39 features), representing the input features of the model. The hidden layer is composed of 256 neurons, each using the exponential linear unit activation function. Finally, the output layer consists of 1 neuron (or 4 neurons when predicting 4 features or 39 neurons when predicting 39 features), representing the output of the model. 
All architectures use a single RNN, GRU or LSTM cell with an input layer containing 1 neuron (or 4 neurons when predicting 4 features or 39 neurons when predicting 39 features), representing the size of output given by the ODE solver. The hidden layer is composed of 256 neurons and the output layer contains 1 neuron (or 4 neurons when predicting 4 features or 39 neurons when predicting 39 features).
To account for random weight initialisation, we trained and tested each model three
times $(R=3)$. To evaluate the performance of the models, we computed the average of the
MSE ($MSE_{avg}$) and standard deviation ($std_{avg}$) values for the test sets, from the three runs.
Furthermore, to analyse and compare the convergence of the proposed networks with the baselines, the training losses of CODE-RNN/-GRU/-LSTM and CODE-BiRNN/-BiGRU/-BiLSTM were plotted, along with ODE-RNN/-GRU/-LSTM baselines, for the missing data imputation and future extrapolation tasks.

\textbf{Remark:} For the backward extrapolation task, the training losses were not plotted since the Neural ODE-based architectures (ODE-RNN/-GRU/-LSTM) solely compute the losses in the forward direction of time.

\subsubsection{Daily Climate time-series Data}

This dataset consists of 1462 data points with 4 daily features: mean temperature, humidity, wind speed, and mean pressure. These features were experimentally measured in Delhi, India for weather forecasting \citep{vraoDailyClimateTime}. 

\paragraph{Missing data imputation}

\begin{figure}[h]
\centering
\begin{subfigure}{0.329\textwidth}
  \centering
  \includegraphics[width=\linewidth]{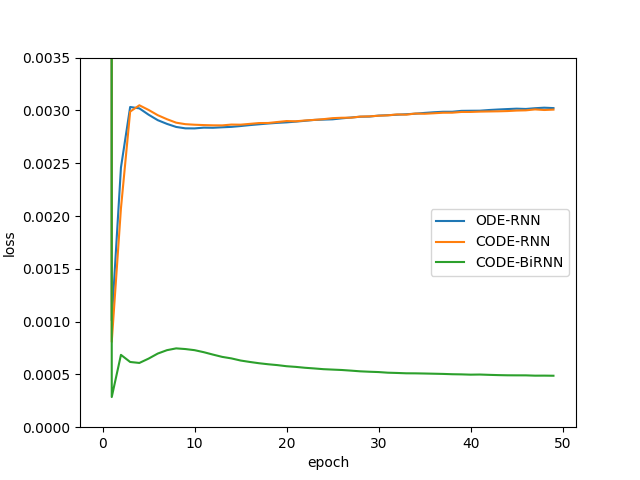}
  \caption{}
  \label{fig:climateLoss1sub1}
\end{subfigure}%
\begin{subfigure}{.329\textwidth}
  \centering
  \includegraphics[width=\linewidth]{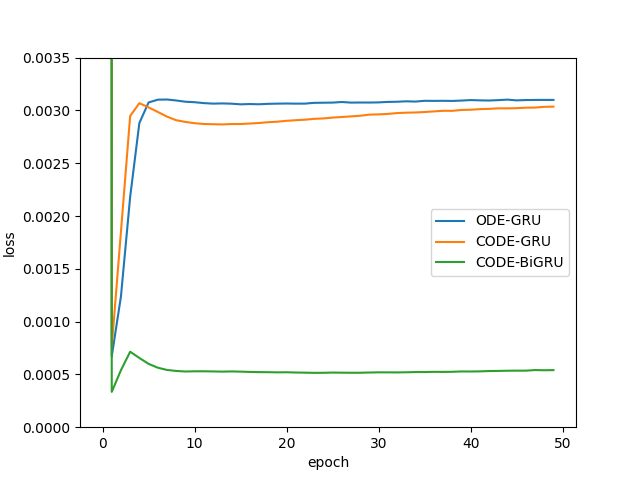}
  \caption{}
  \label{fig:climateLoss1sub2}
\end{subfigure}
\begin{subfigure}{.329\textwidth}
  \centering
  \includegraphics[width=\linewidth]{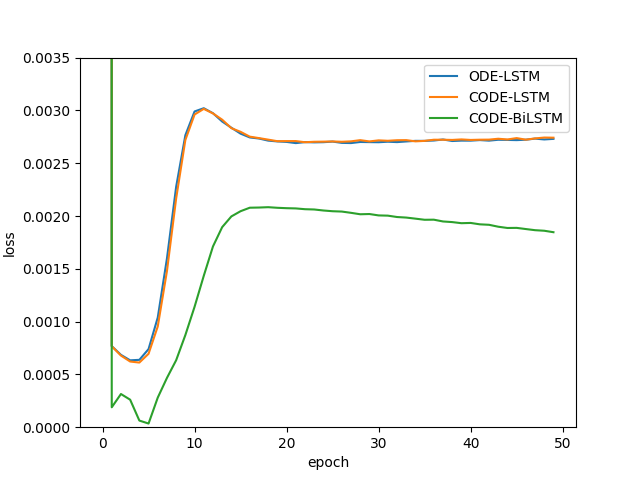}
  \caption{}
  \label{fig:climateLoss1sub3}
\end{subfigure}
\caption{Training loss through the epochs for the missing data imputation task for the DC dataset, for 1/1. (a) ODE-RNN, CODE-RNN, CODE-BiRNN; (b) ODE-GRU, CODE-GRU, CODE-BiGRU; (c) ODE-LSTM, CODE-LSTM, CODE-BiLSTM.}
\label{fig:climateLoss1}
\end{figure}

\begin{figure}[h]
\centering
\begin{subfigure}{0.329\textwidth}
  \centering
  \includegraphics[width=\linewidth]{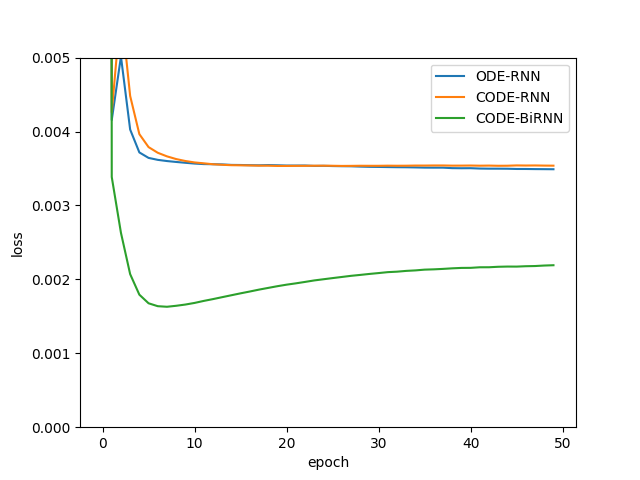}
  \caption{}
  \label{fig:sub1}
\end{subfigure}%
\begin{subfigure}{.329\textwidth}
  \centering
  \includegraphics[width=\linewidth]{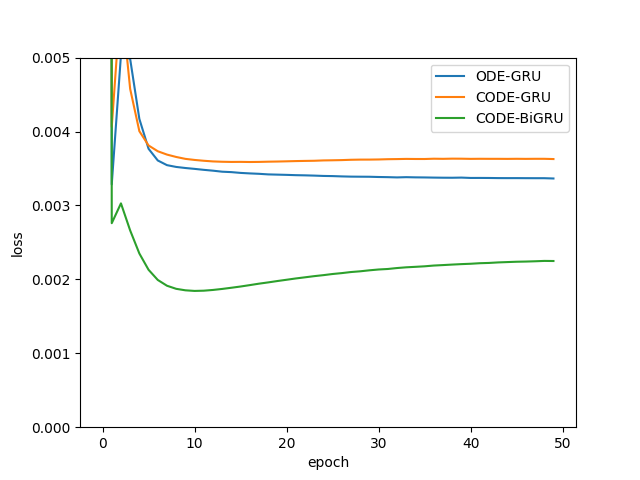}
  \caption{}
  \label{fig:sub2}
\end{subfigure}
\begin{subfigure}{.329\textwidth}
  \centering
  \includegraphics[width=\linewidth]{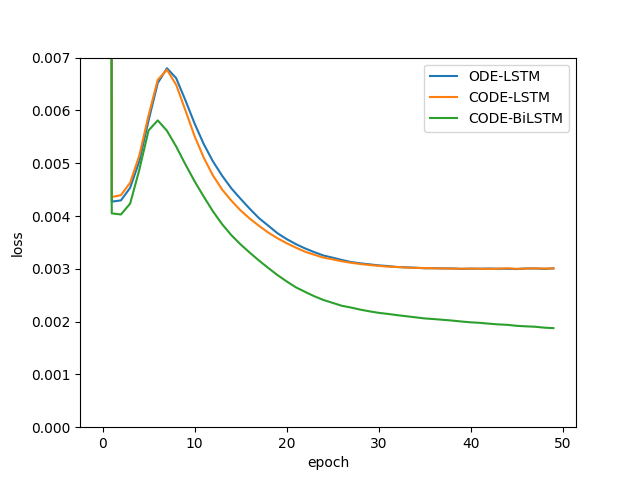}
  \caption{}
  \label{fig:sub2}
\end{subfigure}
\caption{Training loss through the epochs for the missing data imputation task for the DC dataset, for 4/4. (a) ODE-RNN, CODE-RNN, CODE-BiRNN; (b) ODE-GRU, CODE-GRU, CODE-BiGRU; (c) ODE-LSTM, CODE-LSTM, CODE-BiLSTM.}
\label{fig:climateLoss2}
\end{figure}

We performed the missing data imputation task to predict data points with 1 feature (mean temperature) as well as all 4 available features, denoted by 1/1 and 4/4 respectively. The numerical results are shown in Table \ref{tab:climate1.2.1} for architectures ODE-RNN, CODE-RNN and CODE-BiRNN, Table \ref{tab:climate1.2.2} for architectures ODE-GRU, CODE-GRU and CODE-BiGRU, and Table \ref{tab:climate1.2.3} for architectures ODE-LSTM, CODE-LSTM and CODE-BiLSTM. The evolution of the loss during training for the predictions with 1/1 and 4/4 features are depicted in Figures \ref{fig:climateLoss1} and \ref{fig:climateLoss2}, respectively.

From \textcolor{black}{Tables \ref{tab:climate1.2.1}-\ref{tab:climate1.2.3}}, the performance of CODE-BiRNN/-BiGRU/-BiLSTM stand out, from the others, by offering the best performance. Among these three variants, they present similar performance.
They achieve $MSE_{avg}$ values smaller by an order of magnitude for 1/1 and 4/4.
When comparing ODE-RNN/-GRU/-LSTM with CODE-RNN/-GRU/-LSTM the performances are similar.
In general, all architectures exhibit a higher $MSE_{avg}$ values when the number of features increased.

\begin{table}[h]
\centering
\resizebox{\textwidth}{!}{
\begin{tabular}{@{}ccccc@{}}
\toprule
                      &            & ODE-RNN                  & CODE-RNN                 & CODE-BiRNN                     \\ \midrule
seen/predict features & $(t_0, t_f)$ & $MSE_{avg}\pm std_{avg}$ & $MSE_{avg}\pm std_{avg}$ & $MSE_{avg}\pm std_{avg}$       \\ \midrule
1/1                   & (0, 1462)  & 1.15e-2$\pm$4.15e-6      & 1.14e-2$\pm$ 4.86e-6    & \textbf{7.20e-3$\pm$ 6.03e-6} \\
4/4                   & (0, 1462)  & 1.80e-2$\pm$ 2.50e-5    & 1.80e-2$\pm$ 2.10e-5    & \textbf{1.23e-2$\pm$ 4.54e-5} \\ \bottomrule
\end{tabular}
}
\caption{\textcolor{black}{Numerical results on the DC dataset at the missing data imputation task, for ODE-RNN, CODE-RNN and CODE-BiRNN. }}
\label{tab:climate1.2.1}
\end{table}

\begin{table}[h]
\centering
\resizebox{\textwidth}{!}{
\begin{tabular}{@{}ccccc@{}}
\toprule
                      &              & ODE-GRU                  & CODE-GRU                 & CODE-BiGRU                     \\ \midrule
seen/predict features & $(t_0, t_f)$ & $MSE_{avg}\pm std_{avg}$ & $MSE_{avg}\pm std_{avg}$ & $MSE_{avg}\pm std_{avg}$       \\ \midrule
1/1                   & (0, 1462)    & 1.15e-2$\pm$4.54e-6      & 1.14e-2$\pm$ 6.74e-6    & \textbf{7.10e-3$\pm$ 4.85e-6} \\
4/4                   & (0, 1462)    & 1.77e-2$\pm$ 2.47e-5    & 1.80e-2$\pm$ 7.15e-6    & \textbf{1.21e-2$\pm$ 2.24e-5} \\ \bottomrule
\end{tabular}
}
\caption{\textcolor{black}{Numerical results on the DC dataset at the missing data imputation task, for ODE-GRU, CODE-GRU and CODE-BiGRU.}}
\label{tab:climate1.2.2}
\end{table}

\begin{table}[h]
\centering
\resizebox{\textwidth}{!}{
\begin{tabular}{@{}ccccc@{}}
\toprule
                      &              & ODE-LSTM                 & CODE-LSTM                & CODE-BiLSTM                    \\ \midrule
seen/predict features & $(t_0, t_f)$ & $MSE_{avg}\pm std_{avg}$ & $MSE_{avg}\pm std_{avg}$ & $MSE_{avg}\pm std_{avg}$       \\ \midrule
1/1                   & (0, 1462)    & 1.11e-2$\pm$ 1.30e-5    & 0.0110$\pm$ 2.42e-5    & \textbf{7.90e-3$\pm$ 8.29e-6} \\
4/4                   & (0, 1462)    & 1.73e-2$\pm$ 1.31e-5    & 0.0173$\pm$ 6.20e-6    & \textbf{1.29e-2$\pm$ 7.84e-5} \\ \bottomrule
\end{tabular}
}
\caption{\textcolor{black}{Numerical results on the DC dataset at the missing data imputation task, for ODE-LSTM, CODE-LSTM and CODE-BiLSTM.}}
\label{tab:climate1.2.3}
\end{table}


Figures \ref{fig:climateLoss1} and \ref{fig:climateLoss2} show that the CODE-BiRNN/-BiGRU/-BiLSTM stand out, having faster convergence and achieving lower loss values. When analysing the loss values of ODE-RNN/-GRU/-LSTM and CODE-RNN/-GRU/-LSTM, they present similar behaviour. We note that, analysing the loss values throughout the training process corroborate the test results (\textcolor{black}{Tables \ref{tab:climate1.2.1}-\ref{tab:climate1.2.3}}).

\paragraph{Future extrapolation}

\begin{figure}[h]
\centering
\begin{subfigure}{0.329\textwidth}
  \centering
  \includegraphics[width=\linewidth]{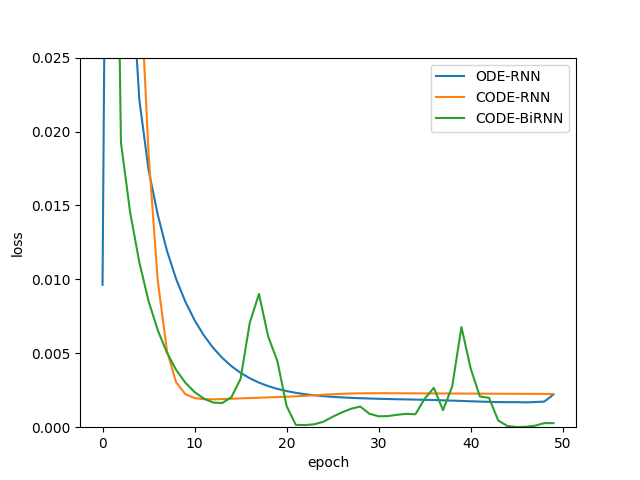}
  \caption{}
  \label{fig:sub1}
\end{subfigure}%
\begin{subfigure}{.329\textwidth}
  \centering
  \includegraphics[width=\linewidth]{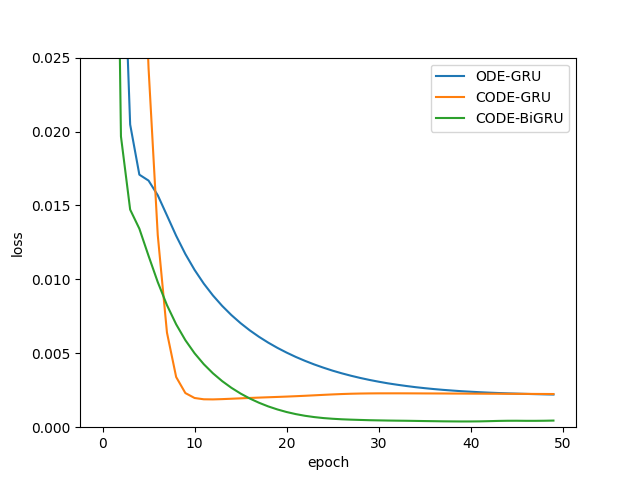}
  \caption{}
  \label{fig:sub2}
\end{subfigure}
\begin{subfigure}{.329\textwidth}
  \centering
  \includegraphics[width=\linewidth]{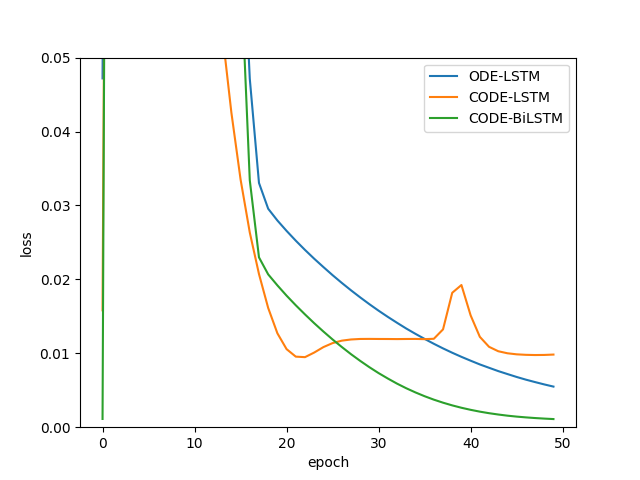}
  \caption{}
  \label{fig:sub2}
\end{subfigure}
\caption{Training loss through the epochs for the future extrapolation task for the DC dataset, for $7/7$. (a) ODE-RNN, CODE-RNN, CODE-BiRNN; (b) ODE-GRU, CODE-GRU, CODE-BiGRU; (c) ODE-LSTM, CODE-LSTM, CODE-BiLSTM.}
\label{fig:climateLoss3}
\end{figure}
\begin{figure}[h]
\centering
\begin{subfigure}{0.329\textwidth}
  \centering
  \includegraphics[width=\linewidth]{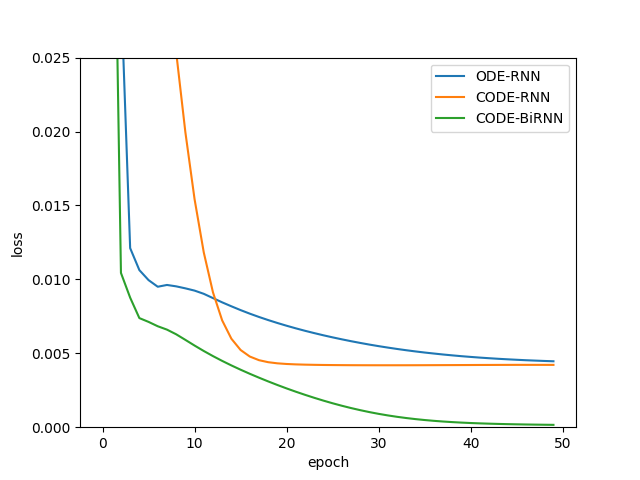}
  \caption{}
  \label{fig:sub1}
\end{subfigure}%
\begin{subfigure}{.329\textwidth}
  \centering
  \includegraphics[width=\linewidth]{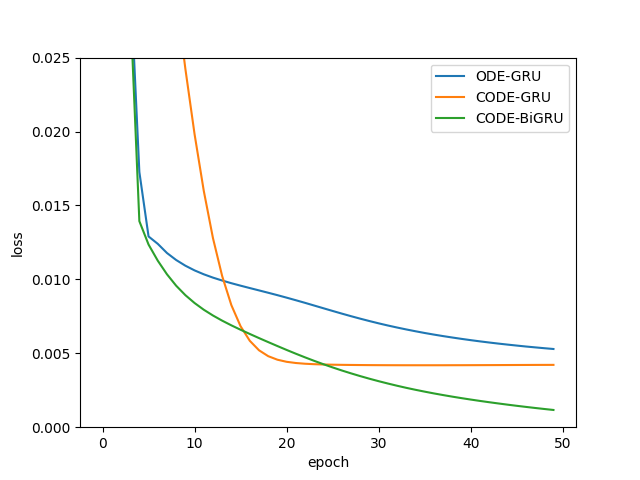}
  \caption{}
  \label{fig:sub2}
\end{subfigure}
\begin{subfigure}{.329\textwidth}
  \centering
  \includegraphics[width=\linewidth]{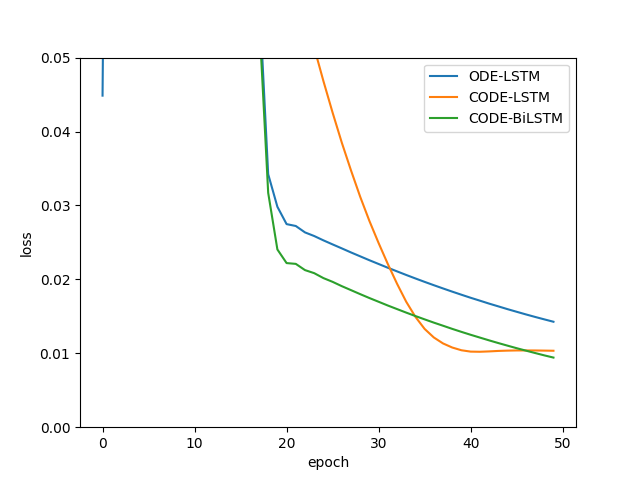}
  \caption{}
  \label{fig:sub2}
\end{subfigure}
\caption{Training loss through the epochs for future extrapolation task for the DC dataset, for $15/15$. (a) ODE-RNN, CODE-RNN, CODE-BiRNN; (b) ODE-GRU, CODE-GRU, CODE-BiGRU; (c) ODE-LSTM, CODE-LSTM, CODE-BiLSTM.}
\label{fig:climateLoss4}
\end{figure}

We performed the future extrapolation task to predict the mean temperature for the next 7 or 15 days after receiving the mean temperature for past 7 or 15 days, denoted by 7/7 and 15/15.
The numerical results are shown in Table \ref{tab:climate2.1} for architectures ODE-RNN, CODE-RNN and CODE-BiRNN, Table \ref{tab:climate2.2} for architectures ODE-GRU, CODE-GRU, CODE-BiGRU and Table \ref{tab:climate2.3} for architectures ODE-LSTM, CODE-LSTM and CODE-BiLSTM.  
The evolution of the losses during training for predictions with 7/7 and 15/15 are depicted in Figures \ref{fig:climateLoss3} and \ref{fig:climateLoss4}, respectively.

The results in \textcolor{black}{Tables \ref{tab:climate2.1}-\ref{tab:climate2.3}} show that, CODE-BiRNN/-BiGRU/-BiLSTM stand out from the others, consistently achieving the best predictive performance for 7/7 and 15/15. Among these variants, CODE-BiGRU has the best performance.
CODE-RNN and ODE-RNN have similar performance, while CODE-RNN present slightly better performance than CODE-GRU and CODE-LSTM, for 7/7.
 CODE-RNN/-GRU/-LSTM outperforms ODE-RNN/-GRU/-LSTM, for 15/15.

\begin{table}[h]
\centering
\resizebox{\textwidth}{!}{
\begin{tabular}{@{}ccccc@{}}
\toprule
                      &              & ODE-RNN                  & CODE-RNN                 & CODE-BiRNN                   \\ \midrule
days seen/predict & $(t_0, t_f)$ & $MSE_{avg}\pm std_{avg}$ & $MSE_{avg}\pm std_{avg}$ & $MSE_{avg}\pm std_{avg}$     \\ \midrule
7/7                   & (0, 1462)    & 1.56e-2 $\pm$ 7.00e-4      & 1.51e-2 $\pm$ 0.0003      & \textbf{5.90e-3 $\pm$ 2.40e-3} \\
15/15                   & (0, 1462)    & 6.303e-2 $\pm$ 3.82e-3    & 3.81e-2 $\pm$ 0.00056    & \textbf{5.6e-4 $\pm$ 4.00e-5} \\ \bottomrule
\end{tabular}
}
\caption{\textcolor{black}{Numerical results on the DC dataset at the future extrapolation task, for ODE-RNN, CODE-RNN and CODE-BiRNN.}}
\label{tab:climate2.1}
\end{table}

\begin{table}[h]
\centering
\resizebox{\textwidth}{!}{
\begin{tabular}{@{}ccccc@{}}
\toprule
                  &              & ODE-GRU                  & CODE-GRU                 & CODE-BiGRU                     \\ \midrule
days seen/predict & $(t_0, t_f)$ & $MSE_{avg}\pm std_{avg}$ & $MSE_{avg}\pm std_{avg}$ & $MSE_{avg}\pm std_{avg}$       \\ \midrule
7/7               & (0, 1462)    & 1.41e-2 $\pm$ 4.00e-5       & 1.53e-2 $\pm$ 3.00e-4      & \textbf{2.00e-3 $\pm$ 1.00e-4}   \\
15/15             & (0, 1462)    & 9.52e-2 $\pm$ 1.57e-2    & 3.81e-2 $\pm$ 2.00e-4     & \textbf{1.34e-3 $\pm$ 2.10e-4} \\ \bottomrule
\end{tabular}
}
\caption{\textcolor{black}{Numerical results on the DC dataset at the future extrapolation task, for ODE-GRU, CODE-GRU and CODE-BiGRU.}}
\label{tab:climate2.2}
\end{table}

\begin{table}[h]
\centering
\resizebox{\textwidth}{!}{
\begin{tabular}{ccccc}
\hline
                  &              & ODE-LSTM                 & CODE-LSTM                & CODE-BiLSTM                   \\ \hline
days seen/predict & $(t_0, t_f)$ & $MSE_{avg}\pm std_{avg}$ & $MSE_{avg}\pm std_{avg}$ & $MSE_{avg}\pm std_{avg}$      \\ \hline
7/7               & (0, 1462)    & 1.88e-2 $\pm$ 2.00e-4      & 3.87e-2 $\pm$ 3e-4      & \textbf{4.70e-3 $\pm$ 2.00e-4}  \\
15/15             & (0, 1462)    & 7.24e-2 $\pm$ 4.40e-4    & 8.51e-2 $\pm$ 3.80e-3     & \textbf{1.08e-2 $\pm$ 3.30e-4} \\ \hline
\end{tabular}
}
\caption{\textcolor{black}{Numerical results on the DC dataset at the future extrapolation task, for ODE-LSTM, CODE-LSTM and CODE-BiLSTM.}}
\label{tab:climate2.3}
\end{table}




Figures \ref{fig:climateLoss3} and \ref{fig:climateLoss4} show CODE-BiRNN/-BiGRU/-BiLSTM, in general, exhibit the fastest convergence rates and the lowest loss values. 
As the number of epochs increases, CODE-RNN/-GRU/-LSTM show a slightly rising trend in the loss values. This suggests that implementing an early stopping criterion could be beneficial for CODE-RNN/-GRU/-LSTM.

\paragraph{Backward extrapolation}




We performed the backward extrapolation task to predict the mean temperature for the past 7 or 15 days after receiving the mean temperature for the next 7 or 15 days, denoted by 7/7 and 15/15.
The numerical results are shown in Table \ref{tab:climate3.1} for architectures ODE-RNN, CODE-RNN and CODE-BiRNN, Table \ref{tab:climate3.2} for architectures ODE-GRU, CODE-GRU, CODE-BiGRU and Table \ref{tab:climate3.3} for architectures ODE-LSTM, CODE-LSTM and CODE-BiLSTM.  

From Tables \ref{tab:climate3.1}-\ref{tab:climate3.3}, the performance of the models on the backward extrapolation task are similar to that observed for the forward extrapolation task. 
Once again, it is worth emphasising that CODE-BiRNN/-BiGRU/-BiLSTM demonstrates the highest performance for 7/7 and 15/15. Among these variants, CODE-BiGRU performs best at 7/7, while CODE-BiRNN excels at 15/15.

Although predicting longer time horizons is more challenging, CODE-BiRNN,/-BiGRU/-BiLSTM achieve lower $MSE_{avg}$ values compared to smaller time horizons.

\begin{table}[h]
\centering
\resizebox{\textwidth}{!}{
\begin{tabular}{ccccc}
\hline
                  &              & ODE-RNN                  & CODE-RNN                     & CODE-BiRNN                   \\ \hline
days seen/predict & $(t_f, t_0)$ & $MSE_{avg}\pm std_{avg}$ & $MSE_{avg}\pm std_{avg}$     & $MSE_{avg}\pm std_{avg}$     \\ \hline
7/7               & (1462, 0)    & 6.70e-2 $\pm$ 2.40e-2      & \textbf{1.50e-2 $\pm$ 2.00e-4} & 0.0223 $\pm$ 0.0093          \\
15/15             & (1462, 0)    & 5.89e-2 $\pm$ 3.98e-3    & 3.38e-2 $\pm$ 5.60e-4        & \textbf{6.93e-3 $\pm$ 4.00e-5} \\ \hline
\end{tabular}
}
\caption{\textcolor{black}{Numerical results on the DC dataset at the backward extrapolation task, for ODE-RNN, CODE-RNN and CODE-BiRNN.}}
\label{tab:climate3.1}
\end{table}

\begin{table}[h]
\centering
\resizebox{\textwidth}{!}{
\begin{tabular}{ccccc}
\hline
                  &              & ODE-GRU                  & CODE-GRU                 & CODE-BiGRU                     \\ \hline
days seen/predict & $(t_f, t_0)$ & $MSE_{avg}\pm std_{avg}$ & $MSE_{avg}\pm std_{avg}$ & $MSE_{avg}\pm std_{avg}$       \\ \hline
7/7               & (1462, 0)    & 1.48e-2 $\pm$ 1.00e-4      & 1.51e-2 $\pm$ 2.00e-4      & \textbf{8.20e-3 $\pm$ 1.00e-4}   \\
15/15             & (1462, 0)    & 9.28e-2 $\pm$ 1.99e-2    & 3.37e-2 $\pm$ 1.70e-4     & \textbf{7.44e-3 $\pm$ 1.2e-4} \\ \hline
\end{tabular}
}
\caption{\textcolor{black}{Numerical results on the DC dataset at the backward extrapolation task, for ODE-GRU, CODE-GRU and CODE-BiGRU.}}
\label{tab:climate3.2}
\end{table}

\begin{table}[h]
\centering
\resizebox{\textwidth}{!}{
\begin{tabular}{ccccc}
\hline
                  &              & ODE-LSTM                 & CODE-LSTM                & CODE-BiLSTM                   \\ \hline
days seen/predict & $(t_f, t_0)$ & $MSE_{avg}\pm std_{avg}$ & $MSE_{avg}\pm std_{avg}$ & $MSE_{avg}\pm std_{avg}$      \\ \hline
7/7               & (1462, 0)    & 8.18e-2 $\pm$ 1.64e-2      & 4.81e-1 $\pm$ 5.08e-2      & \textbf{2.55e-2 $\pm$ 7.6e-3}  \\
15/15             & (1462, 0)    & 5.97e-2 $\pm$ 6.26e-3    & 13.96e-1 $\pm$ 1.02e-2    & \textbf{9.45e-3 $\pm$ 1.00e-4} \\ \hline
\end{tabular}
}
\caption{\textcolor{black}{Numerical results on the DC dataset at the backward extrapolation task, for ODE-LSTM, CODE-LSTM and CODE-BiLSTM.}}
\label{tab:climate3.3}
\end{table}

\subsubsection{Hydropower Modelling with Hydrological Data}

This dataset consists of weekly hydrological data from 27 European countries spanning the period 2015 to 2019, totalling 208 data points. The dataset includes hydropower inflows and average river discharges measured at national hydropower plants \citep{defeliceHydropowerModellingHydrological}. This dataset was specifically chosen for its limited amount of data, enabling the evaluation of how sparse training data affects the performance of the architectures.

\paragraph{Missing data imputation}

\begin{figure}[h]
\centering
\begin{subfigure}{0.329\textwidth}
  \centering
  \includegraphics[width=\linewidth]{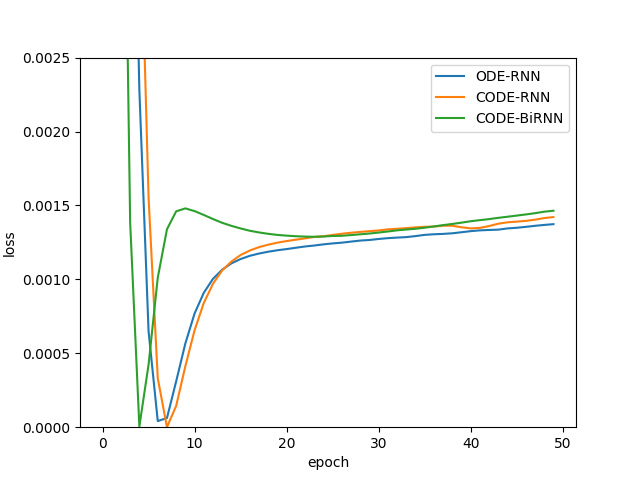}
  \caption{}
  \label{fig:sub1}
\end{subfigure}%
\begin{subfigure}{.329\textwidth}
  \centering
  \includegraphics[width=\linewidth]{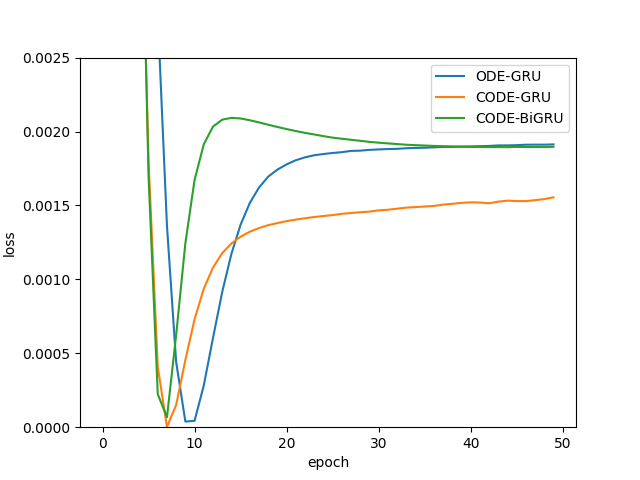}
  \caption{}
  \label{fig:sub2}
\end{subfigure}
\begin{subfigure}{.329\textwidth}
  \centering
  \includegraphics[width=\linewidth]{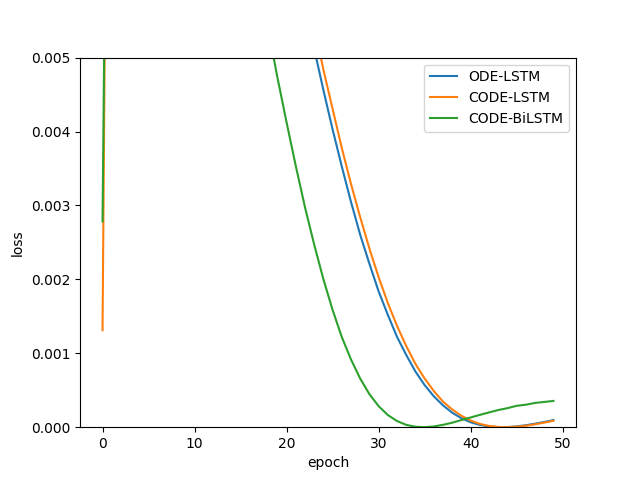}
  \caption{}
  \label{fig:sub2}
\end{subfigure}
\caption{Training loss through the epochs for the missing data imputation task for the HM dataset, for 1/1. (a) ODE-RNN, CODE-RNN, CODE-BiRNN; (b) ODE-GRU, CODE-GRU, CODE-BiGRU; (c) ODE-LSTM, CODE-LSTM, CODE-BiLSTM.}
\label{fig:hydroLoss1}
\end{figure}

\begin{figure}[h]
\centering
\begin{subfigure}{0.329\textwidth}
  \centering
  \includegraphics[width=\linewidth]{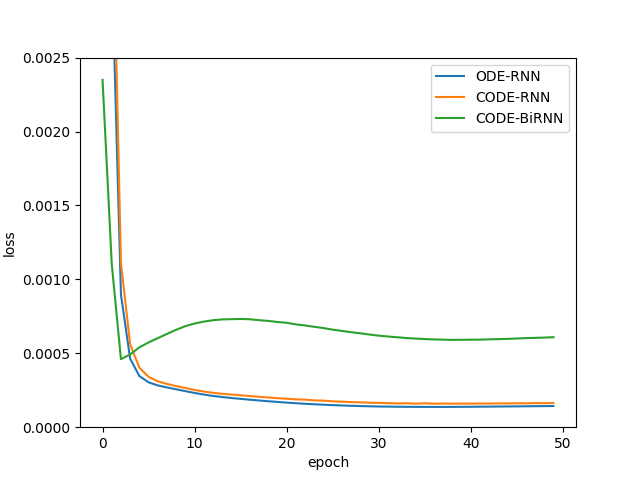}
  \caption{}
  \label{fig:sub1}
\end{subfigure}%
\begin{subfigure}{.329\textwidth}
  \centering
  \includegraphics[width=\linewidth]{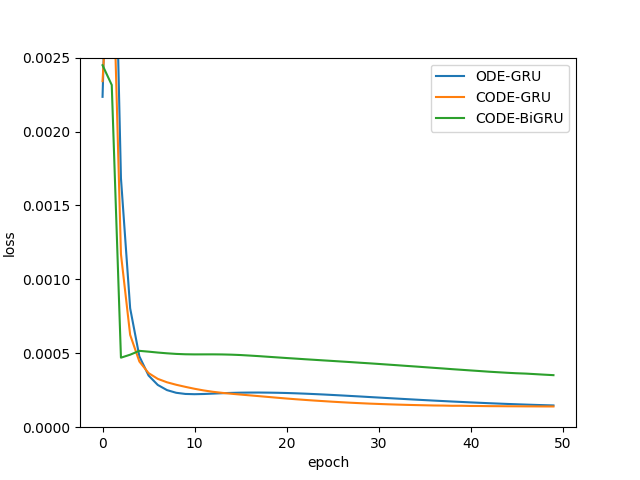}
  \caption{}
  \label{fig:sub2}
\end{subfigure}
\begin{subfigure}{.329\textwidth}
  \centering
  \includegraphics[width=\linewidth]{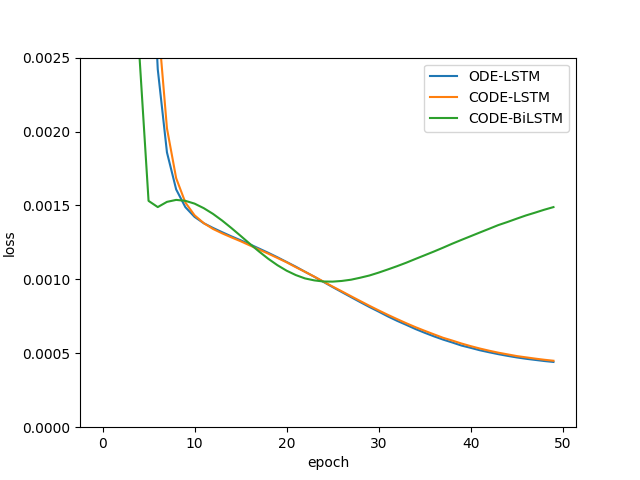}
  \caption{}
  \label{fig:sub2}
\end{subfigure}
\caption{Training loss through the epochs for the missing data imputation task for the HM dataset, for 39/39. (a) ODE-RNN, CODE-RNN, CODE-BiRNN; (b) ODE-GRU, CODE-GRU, CODE-BiGRU; (c) ODE-LSTM, CODE-LSTM, CODE-BiLSTM.}
\label{fig:hydroLoss2}
\end{figure}

We performed the missing data imputation task to predict data points with 1 feature (hydropower inflow) and 39 features (39 average river discharges), denoted by 1/1 and 39/39 respectively. The numerical results are shown in Table \ref{tab:hydro1.1} for architectures ODE-RNN, CODE-RNN and CODE-BiRNN, Table \ref{tab:hydro1.2} for architectures ODE-GRU, CODE-GRU and CODE-BiGRU, and Table \ref{tab:hydro1.3} for architectures ODE-LSTM, CODE-LSTM and CODE-BiLSTM. The evolution of the loss during training for the predictions with 1/1 and 39/39 features are depicted in Figures \ref{fig:hydroLoss1} and \ref{fig:hydroLoss2}, respectively.

From \textcolor{black}{Tables \ref{tab:hydro1.1}-\ref{tab:hydro1.3}}, the performance of CODE-BiRNN/-BiGRU/-BiLSTM stand out, from the others, by offering the best performance. Among these three variants, they present similar performance.
When comparing ODE-RNN/-GRU/-LSTM with CODE-RNN/-GRU/-LSTM the performances are similar. When the number of features is increased, all architectures present increased values of $MSE_{avg}$.

\begin{table}[h]
\centering
\resizebox{\textwidth}{!}{
\begin{tabular}{lcccc}
\hline
                      &              & ODE-RNN                  & CODE-RNN                 & CODE-BiRNN                     \\ \hline
seen/predict features & $(t_0, t_f)$ & $MSE_{avg}\pm std_{avg}$ & $MSE_{avg}\pm std_{avg}$ & $MSE_{avg}\pm std_{avg}$       \\ \hline
1/1                   & (0, 208)     & 9.00e-3$\pm$ 9.83e-5    & 8.90e-3$\pm$ 7.79e-5    & \textbf{2.20e-3$\pm$ 3.83e-6} \\
39/39                 & (0, 208)     & 2.56e-2$\pm$ 3.00e-4       & 2.52e-2$\pm$ 2.00e-4       & \textbf{1.09e-2$\pm$ 9.80e-5} \\ \hline
\end{tabular}
}
\caption{\textcolor{black}{Numerical results on the HM dataset at the missing data imputation task, for ODE-RNN, CODE-RNN and CODE-BiRNN.}}
\label{tab:hydro1.1}
\end{table}

\begin{table}[h]
\centering
\resizebox{\textwidth}{!}{
\begin{tabular}{lcccc}
\hline
                      &              & ODE-GRU                  & CODE-GRU                 & CODE-BiGRU                     \\ \hline
seen/predict features & $(t_0, t_f)$ & $MSE_{avg}\pm std_{avg}$ & $MSE_{avg}\pm std_{avg}$ & $MSE_{avg}\pm std_{avg}$       \\ \hline
1/1                   & (0, 208)     & 8.20e-3$\pm$ 9.55e-6    & 8.70e-3$\pm$ 5.95e-5    & \textbf{2.3e-3$\pm$ 9.68e-6} \\
39/39                 & (0, 208)     & 2.46e-2$\pm$ 4.00e-4       & 2.49e-2$\pm$ 1.00e-4       & \textbf{1.03e-2$\pm$ 2.00e-4}    \\ \hline
\end{tabular}
}
\caption{\textcolor{black}{Numerical results on the HM dataset at the missing data imputation task, for ODE-GRU, CODE-GRU and CODE-BiGRU.}}
\label{tab:hydro1.2}
\end{table}

\begin{table}[h]
\centering
\resizebox{\textwidth}{!}{
\begin{tabular}{lcccc}
\hline
                      &              & ODE-LSTM                 & CODE-LSTM                & CODE-BiLSTM                    \\ \hline
seen/predict features & $(t_0, t_f)$ & $MSE_{avg}\pm std_{avg}$ & $MSE_{avg}\pm std_{avg}$ & $MSE_{avg}\pm std_{avg}$       \\ \hline
1/1                   & (0, 208)     & 1.13e-2$\pm$ 3.47e-5    & 1.12e-2$\pm$ 6.35e-5    & \textbf{3.20e-3$\pm$ 6.48e-5} \\
39/39                 & (0, 208)     & 2.38e-2$\pm$ 2.00e-4       & 2.37e-2$\pm$ 1.00e-4       & \textbf{1.04e-2$\pm$ 2.00e-4}    \\ \hline
\end{tabular}
}
\caption{\textcolor{black}{Numerical results on the HM dataset at the missing data imputation task, for ODE-LSTM, CODE-LSTM and CODE-BiLSTM.}}
\label{tab:hydro1.3}
\end{table}

Figures \ref{fig:hydroLoss1} and \ref{fig:hydroLoss2} show that CODE-RNN/-GRU/-LSTM achieve the lowest $MSE_{avg}$ values out of all architectures. CODE-BiRNN/-BiGRU/-BiLSTM have faster convergence although ODE-RNN/-GRU/-LSTM and CODE-RNN/-GRU/-LSTM reach lower $MSE_{avg}$ values.
When analysing and comparing the loss values, during training, for ODE-RNN/-GRU/-LSTM and CODE-RNN/-GRU/-LSTM, they present similar behaviour. corroborating the test results in (\textcolor{black}{Tables \ref{tab:climate1.2.1}-\ref{tab:climate1.2.3}}).
From Figures \ref{fig:hydroLoss1} and \ref{fig:hydroLoss2}, CODE-BiRNN/-BiGRU/-BiLSTM present the highest training loss values however the test results show they have higher performance than the other architectures. Taking these conflicting pieces of information into account, we can conclude that the ODE-RNN and CODE-RNN models, along with their respective variants, exhibit poorer generalisation and may be prone to overfitting.

\paragraph{Future extrapolation}



\begin{figure}[h]
\centering
\begin{subfigure}{0.329\textwidth}
  \centering
  \includegraphics[width=\linewidth]{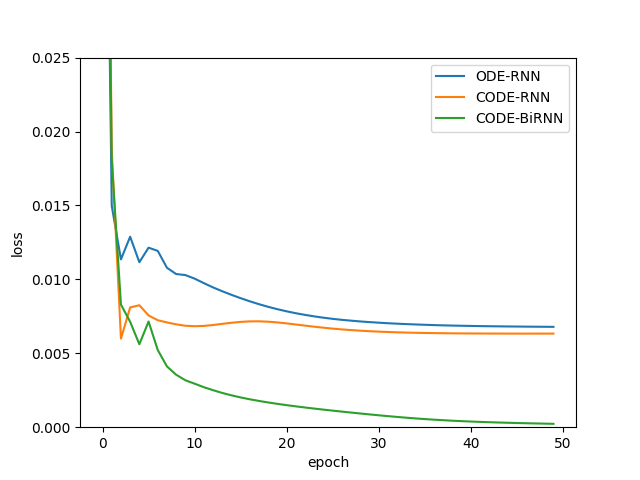}
  \caption{}
  \label{fig:sub1}
\end{subfigure}%
\begin{subfigure}{.329\textwidth}
  \centering
  \includegraphics[width=\linewidth]{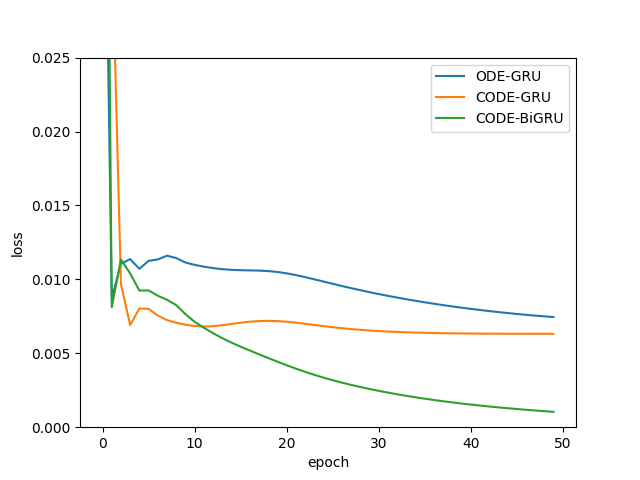}
  \caption{}
  \label{fig:sub2}
\end{subfigure}
\begin{subfigure}{.329\textwidth}
  \centering
  \includegraphics[width=\linewidth]{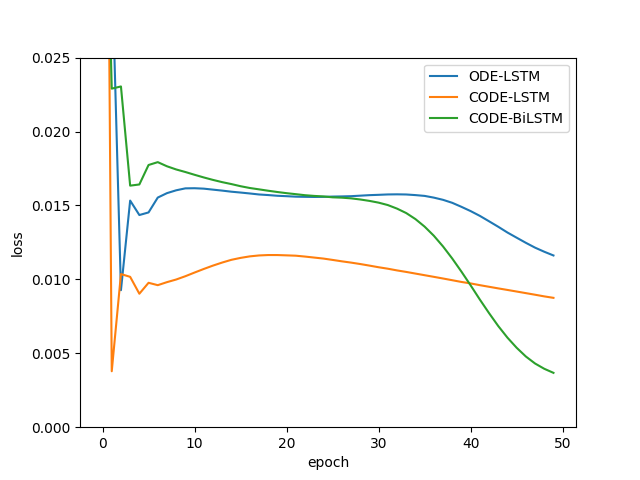}
  \caption{}
  \label{fig:sub2}
\end{subfigure}
\caption{Training loss through the epochs for future extrapolation task for the DC dataset, for $7/7$. (a) ODE-RNN, CODE-RNN, CODE-BiRNN; (b) ODE-GRU, CODE-GRU, CODE-BiGRU; (c) ODE-LSTM, CODE-LSTM, CODE-BiLSTM.}
\label{fig:hydroLoss3}
\end{figure}

\begin{figure}[h]
\centering
\begin{subfigure}{0.329\textwidth}
  \centering
  \includegraphics[width=\linewidth]{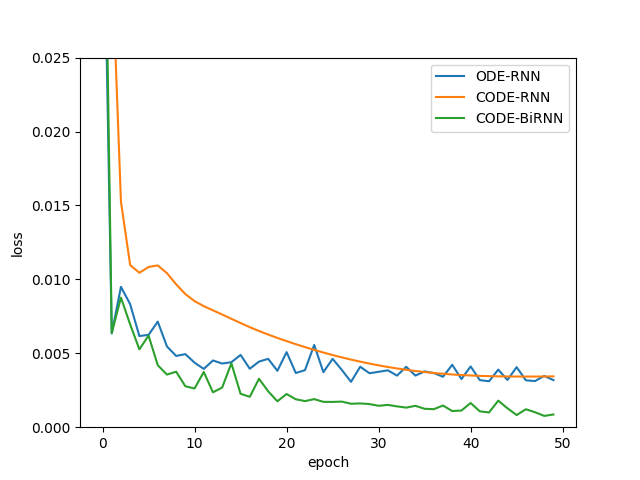}
  \caption{}
  \label{fig:sub1}
\end{subfigure}%
\begin{subfigure}{.329\textwidth}
  \centering
  \includegraphics[width=\linewidth]{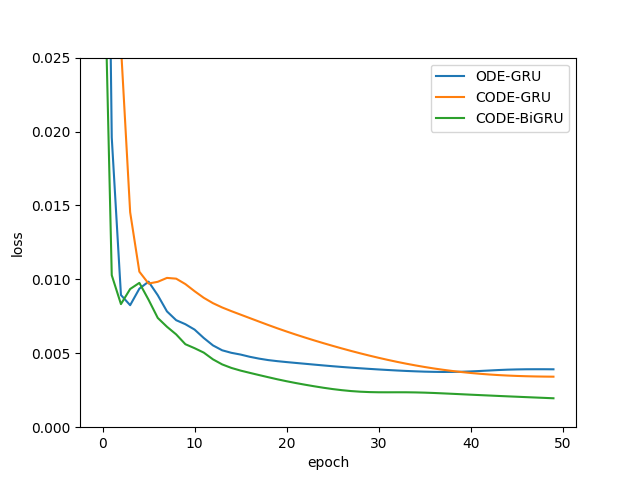}
  \caption{}
  \label{fig:sub2}
\end{subfigure}
\begin{subfigure}{.329\textwidth}
  \centering
  \includegraphics[width=\linewidth]{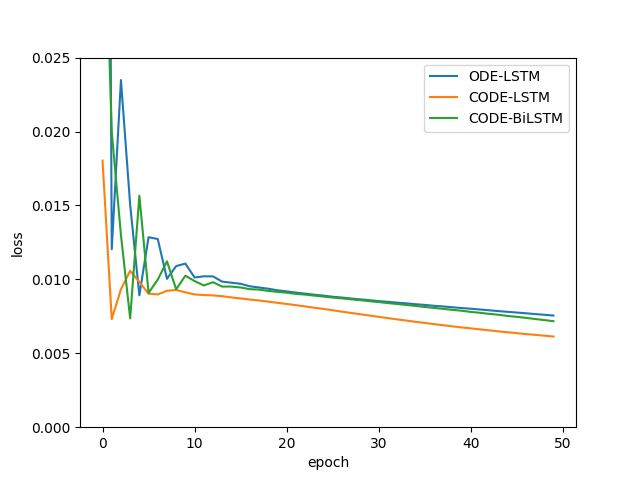}
  \caption{}
  \label{fig:sub2}
\end{subfigure}
\caption{Training loss through the epochs for future extrapolation task for the DC dataset, for $15/15$. (a) ODE-RNN, CODE-RNN, CODE-BiRNN; (b) ODE-GRU, CODE-GRU, CODE-BiGRU; (c) ODE-LSTM, CODE-LSTM, CODE-BiLSTM.}
\label{fig:hydroLoss4}
\end{figure}

We performed the future extrapolation task to predict the hydropower inflow for the next 7 or 15 weeks after receiving the hydropower inflow for past 7 or 15 weeks, denoted by 7/7 and 15/15.
The numerical results are shown in Table \ref{tab:hydro2.1} for architectures ODE-RNN, CODE-RNN and CODE-BiRNN, Table \ref{tab:hydro2.2} for architectures ODE-GRU, CODE-GRU, CODE-BiGRU and Table \ref{tab:hydro2.3} for architectures ODE-LSTM, CODE-LSTM and CODE-BiLSTM.  
The evolution of the losses during training for predictions with 7/7 and 15/15 are depicted in Figures \ref{fig:hydroLoss3} and \ref{fig:hydroLoss4}, respectively.

The results in \ref{tab:hydro2.1}-\ref{tab:hydro2.3} show that, CODE-BiRNN/-BiGRU/-BiLSTM stand out from the others, consistently achieving the best predictive performance for 7/7 and 15/15. Among these variants, CODE-BiRNN has the best performance.
When comparing ODE-RNN/-GRU/-LSTM and CODE-RNN/-GRU/-LSTM, the architectures present similar performance for 7/7 and 15/15.

\begin{table}[h]
\centering
\resizebox{\textwidth}{!}{
\begin{tabular}{ccccc}
\hline
                   &              & ODE-RNN                  & CODE-RNN                 & CODE-BiRNN                   \\ \hline
weeks seen/predict & $(t_0, t_f)$ & $MSE_{avg}\pm std_{avg}$ & $MSE_{avg}\pm std_{avg}$ & $MSE_{avg}\pm std_{avg}$     \\ \hline
7/7                & (0, 208)     & 7.33e-2 $\pm$ 2.00e-4      & 8.53e-2 $\pm$ 2.00e-4      & \textbf{1.00e-3 $\pm$ 6.00e-5}  \\
15/15              & (0, 208)     & 1.40e-1 $\pm$ 5.00e-4      & 1.05e-1 $\pm$ 1.20e-3      & \textbf{1.61e-2 $\pm$ 2.30e-3} \\ \hline
\end{tabular}
}
\caption{\textcolor{black}{Numerical results on the HM dataset at the future extrapolation task, for ODE-RNN, CODE-RNN and CODE-BiRNN.}}
\label{tab:hydro2.1}
\end{table}

\begin{table}[h]
\centering
\resizebox{\textwidth}{!}{
\begin{tabular}{ccccc}
\hline
                   &              & ODE-GRU                  & CODE-GRU                 & CODE-BiGRU                   \\ \hline
weeks seen/predict & $(t_0, t_f)$ & $MSE_{avg}\pm std_{avg}$ & $MSE_{avg}\pm std_{avg}$ & $MSE_{avg}\pm std_{avg}$     \\ \hline
7/7                & (0, 208)     & 8.35e-2 $\pm$ 9.00e-4      & 8.50e-2 $\pm$ 1.00e-4      & \textbf{4.00e-3 $\pm$ 2.00e-4} \\
15/15              & (0, 208)     & 1.76e-1 $\pm$ 1.01e-2      & 1.10e-1 $\pm$ 9.00e-5       & \textbf{3.87e-2 $\pm$ 2.00e-3} \\ \hline
\end{tabular}
}
\caption{\textcolor{black}{Numerical results on the HM dataset at the future extrapolation task, for ODE-GRU, CODE-GRU and CODE-BiGRU.}}
\label{tab:hydro2.2}
\end{table}

\begin{table}[h]
\centering
\resizebox{\textwidth}{!}{
\begin{tabular}{ccccc}
\hline
                   &              & ODE-LSTM                 & CODE-LSTM                & CODE-BiLSTM                  \\ \hline
weeks seen/predict & $(t_0, t_f)$ & $MSE_{avg}\pm std_{avg}$ & $MSE_{avg}\pm std_{avg}$ & $MSE_{avg}\pm std_{avg}$     \\ \hline
7/7                & (0, 208)     & 1.03e-1 $\pm$ 8.00e-4      & 9.63e-2 $\pm$ 9.00e-5       & \textbf{4.18e-2 $\pm$ 1.30e-3} \\
15/15              & (0, 208)     & 1.40e-1 $\pm$ 3.00e-4      & 1.41e-1 $\pm$ 3.00e-4      & \textbf{1.10e-1 $\pm$ 7.00e-4} \\ \hline
\end{tabular}
}
\caption{\textcolor{black}{Numerical results on the HM dataset at the future extrapolation task, for ODE-LSTM, CODE-LSTM and CODE-BiLSTM.}}
\label{tab:hydro2.3}
\end{table}

Figures \ref{fig:hydroLoss3} and \ref{fig:hydroLoss4} show CODE-BiRNN/-BiGRU/-BiLSTM, in general, exhibit the fastest convergence rates and the lowest loss values. 
ODE-RNN/-GRU/-LSTM and CODE-RNN/-GRU/-LSTM present achieve similar $MSE_{avg}$ values at 50 epochs although, in general, CODE-RNN/-GRU/-LSTM achieve lower $MSE_{avg}$ values faster.

\paragraph{Backward extrapolation}


We performed the backward extrapolation task to predict the hydropower inflow for the past 7 or 15 weeks after receiving the hydropower inflow for the next 7 or 15 weeks, denoted by 7/7 and 15/15.
The numerical results are shown in Table \ref{tab:hydro3.1} for architectures ODE-RNN, CODE-RNN and CODE-BiRNN, Table \ref{tab:hydro3.2} for architectures ODE-GRU, CODE-GRU, CODE-BiGRU and Table \ref{tab:hydro3.3} for architectures ODE-LSTM, CODE-LSTM and CODE-BiLSTM.  

From Tables \ref{tab:hydro3.1}-\ref{tab:hydro3.3}, the performance of the models on the backward extrapolation task are similar to that observed for the forward extrapolation task. 
Once again, it is worth emphasising that CODE-BiRNN/-BiGRU/-BiLSTM demonstrates the highest performance for 7/7 and 15/15. Among these variants, CODE-BiRNN performs the best. 
As expected, when predicting longer time horizons, all architectures show an increase in the $MSE_{avg}$ values compared to smaller time horizons.

\begin{table}[h]
\centering
\resizebox{\textwidth}{!}{
\begin{tabular}{ccccc}
\hline
                   &              & ODE-RNN                  & CODE-RNN                 & CODE-BiRNN                   \\ \hline
weeks seen/predict & $(t_f, t_0)$ & $MSE_{avg}\pm std_{avg}$ & $MSE_{avg}\pm std_{avg}$ & $MSE_{avg}\pm std_{avg}$     \\ \hline
7/7                & (208, 0)     & 5.24e-2 $\pm$ 3.00e-4      & 4.26e-2 $\pm$ 2.00e-5       & \textbf{2.25e-2 $\pm$ 1.00e-4} \\
15/15              & (208, 0)     & 1.41e-1 $\pm$ 0.0019      & 1.04e-1 $\pm$ 1.80e-3      & \textbf{4.25e-2 $\pm$ 1.20e-3} \\ \hline
\end{tabular}
}
\caption{\textcolor{black}{Numerical results on the HM dataset at the backward extrapolation task, for ODE-RNN, CODE-RNN and CODE-BiRNN.}}
\label{tab:hydro3.1}
\end{table}

\begin{table}[h]
\centering
\resizebox{\textwidth}{!}{
\begin{tabular}{ccccc}
\hline
                   &              & ODE-GRU                  & CODE-GRU                 & CODE-BiGRU                   \\ \hline
weeks seen/predict & $(t_f, t_0)$ & $MSE_{avg}\pm std_{avg}$ & $MSE_{avg}\pm std_{avg}$ & $MSE_{avg}\pm std_{avg}$     \\ \hline
7/7                & (208, 0)     & 5.56e-2 $\pm$ 2.00e-3      & 4.26e-2 $\pm$ 7.00e-5       & \textbf{2.58e-2 $\pm$ 2.00e-4} \\
15/15              & (208, 0)     & 1.86e-1 $\pm$ 1.27e-2      & 1.12e-1 $\pm$ 8.00e-5       & \textbf{4.34e-2 $\pm$ 7.00e-4} \\ \hline
\end{tabular}
}
\caption{\textcolor{black}{Numerical results on the HM dataset at the backward extrapolation task, for ODE-GRU, CODE-GRU and CODE-BiGRU.}}
\label{tab:hydro3.2}
\end{table}

\begin{table}[h]
\centering
\resizebox{\textwidth}{!}{
\begin{tabular}{ccccc}
\hline
                   &              & ODE-LSTM                 & CODE-LSTM                & CODE-LSTM                    \\ \hline
weeks seen/predict & $(t_f, t_0)$ & $MSE_{avg}\pm std_{avg}$ & $MSE_{avg}\pm std_{avg}$ & $MSE_{avg}\pm std_{avg}$     \\ \hline
7/7                & (208, 0)     & 1.18e-1 $\pm$ 5.70e-3      & 9.67e-2 $\pm$ 6.80e-3      & \textbf{3.36e-2 $\pm$ 5.00e-4} \\
15/15              & (208, 0)     & 1.38e-1 $\pm$ 2.00e-4      & 1.37e-1 $\pm$ 2.00e-4      & \textbf{1.09e-1 $\pm$ 8.00e-4} \\ \hline
\end{tabular}
}
\caption{\textcolor{black}{Numerical results on the HM dataset at the backward extrapolation task, for ODE-LSTM, CODE-LSTM and CODE-BiLSTM.}}
\label{tab:hydro3.3}
\end{table}

\footnotetext{For doing predictions backward in time using the $ODESolve$ in ODE-LSTM receives the time interval in the reverse order.}

\subsubsection{DJIA 30 Stock time-series}






The architectures were tested on an irregularly sampled dataset called DJIA 30 Stock time-series. The dataset consisted of $3019$ data points and included 6 daily features: the stock's price at open and close, the highest and lowest price, the number of shares traded, and the stock's ticker name. The data was collected between January 3, 2006, and December 29, 2017, for 29 DJIA companies \citep{szrleeDJIA30Stock}. However, for this particular study, data from a single company was utilised.

\paragraph{Missing data imputation}

\begin{figure}[h]
\centering
\begin{subfigure}{0.329\textwidth}
  \centering
  \includegraphics[width=\linewidth]{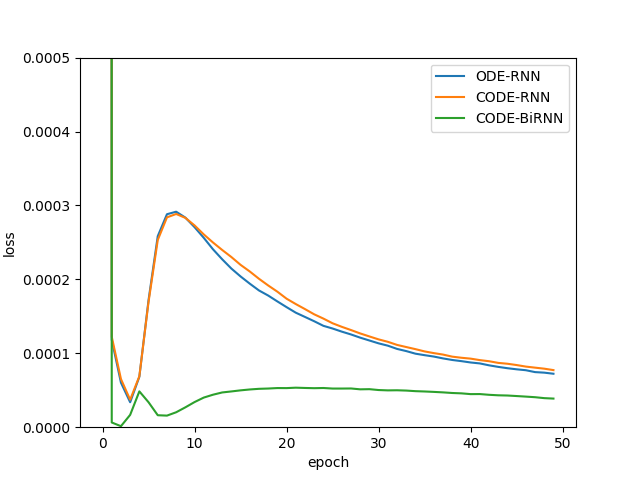}
  \caption{}
  \label{fig:sub1}
\end{subfigure}%
\begin{subfigure}{.329\textwidth}
  \centering
  \includegraphics[width=\linewidth]{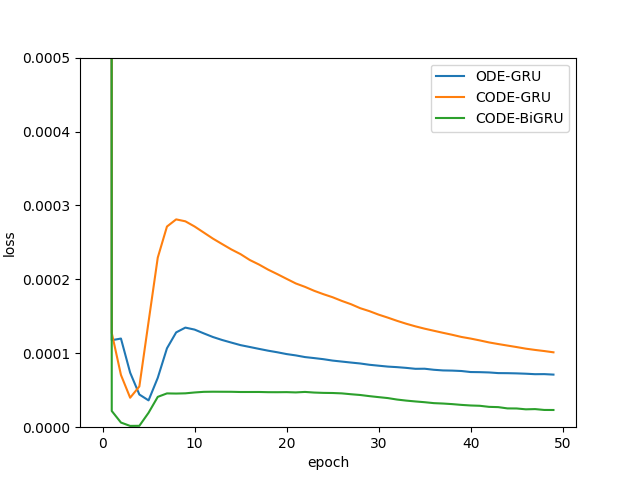}
  \caption{}
  \label{fig:sub2}
\end{subfigure}
\begin{subfigure}{.329\textwidth}
  \centering
  \includegraphics[width=\linewidth]{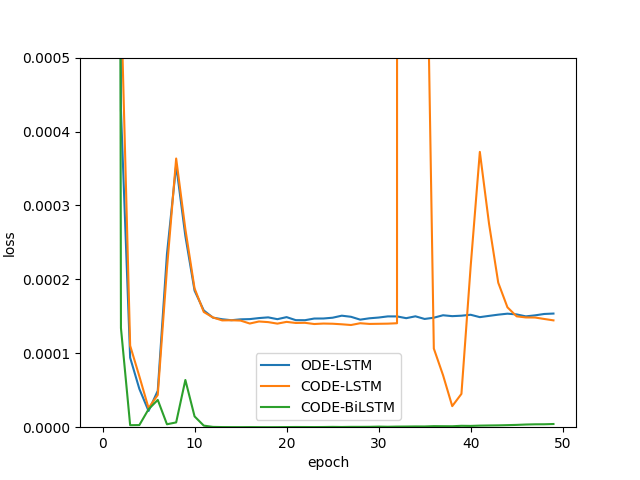}
  \caption{}
  \label{fig:sub2}
\end{subfigure}
\caption{Training loss through the epochs for the missing data imputation task for the HM dataset, for 1/1. (a) ODE-RNN, CODE-RNN, CODE-BiRNN; (b) ODE-GRU, CODE-GRU, CODE-BiGRU; (c) ODE-LSTM, CODE-LSTM, CODE-BiLSTM.}
\label{fig:djiaLoss1}
\end{figure}

\begin{figure}[h]
\centering
\begin{subfigure}{0.329\textwidth}
  \centering
  \includegraphics[width=\linewidth]{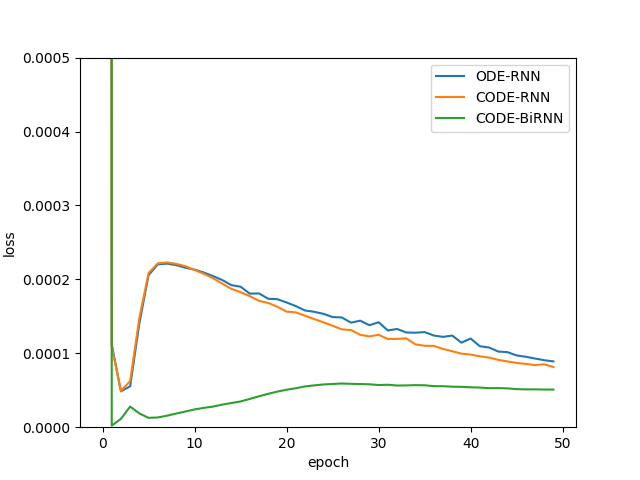}
  \caption{}
  \label{fig:sub1}
\end{subfigure}%
\begin{subfigure}{.329\textwidth}
  \centering
  \includegraphics[width=\linewidth]{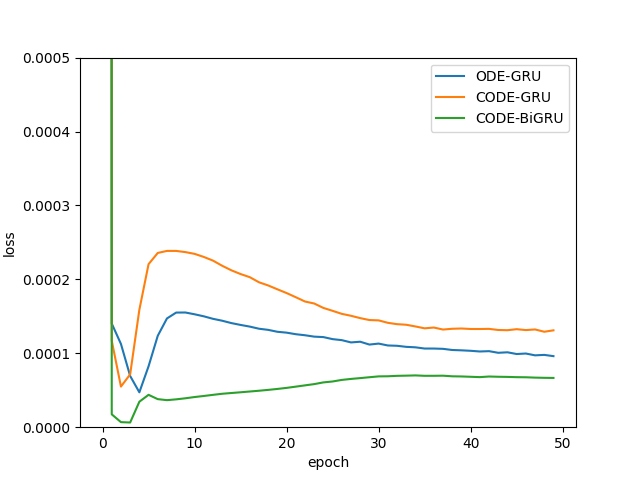}
  \caption{}
  \label{fig:sub2}
\end{subfigure}
\begin{subfigure}{.329\textwidth}
  \centering
  \includegraphics[width=\linewidth]{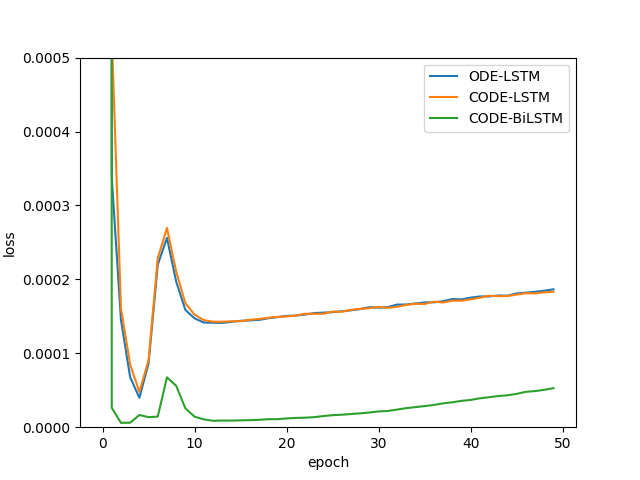}
  \caption{}
  \label{fig:sub2}
\end{subfigure}
\caption{Training loss through the epochs for the missing data imputation task for the HM dataset, for 4/4. (a) ODE-RNN, CODE-RNN, CODE-BiRNN; (b) ODE-GRU, CODE-GRU, CODE-BiGRU; (c) ODE-LSTM, CODE-LSTM, CODE-BiLSTM.}
\label{fig:djiaLoss2}
\end{figure}

We performed the missing data imputation task to predict data points with 1 feature (stock's price at close) and 4 features (stock's price at open and close, and the highest and lowest price), denoted by 1/1 and 39/39 respectively. The numerical results are shown in Table \ref{tab:DJIA1.1} for architectures ODE-RNN, CODE-RNN and CODE-BiRNN, Table \ref{tab:DJIA1.2} for architectures ODE-GRU, CODE-GRU and CODE-BiGRU, and Table \ref{tab:DJIA1.3} for architectures ODE-LSTM, CODE-LSTM and CODE-BiLSTM. The evolution of the loss during training for the predictions with 1/1 and 39/39 features are depicted in Figures \ref{fig:djiaLoss1} and \ref{fig:djiaLoss2}, respectively.

From \textcolor{black}{Tables \ref{tab:DJIA1.1}-\ref{tab:DJIA1.3}}, the performance of CODE-BiRNN/-BiGRU/-BiLSTM stand out, from the others, by offering the best performance. Among these three variants, CODE-BiRNN/-BiGRU offer the best performance.
When comparing ODE-RNN/-GRU/-LSTM with CODE-RNN/-GRU/-LSTM the performances are similar. When the number of features is increased, the $MSE_{avg}$ values do not suffer a significant increase.

\begin{table}[h]
\centering
\resizebox{\textwidth}{!}{
\begin{tabular}{lclll}
\hline
\multicolumn{1}{c}{}  &              & ODE-RNN                                      & CODE-RNN                                     & CODE-BiRNN                                   \\ \hline
seen/predict features & $(t_0, t_f)$ & \multicolumn{1}{c}{$MSE_{avg}\pm std_{avg}$} & \multicolumn{1}{c}{$MSE_{avg}\pm std_{avg}$} & \multicolumn{1}{c}{$MSE_{avg}\pm std_{avg}$} \\ \hline
1/1                   & (0, 3019)    & 4.00e-4$\pm$ 2.84e-5                        & 4.00e-4$\pm$ 2.91e-5                        & \textbf{9.19e-5 $\pm$ 2.28e-6}           \\
4/4                   & (0, 3019)    & 3.00e-4$\pm$ 7.80e-7                        & 3.00e-4$\pm$ 1.68e-6                        & \textbf{2.00e-4$\pm$ 1.43e-5}               \\ \hline
\end{tabular}
}
\caption{\textcolor{black}{Numerical results on the DJIA dataset at the missing data imputation task, for ODE-RNN, CODE-RNN and CODE-BiRNN.}}
\label{tab:DJIA1.1}
\end{table}

\begin{table}[h]
\centering
\resizebox{\textwidth}{!}{
\begin{tabular}{lcccc}
\hline
\multicolumn{1}{c}{}  &              & ODE-GRU                  & CODE-GRU                 & CODE-BiGRU                         \\ \hline
seen/predict features & $(t_0, t_f)$ & $MSE_{avg}\pm std_{avg}$ & $MSE_{avg}\pm std_{avg}$ & $MSE_{avg}\pm std_{avg}$           \\ \hline
1/1                   & (0, 3019)    & 4.00e-4$\pm$ 2.18e-5    & 3.00e-4$\pm$ 2.17e-5    & \textbf{8.96e-5 $\pm$ 4.68e-6} \\
4/4                   & (0, 3019)    & 3.00e-4$\pm$ 1.31e-6    & 3.00e-4$\pm$ 7.50e-6    & \textbf{1.00e-4$\pm$ 8.62e-6}     \\ \hline
\end{tabular}
}
\caption{\textcolor{black}{Numerical results on the DJIA dataset at the missing data imputation task, for ODE-GRU, CODE-GRU and CODE-BiGRU.}}
\label{tab:DJIA1.2}
\end{table}

\begin{table}[h]
\centering
\resizebox{\textwidth}{!}{
\begin{tabular}{lcccc}
\hline
\multicolumn{1}{c}{}  &              & ODE-LSTM                 & CODE-LSTM                & CODE-BiLSTM                    \\ \hline
seen/predict features & $(t_0, t_f)$ & $MSE_{avg}\pm std_{avg}$ & $MSE_{avg}\pm std_{avg}$ & $MSE_{avg}\pm std_{avg}$       \\ \hline
1/1                   & (0, 3019)    & 2.00e-4$\pm$ 1.97e-6    & 2.00e-4$\pm$ 1.11e-6    & 2.00e-2$\pm$ 9.58e-6          \\
4/4                   & (0, 3019)    & 3.00e-4$\pm$ 3.68e-6    & 4.00e-4$\pm$ 4.56e-6    & \textbf{2.00e-4$\pm$ 9.03e-6} \\ \hline
\end{tabular}
}
\caption{\textcolor{black}{Numerical results on the DJIA dataset at the missing data imputation task, for ODE-LSTM, CODE-LSTM and CODE-BiLSTM.}}
\label{tab:DJIA1.3}
\end{table}

Figures \ref{fig:djiaLoss1} and \ref{fig:djiaLoss2} show that CODE-BiRNN/-BiGRU/-BiLSTM achieve faster convergence and lower $MSE_{avg}$ values, out of all architectures. When comparing ODE-RNN/-GRU/-LSTM and CODE-RNN/-GRU/-LSTM they present similar loss evolution being ODE-GRU able to achieve slightly lower $MSE_{avg}$ values than CODE-GRU.

\paragraph{Future extrapolation}

\begin{figure}[h]
\centering
\begin{subfigure}{0.329\textwidth}
  \centering
  \includegraphics[width=\linewidth]{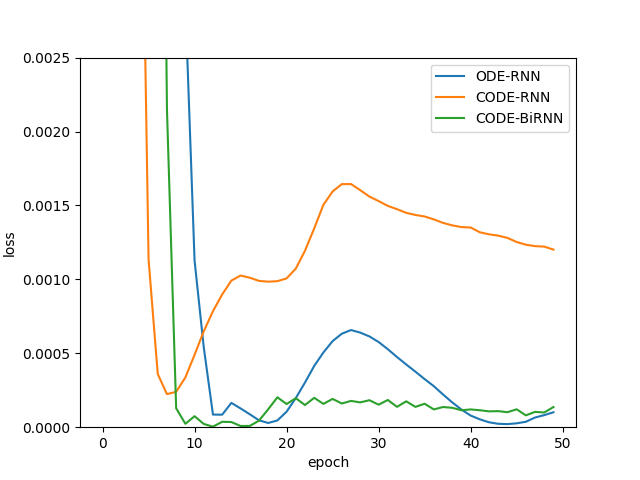}
  \caption{}
  \label{fig:sub1}
\end{subfigure}%
\begin{subfigure}{.329\textwidth}
  \centering
  \includegraphics[width=\linewidth]{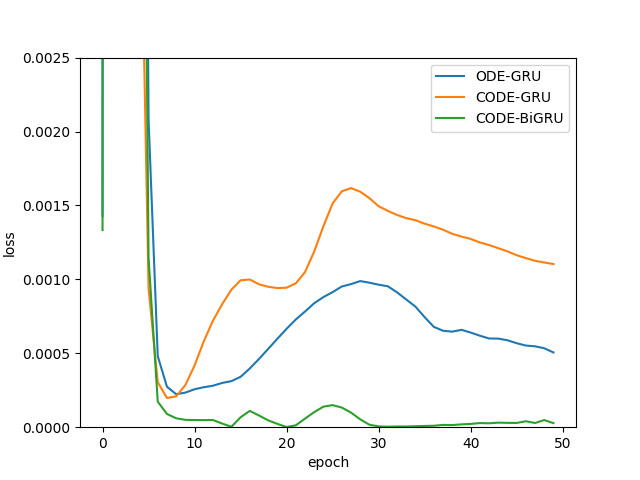}
  \caption{}
  \label{fig:sub2}
\end{subfigure}
\begin{subfigure}{.329\textwidth}
  \centering
  \includegraphics[width=\linewidth]{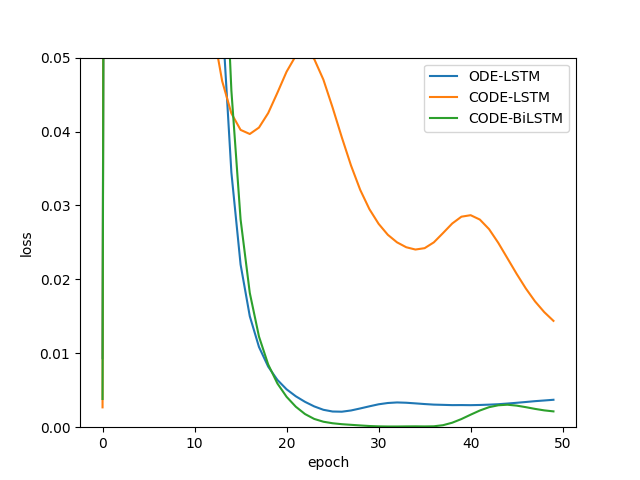}
  \caption{}
  \label{fig:sub2}
\end{subfigure}
\caption{Training loss through the epochs for future extrapolation task for the DC dataset, for $7/7$. (a) ODE-RNN, CODE-RNN, CODE-BiRNN; (b) ODE-GRU, CODE-GRU, CODE-BiGRU; (c) ODE-LSTM, CODE-LSTM, CODE-BiLSTM.}
\label{fig:djiaLoss3}
\end{figure}
\begin{figure}[h]
\centering
\begin{subfigure}{0.329\textwidth}
  \centering
  \includegraphics[width=\linewidth]{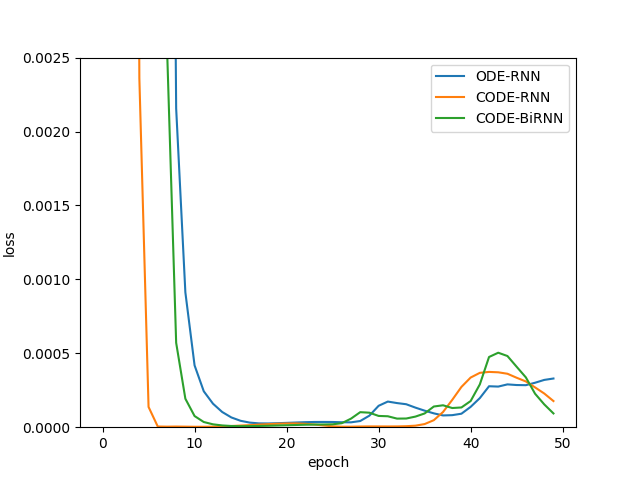}
  \caption{}
  \label{fig:sub1}
\end{subfigure}%
\begin{subfigure}{.329\textwidth}
  \centering
  \includegraphics[width=\linewidth]{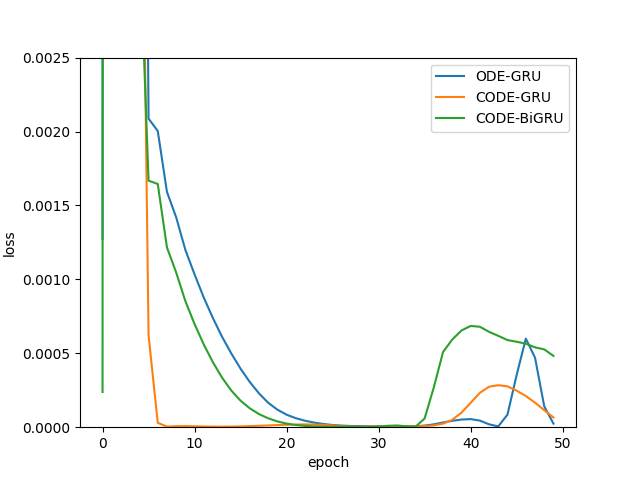}
  \caption{}
  \label{fig:sub2}
\end{subfigure}
\begin{subfigure}{.329\textwidth}
  \centering
  \includegraphics[width=\linewidth]{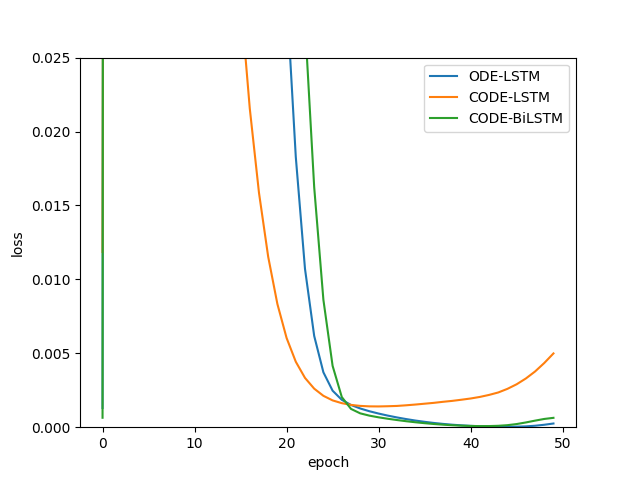}
  \caption{}
  \label{fig:sub2}
\end{subfigure}
\caption{Training loss through the epochs for future extrapolation task for the DC dataset, for $15/15$. (a) ODE-RNN, CODE-RNN, CODE-BiRNN; (b) ODE-GRU, CODE-GRU, CODE-BiGRU; (c) ODE-LSTM, CODE-LSTM, CODE-BiLSTM.}
\label{fig:djiaLoss4}
\end{figure}

We performed the future extrapolation task to predict the stock's price at close for the next 7 or 15 days after receiving the stock's price at close for past 7 or 15 days, denoted by 7/7 and 15/15.
The numerical results are shown in Table \ref{tab:DJIA2.1} for architectures ODE-RNN, CODE-RNN and CODE-BiRNN, Table \ref{tab:DJIA2.2} for architectures ODE-GRU, CODE-GRU, CODE-BiGRU and Table \ref{tab:DJIA2.3} for architectures ODE-LSTM, CODE-LSTM and CODE-BiLSTM.  
The evolution of the losses during training for predictions with 7/7 and 15/15 are depicted in Figures \ref{fig:djiaLoss3} and \ref{fig:djiaLoss4}, respectively.

The results in Tables \ref{tab:DJIA2.1}-\ref{tab:DJIA2.3} show that, CODE-BiRNN/-BiGRU/-BiLSTM stand out from the others, consistently achieving the best predictive performance, for 7/7 and 15/15. Among these variants, CODE-BiGRU offers the best performance for 7/7 while CODE-BiRNN presents the lowest $MSE_{avg}$ for 15/15.
CODE-LSTM presents the lowest performance out of all networks, for 7/7 and 15/15 and, in general ODE-RNN/-GRU/-LSTM have lower or similar $MSE_{avg}$ values when compared to CODE-RNN/-GRU/-LSTM. \textcolor{black}{This was expected since the architecture of these proposed networks do not consider the information given by the input to update the hidden state passed onto the next iteration.}

\begin{table}[h]
\centering
\resizebox{\textwidth}{!}{
\begin{tabular}{ccccc}
\hline
                  &              & ODE-RNN                  & CODE-RNN                 & CODE-BiRNN                   \\ \hline
days seen/predict & $(t_0, t_f)$ & $MSE_{avg}\pm std_{avg}$ & $MSE_{avg}\pm std_{avg}$ & $MSE_{avg}\pm std_{avg}$     \\ \hline
7/7               & (0, 3019)    & 1.60e-3 $\pm$ 8.00e-4      & 3.12e-2 $\pm$ 1.00e-3      & \textbf{1.10e-3 $\pm$ 2.00e-4} \\
15/15             & (0, 3019)    & 1.60e-2 $\pm$ 6.00e-4      & 1.41e-2 $\pm$ 8.10e-3      & \textbf{8.00e-4 $\pm$ 5.00e-4} \\ \hline
\end{tabular}
}
\caption{\textcolor{black}{Numerical results on the DJIA dataset at the future extrapolation task, for ODE-RNN, CODE-RNN and CODE-BiRNN.}}
\label{tab:DJIA2.1}
\end{table}

\begin{table}[h]
\centering
\resizebox{\textwidth}{!}{
\begin{tabular}{ccccc}
\hline
                  &              & ODE-GRU                  & CODE-GRU                 & CODE-BiGRU                   \\ \hline
days seen/predict & $(t_0, t_f)$ & $MSE_{avg}\pm std_{avg}$ & $MSE_{avg}\pm std_{avg}$ & $MSE_{avg}\pm std_{avg}$     \\ \hline
7/7               & (0, 3019)    & 1.70e-2 $\pm$ 1.30e-3     & 3.10e-2 $\pm$ 3.00e-4      & \textbf{3.00e-4 $\pm$ 5.00e-5}  \\
15/15             & (0, 3019)    & 3.00e-3 $\pm$ 1.40e-3      & 1.81e-2 $\pm$ 1.93e-2      & \textbf{1.80e-3 $\pm$ 1.00e-3} \\ \hline
\end{tabular}
}
\caption{\textcolor{black}{Numerical results on the DJIA dataset at the future extrapolation task, for ODE-GRU, CODE-GRU and CODE-BiGRU.}}
\label{tab:DJIA2.2}
\end{table}

\begin{table}[h]
\centering
\resizebox{\textwidth}{!}{
\begin{tabular}{ccccc}
\hline
                  &              & ODE-LSTM                 & CODE-LSTM                & CODE-BiLSTM                  \\ \hline
days seen/predict & $(t_0, t_f)$ & $MSE_{avg}\pm std_{avg}$ & $MSE_{avg}\pm std_{avg}$ & $MSE_{avg}\pm std_{avg}$     \\ \hline
7/7               & (0, 3019)    & 4.96e-2 $\pm$ 8.00e-4      & 3.19e-1 $\pm$ 1.34e-2     & \textbf{1.13e-2 $\pm$ 3.00e-4} \\
15/15             & (0, 3019)    & 1.69e-2 $\pm$ 4.50e-3      & 3.06e-1 $\pm$ 1.21e-1      & \textbf{3.20e-3 $\pm$ 5.00e-4} \\ \hline
\end{tabular}
}
\caption{\textcolor{black}{Numerical results on the DJIA dataset at the future extrapolation task, for ODE-LSTM, CODE-LSTM and CODE-BiLSTM.}}
\label{tab:DJIA2.3}
\end{table}

From Figures \ref{fig:djiaLoss3} and \ref{fig:djiaLoss4} CODE-BiRNN/-BiGRU/-BiLSTM exhibit the the lowest loss values. 
In general, reveal unstable loss evolution for all architectures, for 7/7, being CODE-BiRNN/-BiGRU/BiLSTM the most stable and providing the lowest $MSE_{avg}$ values throughout the epochs. Furthermore, ODE-RNN/-GRU/-LSTM shows lower $MSE_{avg}$ values than CODE-RNN/-GRU/-LSTM, corroborating the test results in Tables \ref{tab:DJIA2.1}-\ref{tab:DJIA2.3}.
For 15/15, CODE-RNN/-GRU/-LSTM show fastest convergence. The loss value of all models increases around epoch $40$, suggesting the implementation of an early stopping criterion could be beneficial.

\paragraph{Backward extrapolation}


We performed the backward extrapolation task to predict the stock's price at close for the past 7 or 15 days after receiving the stock's price at close for the next 7 or 15 days, denoted by 7/7 and 15/15.
The numerical results are shown in Table \ref{tab:DJIA3.1} for architectures ODE-RNN, CODE-RNN and CODE-BiRNN, Table \ref{tab:DJIA3.2} for architectures ODE-GRU, CODE-GRU, CODE-BiGRU and Table \ref{tab:DJIA3.3} for architectures ODE-LSTM, CODE-LSTM and CODE-BiLSTM.  

From Tables \ref{tab:DJIA3.1}-\ref{tab:DJIA3.3}, the performance of the models on the backward extrapolation task are similar to that observed for the forward extrapolation task, with CODE-BiRNN/-BiGRU/-BiLSTM demonstrating the highest performance for 7/7 and 15/15. Among these variants, CODE-BiGRU performs best.
As expected, when predicting longer time horizons, all architectures show an increase in the $MSE_{avg}$ values compared to smaller time horizons.
Furthermore, in general, CODE-RNN/-GRU/-LSTM perform less effectively than ODE-RNN/-GRU/-LSTM.

\begin{table}[h]
\centering
\resizebox{\textwidth}{!}{
\begin{tabular}{ccccc}
\hline
                  &              & ODE-RNN                  & CODE-RNN                 & CODE-BiRNN                     \\ \hline
days seen/predict & $(t_f, t_0)$ & $MSE_{avg}\pm std_{avg}$ & $MSE_{avg}\pm std_{avg}$ & $MSE_{avg}\pm std_{avg}$       \\ \hline
7/7               & (3019, 0)    & 1.80e-3 $\pm$ 9.00e-4      & 3.19e-2 $\pm$ 1.00e-3      & \textbf{1.40e-3 $\pm$ 2.00e-4}   \\
15/15             & (3019, 0)    & 1.40e-2 $\pm$ 6.00e-4      & 1.36e-2 $\pm$ 7.00e-3     & \textbf{1.00e-3 $\pm$ 5.00e-4} \\ \hline
\end{tabular}
}
\caption{\textcolor{black}{Numerical results on the DJIA dataset at the backward extrapolation task, for ODE-RNN, CODE-RNN and CODE-BiRNN.}}
\label{tab:DJIA3.1}
\end{table}

\begin{table}[h]
\centering
\resizebox{\textwidth}{!}{
\begin{tabular}{ccccc}
\hline
                  &              & ODE-GRU                  & CODE-GRU                 & CODE-BiGRU                   \\ \hline
days seen/predict & $(t_f, t_0)$ & $MSE_{avg}\pm std_{avg}$ & $MSE_{avg}\pm std_{avg}$ & $MSE_{avg}\pm std_{avg}$     \\ \hline
7/7               & (3019, 0)    & 1.76e-2 $\pm$ 1.40e-3      & 3.17e-2 $\pm$ 3.00e-4      & \textbf{6.00e-4 $\pm$ 7.00e-5}  \\
15/15             & (3019, 0)    & 2.20e-3 $\pm$ 9.00e-4      & 1.78e-2 $\pm$ 1.78e-2      & \textbf{2.10e-3 $\pm$ 1.00e-3} \\ \hline
\end{tabular}
}
\caption{\textcolor{black}{Numerical results on the DJIA dataset at the backward extrapolation task, for ODE-GRU, CODE-GRU and CODE-BiGRU.}}
\label{tab:DJIA3.2}
\end{table}

\begin{table}[h]
\centering
\resizebox{\textwidth}{!}{
\begin{tabular}{ccccc}
\hline
                  &              & ODE-LSTM                 & CODE-LSTM                & CODE-LSTM                    \\ \hline
days seen/predict & $(t_f, t_0)$ & $MSE_{avg}\pm std_{avg}$ & $MSE_{avg}\pm std_{avg}$ & $MSE_{avg}\pm std_{avg}$     \\ \hline
7/7               & (3019, 0)    & 1.01e-1 $\pm$ 2.10e-3      & 1.69 $\pm$ 1.06e-1      & \textbf{1.29e-2 $\pm$ 3.00e-4} \\
15/15             & (3019, 0)    & 4.24e-2 $\pm$ 1.47e-2      & 7.32e-1 $\pm$ 2.24e-1      & \textbf{3.80e-3 $\pm$ 6.00e-4} \\ \hline
\end{tabular}
}
\caption{\textcolor{black}{Numerical results on the DJIA dataset at the backward extrapolation task, for ODE-LSTM, CODE-LSTM and CODE-BiLSTM.}}
\label{tab:DJIA3.3}
\end{table}

\footnotetext{For doing predictions backward in time using the $ODESolve$ in ODE-LSTM receives the time interval in the reverse order.}

\clearpage

\section{Conclusion} \label{sec:conclusion}

In this work, we have introduced Neural CODE, a novel NN architecture that adjusts an ODE dynamics such that it models the data both forward and backward in time.
To accomplish this, Neural CODE minimises a loss function based on predictions forward, through solving an IVP, and backward, through solving an FVP, in time.  With this, Neural CODE effectively leverages the connections between data by considering how previous and future values influence the current value, effectively exploiting the inter-dependencies within the data.

Furthermore, we have proposed two innovative recurrent architectures, CODE-RNN and CODE-BiRNN, which use Neural CODE to model the states between observations. Additionally, GRU and LSTM update cell variants were introduced in these architectures originating CODE-GRU/-LSTM and CODE-BiGRU/-BiLSTM architectures.

Our experimental evaluation clearly demonstrates that Neural CODE surpasses Neural ODE in terms of learning the dynamics of a spiral ODE when making predictions both forward and backward in time. This highlights the significant impact of considering data from both directions on the performance of the trained models.
Furthermore, when applied to modeling real-life time series using the recurrent architectures CODE-RNN/-GRU/-LSTM and CODE-BiRNN/-BiGRU/-BiLSTM, our results indicate that CODE-BiRNN/-BiGRU/-BiLSTM stands out as the architecture with the highest accuracy, faster convergence, and the ability to reach the lowest training loss values.
These experimental findings provide strong evidence that Neural CODE-based architectures effectively double the available information for fitting the ODE, enhancing the optimisation algorithm's capabilities and resulting in faster convergence and superior performance. Learning the context from both past and future observations enables the capture of patterns and trends, leading to improved generalisation and increased resilience against possible existing data errors.

Overall, leveraging both the forward and backward solutions of an ODE proves to be a powerful approach for data fitting, offering superior performance, generalisation, and robustness compared to solely relying on forward solutions. However, it is crucial to consider the trade-off between these benefits and the associated computational resources and complexity.

While this work sheds light on the advantages of Neural CODE, there remains an open question regarding the impact of different merging operations on the performance of recurrent Neural CODE architectures, including CODE-RNN, CODE-GRU, CODE-LSTM, CODE-BiRNN, CODE-BiGRU, and CODE-BiLSTM. Further exploration in this area could provide valuable insights into optimising the performance of these architectures.



\acks{The authors acknowledge the funding by Fundação para a Ciência e Tecnologia (Portuguese Foundation for Science
and Technology) through CMAT projects UIDB/00013/2020 and UIDP/00013/2020.
C. Coelho would like to thank FCT for the funding through the scholarship with reference 2021.05201.BD.}

\vskip 0.2in
\bibliography{library.bib}
\bibliographystyle{unsrt}

\end{document}